\documentclass[Afour,sageh,times]{sagej}

\usepackage{moreverb,url}
\usepackage{appendix}
\usepackage{subfig}
\makeatletter
\let\originalappendix\appendix
\renewcommand{\appendix}{%
  \originalappendix%
  \setcounter{figure}{0}%
  \renewcommand{\thefigure}{\@Alph\c@section.\arabic{figure}}%
  \setcounter{equation}{0}
  \renewcommand{\theequation}{\@Alph\c@section.\arabic{equation}}

}
\makeatother

\usepackage[colorlinks,bookmarksopen,bookmarksnumbered,citecolor=red,urlcolor=red]{hyperref}

\newcommand\BibTeX{{\rmfamily B\kern-.05em \textsc{i\kern-.025em b}\kern-.08em
T\kern-.1667em\lower.7ex\hbox{E}\kern-.125emX}}

\def\volumeyear{XXXX}

\theoremstyle{plain}

\theoremstyle{remark}
\newtheorem{rem}{\protect\remarkname}
\theoremstyle{definition}

\usepackage{babel}
\providecommand{\definitionname}{Definition}
\providecommand{\remarkname}{Remark}
\providecommand{\theoremname}{Theorem}

\begin{document}


\title{Data-Driven Soft Robot Control via Adiabatic Spectral Submanifolds}

\author{Roshan S. Kaundinya\affilnum{1}, John Irvin Alora\affilnum{2}, Jonas G. Matt\affilnum{3}, Luis A. Pabon\affilnum{2}, Marco Pavone\affilnum{2} and George Haller\affilnum{1}}

\affiliation{\affilnum{1}Institute for Mechanical Systems, ETH Z\"{u}rich\\
\affilnum{2}Autonmous Systems Lab, Stanford University\\
\affilnum{3}Automatic Control Laboratory, ETH Z\"{u}rich}

\corrauth{George Haller, Institute for Mechanical Systems, ETH Z\"{u}rich, CH}

\email{georgehaller@ethz.ch}

\begin{abstract}
The mechanical complexity of soft robots creates significant challenges for their model-based control. Specifically, linear data-driven models have struggled to control soft robots on complex, spatially extended paths that explore regions with significant nonlinear behavior. To account for these nonlinearities, we develop here a model-predictive control strategy based on the recent theory of adiabatic spectral submanifolds (aSSMs). This theory is applicable because the internal vibrations of heavily overdamped robots decay at a speed that is much faster than the desired speed of the robot along its intended path. In that case, low-dimensional attracting invariant manifolds (aSSMs) emanate from the path and carry the dominant dynamics of the robot. Aided by this recent theory, we devise an aSSM-based model-predictive control scheme purely from data. We demonstrate the effectiveness of our data-driven model in tracking dynamic trajectories across diverse tasks. We validate on high-fidelity, high-dimensional finite-element models of a soft trunk robot and Cosserat-rod-based elastic soft arms, with additional experiments confirming robust performance even in the presence of experimental noise. Notably, we find that five- or six-dimensional aSSM-reduced models outperform the tracking performance of other data-driven modeling methods by a factor up to $10$ across all closed-loop control tasks.
\end{abstract}


\maketitle
\section{Introduction}

Soft robots are primarily built of compliant materials, which renders them flexible and dexterous in navigating complex environments, adapting to unpredictable interactions, and manipulating delicate objects with ease. These key characteristics enable the use of these robots in surgical assistance (see \citet{medical_robotics}), underwater deep sea exploration (see \citet{li_self-powered_2021}), and space engineering (see \citet{Nahar2017RobotTL}). In these applications, the softness of the robots allows them to safely interact with humans and the environment. However, to overcome the challenges of their delicate and demanding environments, these robots require controllers with precision and real-time constraints. 

Soft robots are more challenging to control than rigid robots because they are underactuated, and hence can produce complex and unpredictable responses (see \citet{actuation_review18}, \citet{soft_actuators_review2} and \citet{soft_robot_review} for reviews). In practice, given the complexity of the actuating mechanisms, one typically designs motions for the robot that are slow enough to ensure accurate positioning, an approach widely used in surgical soft robots. For instance, \citet{mis_slow21} build a soft robotic manipulator based on hydraulic actuation for laser-assisted micro-surgery with a maximal speed of $2 \text{ [mm/s]}$. \citet{slow_mis_3} use a snake-like endoscope based on cable-driven actuation for cardiac surgery. In that application, the maximal operating speed is $20 \text{ [mm/s]}$. Even small-scale insect-like soft robots powered by a dielectric actuation achieve a maximum speed of only $30 \text{ [mm/s]}$ (see \citet{slow_insect_1}).  

From modeling perspective, the softness coupled with geometric design constraints already cause soft robot motions to be intrinsically nonlinear. At the same time, soft robot geometries tend to be symmetric to mimic naturally occurring animal morphologies (e.g., an elephant trunk), adding small time-periodic actuation to these structures can trigger a nonlinear resonant response (see \citet{li22a} and \citet{axas24}). Moreover, soft robots are typically built without precise prior knowledge of their governing constitutive laws, which further complicates model development. Thus, when modeling these robots, we seek simple yet accurate models and rely on feedback control to attenuate modeling errors. 

Most precision control problems for soft robots involve dynamic trajectory tracking, often handled by model predictive control (MPC). MPC schemes require low-dimensional models for effective performance. For a soft robot, an accurate physical model is a finite-element model whose complexity is an obstacle to its use in MPC. Additionally, soft robots in practice are built without prior precise knowledge of the constitutive laws that govern them.

To mitigate the trade-off between model accuracy and complexity, three main modeling approaches are common in the literature: (a) model-free policies, (b) physics-based models, and (c) dynamics inspired models. The reinforcement learning approach of \citet{pmlr-v28-levine13} seeks to learn an effective control policy using a neural network (NN) that maps states directly to control inputs. However, fitting the various parameters of such a network requires generating large amounts of training data offline (see \citet{Thuruthel2019ModelBasedRL}, \citet{naughton21}, \citet{pmlr-v229-jitosho23a}, and \citet{alessi_23}) or careful tuning and updating of NN's parameters online (see \citet{pique2022}). All this is especially challenging for real-world soft robots due to their high-dimensionality and unpredictable interactions with the environment. To this end, these approaches rely on data generated by numerical models or black-box-type neural network models of the robot. Hence, even model-free methods end up relying heavily on a sufficiently accurate model of the robot's dynamics. 

In stark contrast to approach (a), (b) is grounded in strong physical assumptions on the robotic system. A commonly used physics-based assumption for soft robots is the piecewise constant curvature (PCC) approximation (see \citet{pcc2010} for a review). This assumes that the soft robot is composed of rigid material segments, and each segment comes with a fixed constant curvature. The simplicity of the PCC approximations is appealing but comes at the cost of losing model accuracy (see \citet{pcc_main}). Other approaches train neural networks with physical constraints which are derived from assuming differentiable actuation forces (\citet{Gao2024}), soft robot kinematics (\citet{bern2020}), or specific potential energy functions (\citet{Stozle}) or strain relations (\citet{valadas_2025}). Although innovative, these methods only provide black-box approximations of the relevant physical quantities, thus still complicating their use in control tasks.

Finally, in category (c), the simplest approach is to fit a linear system of differential equations to the available data. To this end, dynamic mode decomposition (DMD) and extended DMD inspired control methodologies (see \citet{bruder21}) have been developed for closed-loop control. These methods, however, have two major limitations: first, they assume that the input data evolves linearly in a postulated set of variables, which will happen with probability zero in realistic experiments  (see \citet{haller24_linear}). Second, these methods fit to data from nonlinear, time-dependent systems, resulting in a low level of predictability under previously unseen control forces. For these reasons, \citet{haggerty23} proposed  a modified training approach that learns static steady state data as linear functions of the input forces, and a linear nonautonomous dynamical model. This method results in a controller that operates the robot in high-acceleration regimes, but still suffers from the general drawbacks of linear modeling approaches. 

A recent approach in category (c) that addresses the limitations of linear modeling methods is based on the mathematical theory of spectral submanifolds (SSMs). SSMs are low-dimensional attracting invariant surfaces in the phase space of a physical system that are tangent to dominant eigenspaces of fixed points in nonlinear dynamical systems. Rigorous theories for SSMs in autonomous systems (see \citet{haller16} and \citet{haller23}) and in forced, non-autonomous dynamical systems are now available (see \citet{haller24_wa}). The theory has also been implemented in practical settings by learning SSMs from data. Specifically, the \textit{SSMLearn} algorithm of \citet{cenedese22a} has been successfully used to identify low-dimensional nonlinear reduced models for fluid sloshing experiments (see \citet{axas22}), fluid-structure interaction problems (see \citet{Xu_Kaszas_Cenedese_Berti_Coletti_Haller_2024}), shear flows (see \citet{kaszas22}), pipe flows (see \citet{Kaszas_Haller_2024}), non-smooth jointed structures (see \citet{morsy2024reducingfiniteelementmodels}) and MPC  for a  continuum diamond robot (see \citet{alora24}). 

Notably, \citet{alora24} show that SSM-reduced MPC schemes achieve the best control performance when compared with the linear modeling methods we have discussed earlier. More recently, \citet{yan2024refinedmotioncompensationsoft}, used data-driven SSM-based MPC to control laser-assisted procedural tasks. They find that SSM-based nonlinear reduced models offer the best performance when compared to extended DMD and the PCC methods.

However, the SSM-reduced models used thus far to model soft robots were constructed near the equilibria of the uncontrolled robots. As a consequence, the accuracy of these models degrades for control tasks that involve large excursions from equilibria. To extend the accuracy of SSM-based reduced models for such excursions, we exploit here the slow nature of typical soft robot motions relative to their fast internal decay rates that arise from high internal damping. This natural slow-fast splitting of time scales in soft robot control makes these robots ideal targets for the application of the recent theory of adiabatic SSMs, or aSSMs for short (see \citet{haller24_wa}). Such aSSMs are no longer confined to the neighborhood of equilibria, and hence reduction to these manifolds results in models that are valid on substantially larger domains.

We illustrate this by deriving and validating low-dimensional aSSM-reduced models for a soft trunk robot and a soft elastic arm robot purely from data in the controls context. We then demonstrate superior closed-loop performance of these aSSM-based models across five challenging target tracks that vary in dimension, size, and speed, compared to standard SSM and linear baselines.

\section{Adiabatic SSMs for control}

\subsection{Problem setup}
\label{subsec:problem_setup}
Dynamic trajectory tracking for robots involves (i) defining the practical workspace of the robot, (ii) designing a target trajectory in this workspace, and (iii) finding optimal control inputs that command the robot to move along the prescribed trajectory. Tasks (i) and (ii) are specific to the robots and the operating domain of interest, whereas task (iii) involves a generally applicable method. To this end, model predictive control (MPC) is applied over a user-defined planning horizon, taking (i) and (ii) as inputs to the optimization. 

We fix the workspace as $\mathbb{R}^{w}$ with some positive integer $w$. We then define $\mathbf{z} \in \mathbb{R}^{w}$ as the coordinates of the workspace and $\boldsymbol{\Gamma}(t) \subset \mathbb{R}^{w}$ as the target trajectory defined on a finite time interval $\mathcal{T}=[t_i,t_f]$. This interval is divided into smaller, equally spaced planning horizons of length $\Delta t$. We therefore have $\mathcal{T} = \bigcup_{j=0}^{N-1}(t_j,t_{j+1}]$ where  $t_0 = t_i$, $t_{N} =t_f$, $t_{j+1}-t_{j}=\Delta t$, and $N = \frac{t_f -t_i}{\Delta t}$. In each planning interval $(t_j,t_{j+1}]$, we have to solve the following optimization problem : 
\begin{align}
\label{eq:optim_problem}
    \text{minimize}_{\mathbf{u}(\cdot)}~ \quad
    &\int_{t_j}^{t_{j+1}} \Big(||\mathbf{z}(t)-\mathbf{\Gamma}(t)||^2_\mathbf{Q_z} + ||\mathbf{u}(t)||^2_\mathbf{R_u} \Big) dt \nonumber \\
    \mathrm{subject~to}~ \quad
        & \mathbf{\dot{x}} =  \mathbf{f}(\mathbf{x}) + \mathbf{g}(\mathbf{x},\mathbf{u}(t))  \\
        & \mathbf{x}(t_j) = \mathbf{c}(\mathbf{y}(t_j)) \nonumber \\ 
        & \mathbf{y}(t) = \mathbf{h}(\mathbf{x}(t)) \quad \mathbf{z}(t) = \mathbf{C} \mathbf{y}(t) \nonumber \\
        &\mathbf{z}(t) \in \mathcal{Z}, \quad  \mathbf{u}(t) \in \mathcal{U}, \nonumber
\end{align}
where $\mathbf{x} \in \mathbb{R}^n$ is the phase space of the positions and velocities of the robot, $\mathbf{y} \in \mathbb{R}^p$ is the observable space of the robot and $\mathbf{u} \in \mathbb{R}^{n_u}$ are the control inputs of the robot. $\mathcal{Z}$ is a set of constraints in the workspace, and $\mathcal{U}$ is a compact set of admissible control inputs. $\mathbf{Q_z}\in \mathbb{R}^{w \times w}$ is a positive semi-definite matrix that represents the workspace variables' cost and $\mathbf{R_u}\in \mathbb{R}^{n_u \times n_u}$ is a positive definite matrix that penalizes the control input. The function $\mathbf{h}:\mathbb{R}^{n} \to \mathbb{R}^{p}$ maps the phase space to the observable space. The function  $\mathbf{c}:\mathbb{R}^{p} \to \mathbb{R}^n$ maps the observable space to the phase space and is only guaranteed to exist and be invertible when $p=n$. Lastly, the operator $\mathbf{C}: \mathbb{R}^{p} \to \mathbb{R}^{w}$ maps the observable space to the workspace.

The initial condition $\mathbf{y}(t_j)$ is the observed configuration of the robot, inferred from a camera at a time instant $t_j$, after applying the optimal control inputs that solve the optimal control problem in the previous planning horizon $(t_{j-1},t_j]$. For the first planning horizon $(t_0,t_1]$, we define the starting configuration of the robot as $\mathbf{y}(t_0) = \mathbf{y}_0$. The quantity $\mathbf{y}(t_j)$ acts effectively as a feedback to the optimal control problem defined on the target's horizon $\mathcal{T}$. Hence, an optimal control input $\mathbf{u}(t)$ for the horizon $\mathcal{T}$ is a solution to a closed-loop control problem. If one decides not to use the observed configurations and rely only on the trajectory of the model $\mathbf{x}(t)$, then the optimization problem becomes an open-loop problem. 

In practical applications, the robot's observability is limited, and its actual dynamics are unknown. Models for those dynamics are at best complicated partial differential equations (PDEs). One numerically approximates such a PDE with a very large finite-dimensional set of ordinary differential equations (ODE). Optimizing such a high-dimensional ODE over a given planning horizon is computationally expensive, necessitating a reduced-order model. In the following sections, we will rigorously construct such reduced-order models for soft robots in practical operating domains of interest. We list key symbols and some notation that appear in this paper in Table \ref{tab:symbols}. 

\begin{table}[h]
\label{tab:symbols} 
\centering
\caption{List of key symbols and their definitions}
\label{tab:symbols}
\begin{tabular}{ll}
\toprule
\textbf{Symbol} & \textbf{Definition} \\
\midrule
$\mathbf{x} \in \mathbb{R}^{n}$ & State space coordinates\\
$\mathbf{y} \in \mathbb{R}^{p}$ & Observable space coordinates\\
$\mathbf{z}\in \mathbb{R}^{w}$ & Workspace coordinates \\
$\boldsymbol{\Gamma}(t)$ & Target trajectory \\
$\mathbf{u}\in \mathbb{R}^{n_u}$ & Control input vector\\
$\mathcal{U}$ & Admissible set of control inputs \\
$\mathcal{L}_0$ & Critical manifold geometry \\
$\mathbf{S}:\mathcal{U} \to \mathcal{L}_0$ & Critical manifold map\\ 
$\mathbf{I}:\mathcal{L}_0 \to \mathcal{U}$ & Inverse map of critical manifold\\
$\mathcal{A}_{\epsilon}$ & Adiabatic SSM (aSSM) geometry\\
$\mathbf{r} \in \mathbb{R}^{d}$ & aSSM reduced coordinates \\ 
$\mathbf{u}^s(\epsilon t)$ & Slow control input \\
$\mathbf{V}^{\mathrm{T}}:\mathbb{R}^{n} \times \mathcal{U} \to \mathbb{R}^{d}$ & aSSM chart map \\
$\mathbf{R}:\mathbb{R}^{d} \times \mathcal{U} \to \mathbb{R}^{d}$ & aSSM reduced dynamics\\
$\mathbf{W}:\mathbb{R}^{d} \times \mathcal{U} \to \mathbb{R}^{n}$ & aSSM parametrization map\\
$\mathbf{u}^d(t)$ & Fast control input deviation \\
$\mathbf{B}:\mathcal{U} \to \mathbb{R}^{d\times n_u}$ & aSSM control deviation term \\
$r_s$ & Slowness measure \\
\bottomrule
\end{tabular}
\end{table}

\subsection{Adiabatic SSM reduction}
\label{sec:assmr}
General robot motion is governed by an ODE 
\begin{equation}
\label{eq:governing_equations}
\dot{\mathbf{x}} = \mathbf{f}(\mathbf{x}) + \mathbf{g}(\mathbf{x},\mathbf{u}(t)), \quad \mathbf{x} \in \mathbb{R}^n,
\end{equation}
where $\mathbb{R}^n$ is the phase space of the robot, $\mathbf{u}(t) \in \mathbb{R}^{n_u}$ is the control input, and $\mathbf{f}$ and $\mathbf{g}$ are nonlinear functions. Setting $\mathbf{u}(t)=0$ in eq.(\ref{eq:governing_equations}), results in an autonomous dynamical system describing the motion of an uncontrolled robot: 
\begin{equation}
\label{eq:unforced_equations}
\dot{\mathbf{x}} = \mathbf{f}(\mathbf{x}).
\end{equation}
By design, most physical robots without control have at least one equilibrium $\mathbf{x}^*$ defined by $\mathbf{f}(\mathbf{x}^*)=0$. We can always perform a coordinate transformation to shift this equilibrium to the origin, and hence we will assume $\mathbf{x}^{*}=0$. We also assume this equilibrium to be asymptotically stable, which translates to $\text{Re}[\text{Spec}(\mathbf{D}_{\mathbf{x}} \mathbf{f}(0))]<0$ for the spectrum of $\mathbf{D}_{\mathbf{x}} \mathbf{f}(0)$. These assumptions hold in our setting, since soft robots, by construction, have an asymptotically stable rest position. 

Under appropriate nonresonance conditions among the eigenvalues of $\mathbf{D}_{\mathbf{x}} \mathbf{f}(0)$, a hierarchy of spectral submanifolds (SSMs) anchored to $\mathbf{x}^* =0$ exists (see \citet{haller16}). In our setting here, (primary) SSMs are the unique smoothest invariant manifolds of system (\ref{eq:unforced_equations}) that are tangent to spectral subspaces of $\mathbf{D}_{\mathbf{x}} \mathbf{f}(0)$ at the origin and have the same dimensions as those subspaces. Secondary SSMs with the same invariance and tangency properties also exist, but they are strictly less smooth than the primary SSMs. In this paper, we will only rely on the primary SSMs and refer to them simply as SSMs for brevity. Specifically, slow SSMs are tangent to spectral subspaces spanned by a set of slowest decaying eigenmodes. A unique slow SSM captures most of the dominant dynamics of the system and hence is an ideal candidate for model reduction in system (\ref{eq:unforced_equations}).  A key feature that makes these structures robust for control or forced response prediction is that they are normally attracting, which means that any normal rates of attraction to the SSM overpower all the tangential rates of attraction within the SSM. 

When the control input $\mathbf{u}(t)$ is non-zero and uniformly bounded, generalized time-varying SSMs anchored to generalized stationary states of the same stability type as the uncontrolled origin are proven to exist (see \citet{haller24_wa}). Those generalized SSMs can be rigorously constructed for the system (\ref{eq:governing_equations}) for moderate control inputs. They may not survive for large control inputs, but one can still devise systematic control strategies based on them. 

To address this limitation on the control input magnitude, \citet{haller24_wa} also discuss an alternative case wherein time-dependent SSMs called adiabatic SSMs (aSSMs), exist under slowly varying inputs $\mathbf{u}(t) = \mathbf{u}^s(\epsilon t)$ for $\epsilon \ll 1$ in system (\ref{eq:governing_equations}). This adiabatic setting allows for larger control inputs, but assumes that control inputs vary in time at a rate that is much slower than the internal decay rates of the uncontrolled robot to its fixed point.  Using the formulation of \citet{haller24_wa}, eq.(\ref{eq:governing_equations}) subjected to those slow control inputs can be re-written as the slow-fast system 

\begin{align}
\label{eq:slow-fast-split}
\dot{\mathbf{x}} &= \mathbf{f}(\mathbf{x}) + \mathbf{g}(\mathbf{x},\mathbf{u}), \nonumber \\ 
\dot{\mathbf{u}} &= \epsilon \mathbf{D}_{\alpha} \mathbf{u}^s(\alpha), \\ \nonumber
\dot{\alpha} &= \epsilon. 
\end{align}

Setting $\epsilon =0$ in these, we obtain

\begin{align}
\label{eq:frozen-limit}
\dot{\mathbf{x}} &= \mathbf{f}(\mathbf{x}) + \mathbf{g}(\mathbf{x},\mathbf{u}), \nonumber \\ 
\dot{\mathbf{u}} &= 0, \\ \nonumber
\dot{\alpha} &= 0.
\end{align}
This is an autonomous system of ODEs similar to (\ref{eq:unforced_equations}) but describes the dynamics of the robot subjected to constant control inputs $\mathbf{u}(t) \equiv \mathbf{\bar{u}}$. We will assume that system (\ref{eq:frozen-limit}) admits a critical manifold $\mathcal{L}_0$ (i.e; a manifold of fixed points) of dimension $n_u$ in the form

\begin{multline}
\label{eq:critical_manifold}
\mathcal{L}_0 = \{ (\mathbf{x},\mathbf{u}) \in \mathbb{R}^n \times \mathbb{R}^{n_u} :   \mathbf{x} = \mathbf{S}(\mathbf{u}), \quad \mathbf{u} \in \mathcal{U} \quad  | \\ \quad  \mathbf{f}( \mathbf{S}(\mathbf{u})) + \mathbf{g}( \mathbf{S}(\mathbf{u}),\mathbf{u}) = 0\}.    
\end{multline}

We further assume that for any fixed $\mathbf{u}(t) \equiv \mathbf{\bar{u}}$, the fixed point $\mathbf{\bar{x}} = \mathbf{S}(\mathbf{\bar{u}})$ of the first equation in (\ref{eq:slow-fast-split}) is asymptotically stable. Under this assumption, we let $\mathbf{A}(\mathbf{u}) = \mathbf{D}_{\mathbf{x}} \left[\mathbf{f}(\mathbf{x}) +  \mathbf{g}(\mathbf{x},\mathbf{u})\right]|_{\mathbf{x}=\mathbf{S}(\mathbf{u})}$ and  $\text{Spec}(\mathbf{A}(\mathbf{u})) = \{\lambda_j(\mathbf{u})\}_{j=1}^{n}$. We further assume that for some non-zero negative real number $K$, strict spectral splitting exists: 
\begin{multline}
    \label{eq:lambda}
\text{Re}[\lambda_n(\mathbf{u})] \leq \text{Re}[\lambda_{n-1}(\mathbf{u})] \dots \leq  \text{Re}[\lambda_{d+1}(\mathbf{u})] <  K \\< \text{Re}[\lambda_d(\mathbf{u})] \leq \dots \leq \text{Re}[\lambda_1(\mathbf{u})] < 0.
\end{multline}

The first $d$ eigenvalues correspond to slow eigenvectors that span a slow spectral subspace denoted $E(\mathbf{u})$. From \citet{haller16}, a unique smoothest $d$-dimensional SSM, $\mathcal{W}(E(\mathbf{u}))$, exists tangent to $E(\mathbf{u})$ and anchored to $\mathbf{x} = \mathbf{S}(\mathbf{u})$, provided that appropriate nonresonance conditions hold among the eigenvalues $\lambda_j(\mathbf{u})$ for all $\mathbf{u} \in \mathcal{U}$. Note that the collection of these SSMs, $\mathcal{A}_0 = \cup_{\mathbf{u} \in \mathcal{U}} \mathcal{W}(E(\mathbf{u}))$, is a $\rho$-normally attracting compact invariant manifold, with 
\begin{equation}
    \label{eq:rho_u}
\rho = \text{min}_{\mathbf{u} \in \mathcal{U}} \text{Int}\Big[\frac{\lambda_{d+1}(\mathbf{u})}{\lambda_{d}(\mathbf{u})}\Big] ,
\end{equation}
as defined by \citet{fenichel73}.

\begin{figure*}
    \begin{centering}
    \subfloat[]{\begin{centering}
    \includegraphics[width=0.3\textwidth]{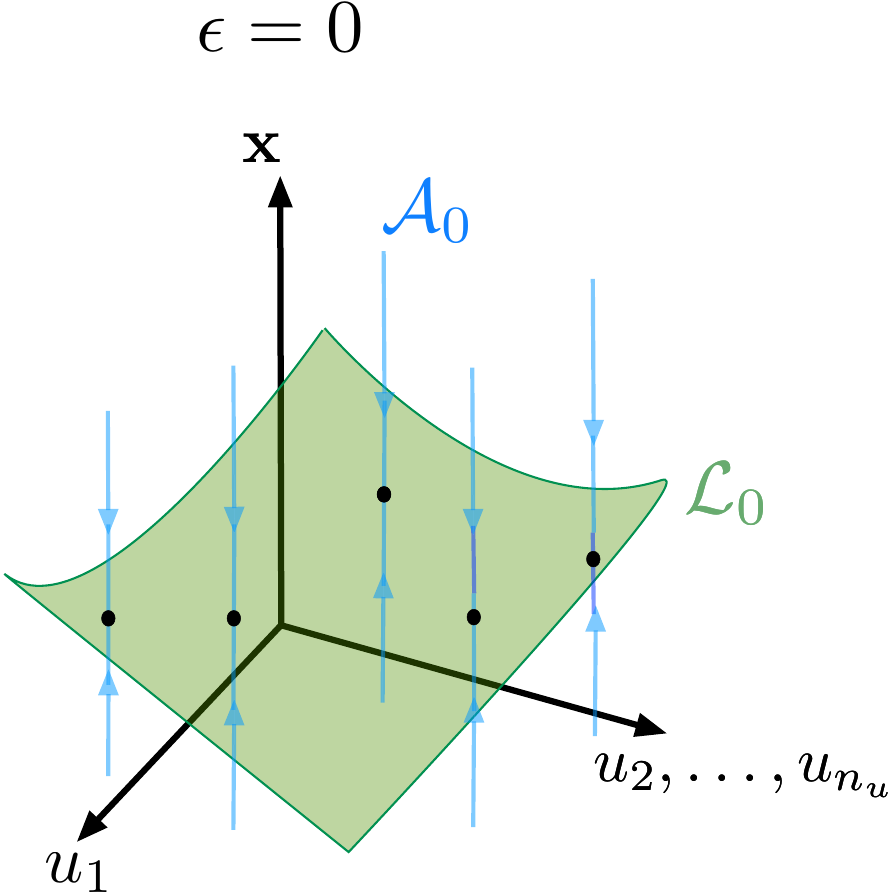}
    \par\end{centering}
    }\subfloat[]{\begin{centering}
    \includegraphics[width=0.3\textwidth]{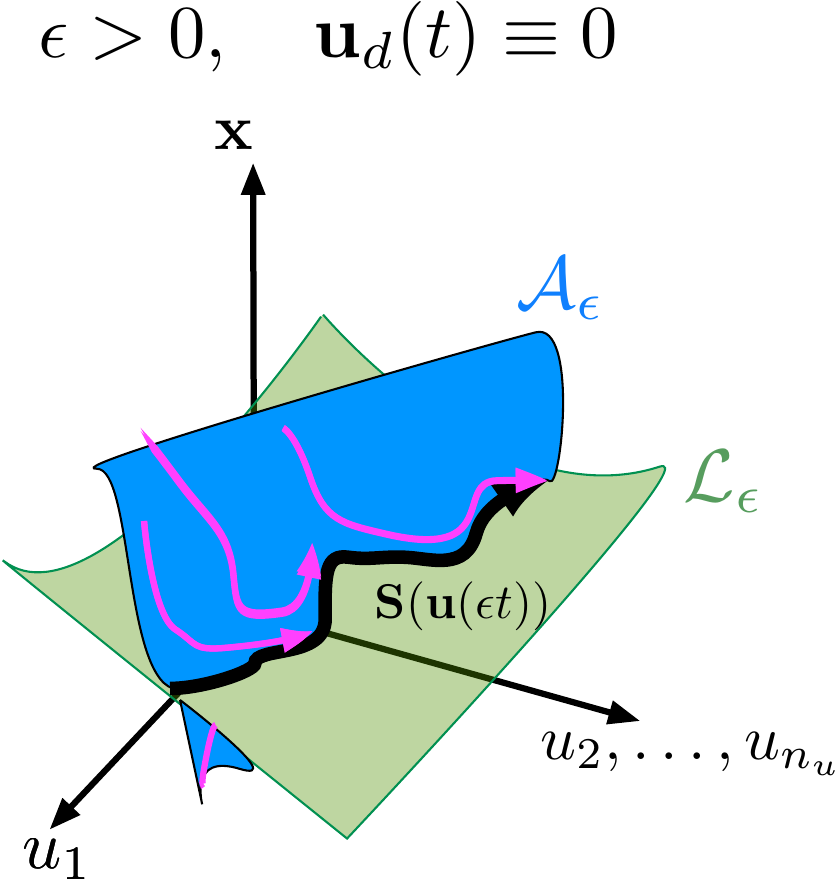}
    \par\end{centering}
    }\subfloat[]{\begin{centering}
        \includegraphics[width=0.3\textwidth]{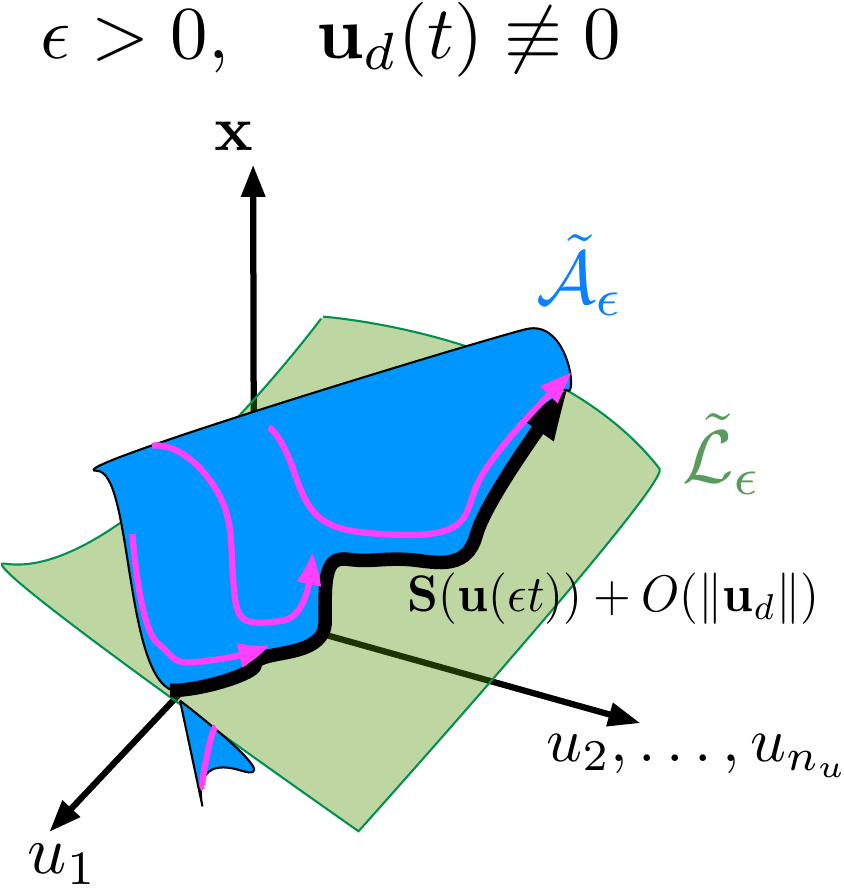}
        \par\end{centering}
        }
    \par\end{centering}
    \caption{(a) For $\epsilon =0$, critical limit of the adiabatic SSM geometry in the phase and actuation space. (b) For $\epsilon >0$ and slow input $\mathbf{u}(\epsilon t)$, the leading order adiabatic SSM geometry of $\mathcal{A}_\epsilon$ anchored to the target $\mathbf{S}(\mathbf{u}(\epsilon t))$. (c) For $\mathbf{u}_d(t) \not \equiv 0$, the perturbed aSSM geometry of $\tilde{\mathcal{A}}_{\epsilon}$.}
    \label{fig:idea_assm}
    \end{figure*}

For $\epsilon >0$, by Theorem 6 of \citet{haller24_wa}, the manifold $\mathcal{A}_0$ survives as a nearly diffeomorphic invariant manifold $\mathcal{A}_{\epsilon}$, which we call an adiabatic SSM (aSSM). The critical manifold $\mathcal{L}_0$ will also survive as a slow trajectory $\mathcal{L}_{\epsilon}$ that acts as an anchor to the aSSM. We can seek $\mathcal{L}_{\epsilon}$ and $\mathcal{A}_{\epsilon}$ as Taylor expansions in $\epsilon$ using the governing equations. Let $\mathbf{V}(\mathbf{u}) \in \mathbb{R}^n \times \mathbb{R}^{d}$ be a matrix whose columns form the basis of the spectral subspace $E(\mathbf{u})$, and let 
\begin{equation}
    \mathbf{r} = \mathbf{V}^{\mathrm{T}}(\mathbf{u}) (\mathbf{x}-\mathbf{S}(\mathbf{u})) \in \mathbb{R}^d
\end{equation}
denote a local coordinate along $E(\mathbf{u})$. Note that $\mathcal{A}_0$ can be parametrized as $\mathcal{A}_0 = \{\mathbf{x}\in \mathbb{R}^n: \mathbf{x} = \mathbf{W}(\mathbf{r},\mathbf{u})\}$ with an appropriate smooth function $\mathbf{W}: \mathbb{R}^d \times \mathcal{U} \subset \mathbb{R}^{n_u} \to \mathbb{R}^n$. Along the unperturbed SSM $\mathcal{A}_0$, we then have 
\begin{equation}
    \label{eq:aSSM_maps}
    \mathbf{x} = \mathbf{W}(\mathbf{r},\mathbf{u}(\alpha)) + \mathbf{S}(\mathbf{u}(\alpha)).
\end{equation}
The dynamics on the perturbed aSSM $\mathcal{A}_{\epsilon}$ can be written as   
\begin{align}
\label{eq:aSSM_reduced}
   \dot{\mathbf{r}} &= \mathbf{R}(\mathbf{r}, \mathbf{u}(\alpha)) + o(\epsilon,\epsilon|\mathbf{r}|), \\ \nonumber
    \dot{\alpha} &= \epsilon. 
\end{align}
 
At leading-order, the aSSM is anchored along the slow trajectory $\mathbf{x}^*(\epsilon t) \approx \mathbf{S}(\mathbf{u}(\epsilon t))$ which lies on the critical manifold $\mathcal{L}_0$. All trajectories of the aSSM-reduced system (\ref{eq:aSSM_reduced}) will approach $\mathcal{L}_0$ in forward time for any user-defined slow input $\mathbf{u}(\epsilon t)$ with a small enough $\epsilon > 0$ (see Fig. \ref{fig:idea_assm}b). Note that we have chosen the reduced coordinate such that the critical manifold $\mathcal{L}_0$ satisfies $\mathbf{r}=0$.

\begin{rem}
\textbf{[Chart map for the critical manifold $\mathcal{L}_0$]}
For closed-loop control applications, we will need the chart map of the critical manifold $\mathcal{L}_0$. We expect this map to be a homeomorphism, since we know from the physics of the robot that an arbitrary static control input maps to a unique asymptotically stable steady state. We define the chart map $\mathbf{I}: \mathcal{L}_0 \to \mathcal{U}$ as the inverse of $\mathbf{S}: \mathcal{U} \to  \mathcal{L}_0$, mapping steady states on the critical manifold to commensurate inputs.
\end{rem}    

\begin{rem}
\textbf{[Slow target prediction]}
The leading-order reduced dynamics on the aSSM predicts approach and synchronization to a unique slow trajectory $\mathbf{S}(\mathbf{u}(\epsilon t))$ that is generated by the slow control force $\mathbf{u}(\epsilon t)$. In  practice, closed-loop control will still be required to ensure the accurate tracking of the uncontrolled trajectory due to the uncertainties and approximations involved. 
\end{rem}

\begin{rem}
\textbf{[Zeroth-order aSSM approximation]}
 We can recover the SSM-reduced model that describes the unforced robot's motion by setting $\mathbf{u}(\epsilon t)\equiv \mathbf{0}$ and $\mathbf{S}(\mathbf{u})\equiv 0$ in eqs.(\ref{eq:aSSM_maps})-(\ref{eq:aSSM_reduced}).
 This gives us the origin-based SSM parametrization  $\mathbf{W}(\mathbf{r},\mathbf{0})$, projection $\mathbf{V}(\mathbf{0})$, and reduced dynamics $\mathbf{R}(\mathbf{r},\mathbf{0})$, providing a zeroth-order approximation of the aSSM-reduced model (\ref{eq:aSSM_reduced}). This approximation comprises an SSM anchored to the $\mathbf{x} = \mathbf{0}$ equilibrium of the soft robot and is described by the equations
 \begin{align}
    \label{eq:zeroth-order-aSSM}
    \mathbf{r} &= \mathbf{V}^\top(\mathbf{0}) \mathbf{x}, \nonumber\\ 
    \mathbf{x} &= \mathbf{W}(\mathbf{r},\mathbf{0}) , \\ 
    \dot{\mathbf{r}} &= \mathbf{R}(\mathbf{r},\mathbf{0}) + o(\epsilon,\epsilon |\mathbf{r}|). \nonumber
 \end{align}
 
 In \citet{alora23}, this approximation is combined with a model predictive control scheme to provide controlled response predictions close to the equilibrium of the robot.   
\end{rem}

\begin{rem}
\textbf{[First-order aSSM approximation]}
A next level of approximation to the aSSM-reduced dynamics (\ref{eq:aSSM_maps})-(\ref{eq:aSSM_reduced}) is obtained by assuming that the aSSM parametrization is constant along the path $\mathbf{S}(\mathbf{u}(\alpha))$ and equals to $\mathbf{S}(\mathbf{0})$. The first-order approximation to the aSSM yields the reduced model 
\begin{align}
    \label{eq:first-order-aSSM}
        \mathbf{r} &= \mathbf{V}^\mathrm{T}(\mathbf{0}) (\mathbf{x}-\mathbf{S}(\mathbf{u}(\alpha))) , \nonumber \\ 
        \mathbf{x} &= \mathbf{W}(\mathbf{r}, \mathbf{0}) + \mathbf{S}(\mathbf{u}(\alpha)),  \\ \nonumber
        \dot{\mathbf{r}} &= \mathbf{R}(\mathbf{r}, \mathbf{0}) + o(\epsilon,\epsilon|\mathbf{r}|), \\ \nonumber
        \dot{\alpha} &= \epsilon. \nonumber
\end{align}
This approximation to the aSSM-reduced dynamics is effectively a translation of the zeroth-order aSSM approximation along the response generated by a slow input. It requires less effort to compute compared to the aSSM-reduced model (\ref{eq:aSSM_maps})-(\ref{eq:aSSM_reduced}). 
\end{rem}

\subsection{Predictive capability of aSSM-reduced models for controls}
\label{sec:assmr_perturb}
In dynamic trajectory tracking problems, we are interested in determining the control inputs required to move the robot along a prescribed track. In contrast, in Section \ref{sec:assmr}, we described a reduction methodology for the inverse problem in which a reduced model predicts a track under given control inputs. To solve the original trajectory tracking problem via this reduction, we can substitute the aSSM-reduced model for the full nonlinear dynamical system in the optimization problem (\ref{eq:optim_problem}). In that case, however, we need to employ an additional fast control component that seeks to eliminate the error of the slow, aSSM-based controller. 

Specifically, we express the general control input as $\mathbf{u}(t) = \mathbf{u}^s(\epsilon t) + \mathbf{u}^d(t)$, where $\mathbf{u}^s(\epsilon t)$ is a slow control input and $\mathbf{u}^d(t)$ is a fast control input deviation. We assume $\mathbf{u}^d(t)$ to be uniformly bounded, i.e., let $|\mathbf{u}^d(t)| \leq \delta$ for some $\delta>0$ and for all $t \geq 0$. We extend the non-autonomous system (\ref{eq:governing_equations}) to an autonomous system by letting 
\begin{align}
    \label{eq:extended_slow-fast}
    \dot{\mathbf{x}} &= \mathbf{f}(\mathbf{x}) + \mathbf{g}(\mathbf{x}, \mathbf{u}^s(\epsilon t)) + \mathbf{D}_{\mathbf{u}} \mathbf{g}(\mathbf{x},\mathbf{u}^s(\epsilon t)) \mathbf{u}^d(t) + o(|\mathbf{u}^d|),\nonumber \\
    \dot{t} &= 1. 
\end{align}

When $\mathbf{u}^d(t)\equiv 0$, system (\ref{eq:extended_slow-fast}) is equivalent to system (\ref{eq:slow-fast-split}), for which we have already obtained the aSSM-reduced model (\ref{eq:aSSM_reduced}). We will next show how this aSSM-reduced model carries over to system (\ref{eq:extended_slow-fast}) for $\mathbf{u}^d(t)\not \equiv 0$ small enough. 

The aSSM, $\mathcal{A}_{\epsilon}$, of system (\ref{eq:slow-fast-split}) is constructed by the results of \citet{eldering13} as a small perturbation of the normally hyperbolic invariant manifold $\mathcal{A}_0$ of system (\ref{eq:frozen-limit}). These results do not directly imply that $\mathcal{A}_{\epsilon}$ is also normally hyperbolic, but can be strengthened to do so under further technical assumptions (see Remark 3.319 of \cite{eldering13}). To avoid introducing technicalities beyond the scope of this paper, we simply assume that $\mathcal{A}_{\epsilon}$ is normally hyperbolic. Then, by Theorem 3.1 of \citet{eldering13}, $\mathcal{A}_{\epsilon}$ perturbs into a nearby, diffeomorphic invariant manifold $\tilde{\mathcal{A}}_{\epsilon}$ in system (\ref{eq:extended_slow-fast}) for small enough $\delta >0$ uniform bound in $|\mathbf{u}_d(t)|$. We depict the perturbed aSSM geometry of $\tilde{\mathcal{A}}_{\epsilon}$ in Fig. \ref{fig:idea_assm}c. 

The perturbed aSSM-reduced model on $\tilde{\mathcal{A}}_{\epsilon}$ at leading order in $|\mathbf{u}^d|$ and $\epsilon$ is given by 
\begin{align}
    \label{eq:perturbed_assm}
    \mathbf{r} &= \mathbf{V}^\mathrm{T} (\mathbf{u}^s(\epsilon t)) (\mathbf{x}-\mathbf{S}(\mathbf{u}^s(\epsilon t)))\nonumber \\ 
    \dot{\mathbf{r}} &= \mathbf{R}(\mathbf{r},\mathbf{u}^s(\epsilon t)) + \mathbf{B}(\mathbf{u}^s(\epsilon t)) \mathbf{u}^d(t) + o(\epsilon, \epsilon |\mathbf{r}|,|\mathbf{r}||\mathbf{u}^d|), \nonumber\\ 
    \mathbf{x} &= \mathbf{W}(\mathbf{r},\mathbf{u}^s(\epsilon t)) + \mathbf{S}(\mathbf{u}^s(\epsilon t)) + o(\epsilon,\epsilon |\mathbf{r}|,|\mathbf{u}^d|,|\mathbf{r}||\mathbf{u}^d|), 
\end{align}
where $\mathbf{B}(\mathbf{u}^s(\epsilon t)) = \mathbf{V}^\mathrm{T}(\mathbf{u}^s(\epsilon t)) \mathbf{D}_{\mathbf{u}}\mathbf{g}(\mathbf{S}(\mathbf{u}^s(\epsilon t)),\mathbf{u}^s(\epsilon t))$ can be interpreted as the effective reduced linear control matrix, that varies along the slow control inputs. As formula (\ref{eq:perturbed_assm}) shows, the leading order correction to the aSSM parametrization can be ignored when the control deviations are small. Equation (\ref{eq:perturbed_assm}) provides us with an aSSM-reduced model that replaces the full system in the finite-time horizon optimization problem (\ref{subsec:problem_setup}).

\begin{rem}
    \textbf{[Control deviation addition to zeroth-order aSSM]}
     For the case when $\mathbf{u}^s(t)\equiv \mathbf{0}$ and the general control input is solely interpreted as a deviation control term $\mathbf{u}(t) \equiv \mathbf{u}^d(t)$. Substituting these control inputs into  eq. (\ref{eq:perturbed_assm}), we arrive at 
     \begin{align}
        \label{eq:perturbed_zeroth_assm}
        \mathbf{r} &= \mathbf{V}^\mathrm{T} (\mathbf{0}) \mathbf{x}, \nonumber \\ 
        \dot{\mathbf{r}} &= \mathbf{R}(\mathbf{r},\mathbf{0}) + \mathbf{B}(\mathbf{0}) \mathbf{u}(t) + o(|\mathbf{r}||\mathbf{u}|), \\ 
        \mathbf{x} &= \mathbf{W}(\mathbf{r},\mathbf{0}) + o(|\mathbf{u}|,|\mathbf{r}||\mathbf{u}|), \nonumber
    \end{align}
     which is a perturbation of the zeroth-order aSSM-reduced model (\ref{eq:zeroth-order-aSSM}). \citet{alora23} use this model for closed loop control of soft robots. They learn the reduced linear control matrix $\mathbf{B}(\mathbf{0})$ from forced data and apply the model (\ref{eq:perturbed_zeroth_assm}) to control the robots along tracks near the equilibria.
\end{rem}

\begin{rem}
    \textbf{[Control deviation addition to first-order aSSM]}
 Including deviation control in the first-order approximation (\ref{eq:first-order-aSSM}) of the aSSM dynamics gives the reduced model
     \begin{align}
        \label{eq:perturbed_first_order_assm}
        \mathbf{r} &= \mathbf{V}^\mathrm{T} (\mathbf{0}) (\mathbf{x}-\mathbf{S}(\mathbf{u}^s(\epsilon t))), \nonumber \\ 
        \dot{\mathbf{r}} &= \mathbf{R}(\mathbf{r},\mathbf{0}) + \mathbf{B}(\mathbf{u}^s(\epsilon t)) \mathbf{u}^d(t) + o(\epsilon, \epsilon |\mathbf{r}|,|\mathbf{r}||\mathbf{u}^d|), \\ 
        \mathbf{x} &= \mathbf{W}(\mathbf{r},\mathbf{0}) + \mathbf{S}(\mathbf{u}^s(\epsilon t)) + o(\epsilon,\epsilon |\mathbf{r}|,|\mathbf{u}^d|,|\mathbf{r}||\mathbf{u}^d|). \nonumber
    \end{align}
    Since $\mathbf{B}$ depends on $\mathbf{S}(\mathbf{u}^s(\epsilon t))$ explicitly, we expect the linear control matrix to change as we translate the zeroth order aSSM along the critical manifold $\mathcal{L}_0$. As a result, system (\ref{eq:perturbed_first_order_assm}) generally has a larger domain of accuracy than the perturbed zeroth-order aSSM-reduced model (\ref{eq:perturbed_zeroth_assm}).
\end{rem}

\subsection{Finite-time horizon predictions using aSSM-reduced models}
\label{sec:fh_horizon}
We now combine our aSSM-reduced models with the finite-time horizon optimization problem defined in eq.(\ref{eq:optim_problem}). Recall, that the finite horizon optimal control problem takes as inputs the initial condition, the planning horizon, the target trajectory, workspace constraints, and actuation space constraints. Since, the planning horizon in practice is fairly short, the aSSM in this interval can be approximated to be the SSM attached to the initial condition of the optimization problem. 

Using notation from section \ref{subsec:problem_setup} and also the chart map of the critical manifold $\mathcal{L}_0$ from Remark 1, we define the aSSM-reduced MPC scheme for the planning horizon $(t_j,t_{j+1}]$ as 
\begin{align}
    \label{eq:aSSM_mpc}
        \text{minimize}_{\mathbf{u}(\cdot)}~ \quad
        &\int_{t_j}^{t_{j+1}} \Big(||\mathbf{z}(t)-\mathbf{\Gamma}(t)||^2_\mathbf{Q_z} + ||\mathbf{u}(t)||^2_\mathbf{R_u} \Big) dt ,\nonumber \\
        \mathrm{subject~to}~ \quad
            & \mathbf{x}(t_j) = \mathbf{c}(\mathbf{y}(t_j)),  \quad \mathbf{u}^s_0 = \mathbf{I}(\mathbf{x}(t_j)), \nonumber \\
            & \mathbf{r} = \mathbf{V}^\mathrm{T} (\mathbf{u}^s_0) (\mathbf{x}-\mathbf{x}(t_j)), \nonumber\\ 
            & \dot{\mathbf{r}} = \mathbf{R}(\mathbf{r},\mathbf{u}^s_0) + \mathbf{B}(\mathbf{u}^s_0) \left( \mathbf{u}(t) - \mathbf{u}^s_0 \right),  \\ 
            & \mathbf{x} = \mathbf{W}(\mathbf{r},\mathbf{u}^s_0) + \mathbf{x}(t_j),  \nonumber \\ 
            & \mathbf{z}(t) = \mathbf{C}\mathbf{h}(\mathbf{x}(t)), \nonumber \\
            &\mathbf{z}(t) \in \mathcal{Z}, \quad  \mathbf{u}(t) \in \mathcal{U}. \nonumber
\end{align}
Here $\mathbf{u}^s_0$ is the static control input needed to generate the static configuration $\mathbf{x}(t_j)$. This is dependent on observations made in the workspace after successful application of the optimal control inputs in the previous horizon $(t_{j-1},t_{j}]$. The equations of motion constraint for the above optimization problem can be derived by setting $\mathbf{u}^s(\epsilon t) = \mathbf{u}^s_0$ and $\mathbf{u}^d(t) = \mathbf{u}(t) - \mathbf{u}^s_0$ in the perturbed aSSM-reduced model eq.(\ref{eq:perturbed_assm}). By setting $\mathbf{u}^s(\epsilon t) = \mathbf{0}$ and $\mathbf{u}^d(t) = \mathbf{u}(t)$ in eq.(\ref{eq:aSSM_mpc}) we arrive at the zeroth-order aSSM-reduced MPC scheme (see Remark 5). A first-order aSSM-reduced MPC scheme is obtained by assuming that the chart, parametrization and reduced dynamics are the same for all possible static control inputs $\mathbf{u}^s_0$ in the aSSM-reduced MPC scheme (see Remark 6). 

The optimization problem (\ref{eq:aSSM_mpc}) can be solved, for instance, using the \textit{GuSTO} method of \citet{pavone192}. \textit{GuSTO} approximates the non-convex optimization problem into sequential convex optimization problems. The method has already been used in the context of solving SSM based MPC schemes in \citet{alora23b}. 

\begin{rem}
    \textbf{[Applicability of aSSM-reduced methods in practice]} 
    To assess the applicability, we need a non-dimensional quantity that measures the speed along the desired track relative to typical internal time scales of the underlying dynamical system. Motivated by similar quantities defined in  \citet{haller24_wa}, we introduce the non-dimensional slowness measure 
\begin{equation}
    r_{s} = \frac{\frac{1}{t_f-t_1}\int_{t_i}^{t_f} \|\dot{ \boldsymbol{\Gamma}}(s)\| ds}{\frac{1}{t_d}\overline{\int_{0}^{t_d} \|\dot{\mathbf{z}}_{\mathbf{u}^s_0}(s)\| ds}}.
\end{equation} 

Where the overbar denotes averages across various constant control inputs $\mathbf{u}^s_0$, $\mathbf{z}_{\mathbf{u}^s_0}(t)$ is a decaying trajectory of system (\ref{eq:frozen-limit}) for that constant input and $t_d$ is the total duration of that decaying trajectory. The slowness measure $r_s$ is a heuristic measure and quantifies the slowness of the target trajectory $\boldsymbol{\Gamma}(t)$ compared to the intrinsic decay rates of the robot. Our theory is mathematically guaranteed to work for $r_s \ll 1$, but will be seen to perform well in practice even up to $r_s \approx 1$. 
\end{rem}

\begin{figure*}
\centering
\includegraphics[width=1\textwidth]{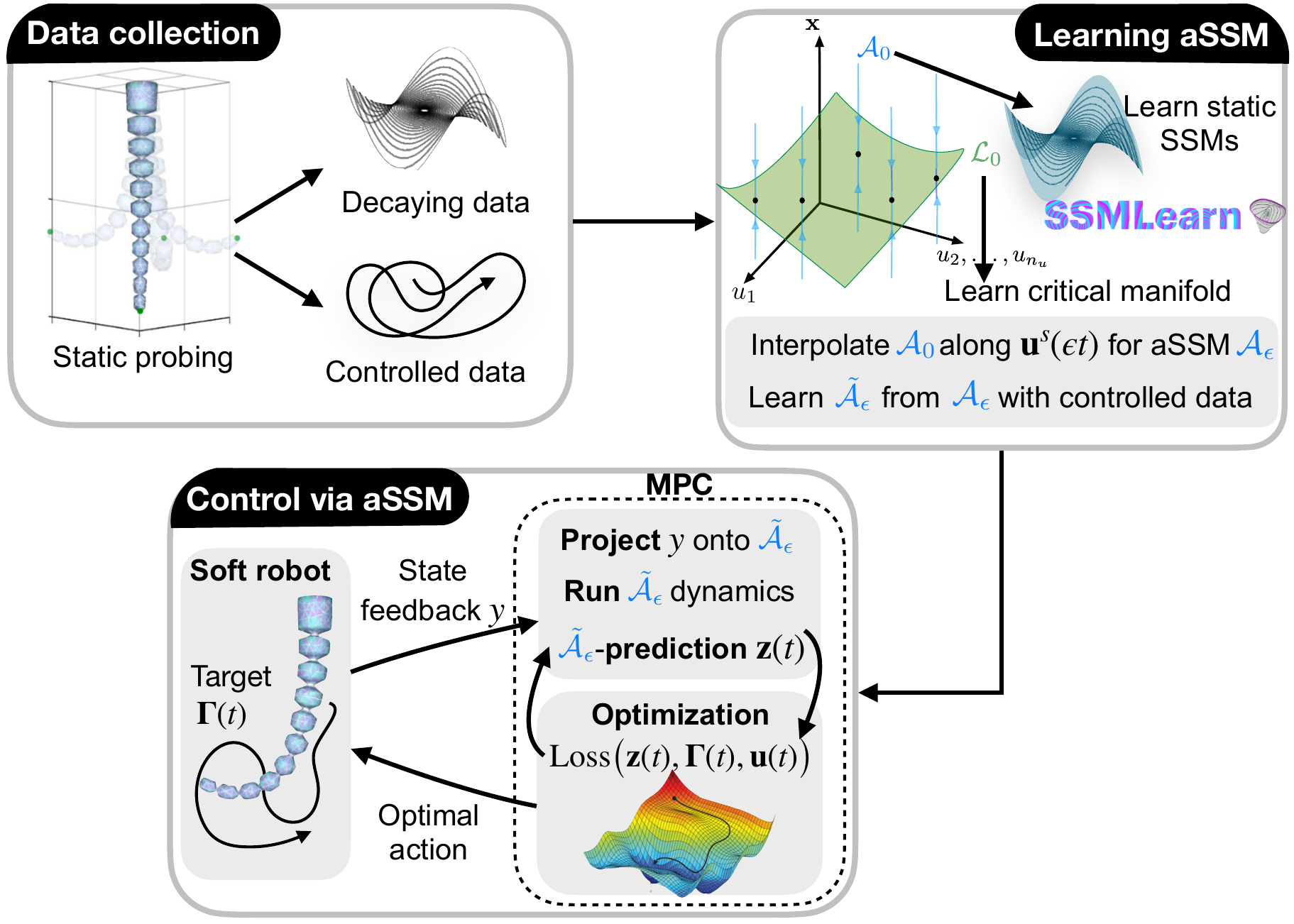}
\caption{Three step procedure for modeling and control of soft robots using aSSMs. Step 1, involves data collection of decaying and controlled data about random static configurations of the soft robot. Step 2, learns the aSSM geometry and dynamics using existing \textit{SSMLearn} algorithm (see \citet{cenedese22b}), for specific details see Section \ref{sec:learn_aSSMs}. Step 3, fuses the learned aSSM-reduced model in a model predictive control scheme, for specific details see Section \ref{sec:fh_horizon}.}
\label{fig:flow_chart_aSSM}
\end{figure*}

We outline our methodology and the control approach in Fig.\ref{fig:flow_chart_aSSM}. The following sections will elaborate in detail the learning methodology for aSSMs.

\section{Learning adiabatic SSMs from data}
\label{sec:data-driven_SSMdictionary}
In practice, one has no access to the governing equations (\ref{eq:governing_equations}) of the robot. At best, we can experimentally probe the robot to obtain trajectory data.  We will also be limited to observing only a few physical coordinates based on sensor availability, and hence $p \ll n$ will hold for the dimension of the observable $\mathbf{y}$ in eq.(\ref{eq:optim_problem}).

Delay embedding of the available observables can be used to increase the dimenion of $\mathbf{y}$. \citet{cenedese22a} and \citet{axas22}, developed the \textit{SSMLearn} and \textit{fastSSM} algorithms to learn a normally attracting, $d$-dimensional SSM embedded in a generic observable space. \citet{alora23b} applied the \textit{SSMLearn} algorithm to learn zeroth-order aSSM-reduced models for soft robots from data lying in a delay-embedded observable space. In what follows, we first briefly recall the procedure from \citet{alora23b}, then describe its extension to more accurate apporximations of aSSMs for controlling soft robots along larger tracks.

\subsection{Learning zeroth-order aSSM-reduced models from data}

\textit{SSMLearn} takes as input trajectory data collected from unforced experiments of the system. The data comprises decaying trajectories starting from a few initial conditions that ultimately converge to the equilibrium of the uncontrolled system.
From a spectrogram analysis of this data and observed symmetries of the underlying physical system, we estimate the minimal dimension $d$ of the dominant slow SSM to which these trajectories converge. With probability one, trajectories will not lie on the slow SSM, but truncating them to their final sections will provide an accurate approximation to the dominant slow SSM. We can use the Takens embedding theorem for delay-embedded observables to deduce the minimal embedding dimension $p=2d+1$ for the SSM to embed smoothly in the $p-$ dimensional observable space. The SSMLearn algorithm chooses the observable space's coordinate frame such that resting position of the robot is always at the origin. 

The truncated decaying data is arranged into a snapshot matrix $\mathbf{Y} \in \mathbb{R}^{p\times N}$. Furthermore, we define $N_{\mathrm{traj}}$ as the number of collected decaying trajectories, if we assume each trajectory has length $n_{i}$, then $N = \sum_{i=1}^{N_{\mathrm{traj}}} n_i$. Performing singular value decomposition (SVD) on $\mathbf{Y}$ and collecting the $d$ dominant singular vectors gives an approximation of the tangent space $\mathbf{V}(\mathbf{0})\in \mathbb{R}^{p \times d}$ of the zeroth-order aSSM.  The zeroth-order aSSM parametrization and the autonomous part of its reduced dynamics are then obtained by polynomial regression. Specifically, we fit the expressions 
\begin{align}
    \label{eq:ssm_param}
    \mathbf{y} = \mathbf{W}(\mathbf{r},\mathbf{0}) &= \mathbf{W}^\star_{\mathbf{0}} \mathbf{r}^{1:n_{w}}, \\
    \dot{\mathbf{r}} = \mathbf{R}(\mathbf{r},\mathbf{0}) &= \mathbf{R}^\star_{\mathbf{0}} \mathbf{r}^{1:n_{r}},
\end{align}
via least-square minimization to trajectory data to determine the parameter matrices $\mathbf{W}^\star_{\mathbf{0}}$ and $\mathbf{R}^\star_{\mathbf{0}}$. Here $\mathbf{r}^{1:n}$ is a column vector comprising all the monomials built out of the coordinate components of $\mathbf{A}$ from order $1$ to $n$. Therefore, we have $\mathbf{r}^{1:n} \in \mathbb{R}^{1 \times m_{n}}$ with $m_{n} = \binom{d+n}{d} -1$. $\mathbf{W}^\star_{\mathbf{0}}$ denotes the optimal parametrization coefficient matrix of size $\mathbb{R}^{p \times m_{n_w}}$ and $\mathbf{R}^\star_{\mathbf{0}}$ is the optimal reduced dynamics coefficient matrix of size $\mathbb{R}^{d \times m_{n_r}}$. The orders $n_r$ and $n_w$ are chosen in a way to avoid overfitting to the training data. We use the normalized mean trajectory error (NMTE) on test predictions $\mathbf{y}_{\mathrm{pred}}(t)$ to optimize these dimensions, where 
\begin{equation}
    \text{NMTE} = \sum_{i=1}^{N_{\mathrm{traj}}} \frac{\| \mathbf{y}_{\mathrm{pred}}(t_i) - \mathbf{y}_{\mathrm{true}}(t_i) \|}{N_{\mathrm{traj}}\text{max}(\| \mathbf{y}_{\mathrm{true}}(t) \|)}.
\end{equation} 
For more detail, see \citet{Kaszas24_2}. To learn the effect of the control deviation term on the SSM-reduced dynamics (\ref{eq:perturbed_zeroth_assm}), we generate controlled training data by subjecting the robot to random time-varying control inputs $\mathbf{u}(t)$. We arrange the controlled data and the random inputs as snapshot matrices $\mathbf{Y}_{u} \in \mathbb{R}^{p \times N_{t}}$ and $\mathbf{U} \in \mathbb{R}^{n_{u} \times N_{t}}$, where $N_{t}$ is the length of the controlled trajectory. The effective reduced linear control matrix $\mathbf{B}(\mathbf{0})$ is then selected as the solution to the minimization problem, 
\begin{multline}
    \label{eq:b0}
    \mathbf{B}^\star(\mathbf{0}) = \operatorname*{arg\,min}_{\mathbf{B}(\mathbf{0})} \| \mathbf{V}^\mathrm{T}(\mathbf{0})\dot{\mathbf{Y}}_{u} \\ - \mathbf{R}_0 (\mathbf{V}^\mathrm{T}(\mathbf{0})\mathbf{Y}_{u})^{1:n_r}  - \mathbf{B}(\mathbf{0})\mathbf{U} \|^2.
\end{multline}

Using this data-driven, zeroth-order aSSM-reduction methodology as a template, we will next discuss methods to learn a $d$-dimensional aSSM-reduced model's chart map, parametrization and reduced dynamics.   

\subsection{Learning higher-order aSSM-reduced models from data}
\label{sec:learn_aSSMs}
\subsubsection{Pointwise static SSM models}
Observe that in our designed control scheme (\ref{eq:aSSM_mpc}) the aSSM-reduced model is approximated by a SSM model attached to a steady state generated by a constant control inputs $\mathbf{u}^s$. We refer to this construct as a static SSM model. Also recall that in the critical limit (\ref{eq:frozen-limit}), the aSSM geometry at leading order comprises a collection of static SSM models. Motivated by this observation, we collect further decaying training data $\mathbf{Y}_{\mathbf{u}^s} \in \mathbb{R}^{p \times N}$ about constant controlled configurations in the robot's workspace. Specifically, to mimic experiments, we collect this type of data for $N_u$ random distinct static control inputs $\{\mathbf{u}^s_{i}\}_{i=1}^{N_u}$. These will lead to a scattered grid of steady states in the observable space, $\{\mathbf{y}^s_i\}_{i=1}^{N_u}$. 

At each of these random steady states, we do not expect any strong symmetries to dictate the static SSM's dimension. If there are special steady states that require a higher dimensional static SSM-reduced model, we pick this as our maximal dimension and keep it fixed across our learning method for the various static SSMs. Similarly, we also keep fixed the polynomial orders for the static SSM parametrization $n_{w}$ and reduced dynamics $n_{r}$. We also center the input training data for each steady state, 
\begin{equation}
    \label{eq:shift_ss}
    \bar{\mathbf{Y}}_{i} =\mathbf{Y}_{\mathbf{u}^s_i} - \mathbf{y}^s_i.
\end{equation}

By repeatedly applying the \textit{SSMLearn} algorithm to each steady state training data $\bar{\mathbf{Y}}_{i}$, we obtain a set of coefficients for the static SSM parametrization, chart map and reduced dynamics. We explicitly define these parameter sets, respectively, as 
\begin{multline}
     \mathbf{V}_{s} = \Big\{ \mathbf{V}_{\mathbf{y}^s_i} \in \mathbb{R}^{p \times d} \Big\}_{i=1}^{N_{u}}, \\ 
    \mathbf{W}_{s} = \Big\{ \mathbf{W}^\star_{\mathbf{y}^s_i} \in \mathbb{R}^{p \times m_{n_{w}}}\Big\}_{i=1}^{N_{u}}, \\ 
    \mathbf{R}_{s} = \Big\{ \mathbf{R}^\star_{\mathbf{y}^s_i} \in \mathbb{R}^{d \times m_{n_{r}}}\Big\}_{i=1}^{N_{u}}.
\end{multline}

Since the \textit{SSMLearn} algorithm uses the numerical SVD procedure, the singular vectors $\mathbf{V}_{y^s_i}$ spanning the tangent space can switch orientations across the training datasets. To this end, we enforce a uniform orientation of the tangent vectors across the dataset to ensure compatible parameter sets.

\subsubsection{Control calibration for static SSMs}
\label{sec:control_calib}
We collect controlled training data by passing bounded random control inputs $\mathbf{u}(t)$ about each static control input $\mathbf{u}^s_i$. Specifically, we generate a random deviation  control input sequence $\mathbf{u}^d(t)$ such that each entry of the control input satisfies $|\mathbf{u}^d_i(t)|\leq \delta$ for all $i\leq n_{u}$. Next, we subject the robot to $N_u$ control inputs $\mathbf{u}_i(t) = \mathbf{u}^d(t) + \mathbf{u}^s_i$, which results in controlled responses that we arrange into a snapshot matrix $\mathbf{Y}^i_{\mathbf{u}}$ and control sequence matrix $\mathbf{U}_{i}$. We further shift the controlled data $\mathbf{Y}^i_{\mathbf{u}}$ (see eq.(\ref{eq:shift_ss})) to $\bar{\mathbf{Y}}^i_{\mathbf{u}}$.

The effective linear control matrix $\mathbf{B}(\mathbf{u}_s)$ appearing in eq.(\ref{eq:aSSM_mpc}) is calculated by solving the minimization problem, 
\begin{multline}
    \label{eq:control_calib}
    \mathbf{B}(\mathbf{u}^i_s) =  \mathbf{B}^\star_{\mathbf{y}^i_s} = \operatorname*{arg\,min}_{\mathbf{B}_{\mathbf{y}^i_s}} \| \mathbf{V}^\mathrm{T}_{\mathbf{y}^i_s} \dot{\bar{\mathbf{Y}}}^i_{\mathbf{u}} \\ - \mathbf{R}_{\mathbf{y}^i_s} (\mathbf{V}^\mathrm{T}_{\mathbf{y}^i_s} \bar{\mathbf{Y}}^i_{\mathbf{u}})^{1:n_r} - \mathbf{B}_{\mathbf{y}^i_s}\mathbf{U}_i \|^2,   
\end{multline} 
where $\dot{\bar{\mathbf{Y}}}^i_{\mathbf{u}}$ is obtained by finite-difference of $\bar{\mathbf{Y}}^i_{\mathbf{u}}$.
This leads to a parameter set of linear constant control action matrices $\mathbf{B}_s = \Big\{ \mathbf{B}^\star_{\mathbf{y}^i_s} \in \mathbb{R}^{d \times n_u} \Big\}_{i=1}^{N_u}$. In the upcoming section, we will need the set of static control inputs $\mathbf{u}^s_i$, which we define as $\mathbf{U}_s = \Big\{ \mathbf{u}^s_{i} \in \mathbb{R}^{n_u \times 1} \Big\}_{i=1}^{N_u}$. 

\subsubsection{Interpolating an aSSM-reduced model from static SSM models}
\label{sec:interp_aSSM}
We can approximate the aSSM-reduced model in the finite horizon optimal control problem (OCP) (eq.(\ref{eq:aSSM_mpc})), by interpolating the static SSM model coefficients. 

Specifically, given grid coordinates $\mathbf{q}\in \mathbb{R}^{g}$, we define the grid of static states $\mathbf{Q}_s = \Big\{ \mathbf{q}_{\mathbf{y}^s_i}  \in \mathbb{R}^{1 \times g } \Big\}_{i=1}^{N_u} $. We also define a generic coefficient set $\mathbf{Z}_s = \Big\{ \mathbf{Z}_{\mathbf{y}^s_i} \in \mathbb{R}^{z_r \times z_c}\Big\}_{i=1}^{N_u}$, where $\mathbf{Z} \in \{\mathbf{V},\mathbf{W},\mathbf{R},\mathbf{B},\mathbf{U}\}$. We then construct a modified inverse distance weighting interpolant for $\mathbf{Z}_s$, given by
\begin{multline}
    \label{eq:midw}
    \mathbf{Z}(\mathbf{q}) = \mathcal{Q}_{\mathrm{MIDW}}(\mathbf{Z}_s, \mathbf{Q}_s,\mathbf{q}) = \sum_{i=1}^{N_u} \alpha_i(\mathbf{q},\mathbf{Q}_s)  \mathbf{Z}_{\mathbf{y}^s_i}, \\ \quad \alpha_i(\mathbf{q},\mathbf{Q}_s) = \frac{w_{i}(\mathbf{q},R,l)}{\sum_{k=1}^{N_u} w_{k}(\mathbf{q},R,l)}.
\end{multline}
Here the weights are defined by $w_{i}(\mathbf{q},R,l) = \left(\frac{\text{max}\left(0,R-\|\mathbf{q}- \mathbf{q}_{\mathbf{y}^s_i}\|\right)}{R\|\mathbf{q}- \mathbf{q}_{\mathbf{y}^s_i}\|}\right)^{l}$. The constant $R$ is a user specified radius of an open ball around a static state in the grid space. Interpolating a query point inside that open ball gives coefficient values with maximal contributions from values at the center. The constant $l$ is an additional user-defined parameter that controls the influence of the grid point with distance. This method scales well with the dimension of $\mathbf{q}$ and is built specially for scattered grids.

Instead of the interpolation scheme (\ref{eq:midw}), we can use regression methods to sample for aSSM coefficients. We also implement a polynomial regression method defined by 
\begin{multline}
    \label{eq:poly_reg}
    \mathbf{Z}(\mathbf{q}) = \mathcal{Q}_{PR}(\mathbf{Z}_s, \mathbf{Q}_s,\mathbf{q}) = \mathcal{Z}^\star\mathbf{q}^{1:n_q}, \quad   \\ \mathcal{Z}^\star  = \operatorname*{arg\,min}_{\mathcal{Z}} \sum_{i=1}^{N_u} \| \mathcal{Z} \mathbf{q}_{\mathbf{y}^s_i}^{1:n_q} - \mathbf{Z}_{\mathbf{y}^s_i} \|^2  \in \mathbb{R}^{z_r \times z_c \times m_{n_q} }.
\end{multline}
Regression methods tend to not enforce the learned data point values. Rather they provide a coarse approximation that captures the overall trends in variations of the SSM parameter sets. They also do not scale well with the dimension of the grid coordinate $\mathbf{q}$. In general, the choice  between interpolation or regression for obtaining the aSSM-reduced model (\ref{eq:aSSM_mpc}) will be decided based on computational efficiency and accuracy of the methods in open loop.

\subsubsection{Systematic parameter selection for aSSM-reduced models}

The parameters used in aSSM reduction are not hyperparameters in the usual sense of machine learning. Instead of being selected by global optimization in a large parameter space, aSSM parameters are selected individually in a systematic fashion. The parametrization order $n_w$ is chosen based on
lowering the reconstruction error of the test dataset. The reduced dynamics order $n_r$ is chosen such that normalized
mean trajectory error on test data predictions is within 1-5\%. Both optimizations simply involve gradually increasing integer numbers from 2 until a minimum error is reached on unseen test data. The embedding dimension $p$ is inferred from the Takens’ delay-embedding theorem and the dimension of the aSSM inferred from a spectrogram analysis of the training data and symmetries of the robot’s geometry. The interpolation radius $R$ and the quadratic regression parameters are optimized to lower reconstruction errors for test data on the critical manifold. In summary, the choices for these parameters are dictated by several theoretical insights unlike the trial-and-error approach used in hyperparameter tuning of Koopman-based, TPWL, and reinforcement learning methods.

\subsection{Illustrative example: Double pendulum under torque control}
\label{sec:dp_torque}
We apply the method outlined in Sections \ref{sec:assmr}-\ref{sec:assmr_perturb} to construct a perturbed aSSM-reduced model (\ref{eq:perturbed_assm}) for open-loop control of a double pendulum. We consider the observable space $\mathbf{y} \in \mathbb{R}^{p}$ to be the full phase space variable $\mathbf{x} = \{\theta_1,\theta_2,\dot{\theta}_1,\dot{\theta}_2\}$, which renders $p =4$. We use the governing equations of the double pendulum derived by \citet{yang24}, who model the double pendulum as a set of two connected rods. We set the lengths and weights of the rods equal and to $1 \text{ [\text{m}]}$ and $1 \text{ [\text{kg}]}$,  respectively. The system is subject to a gravitational acceleration of $g = 9.8 \text{ [m/s$^2$]}$. We assume linear Rayleigh damping for each rod, with the top rod damped twice as much as the lower one. 

The dimension of the control input $\mathbf{u}$ is $n_u = 2$, representing the torques acting on each angular degree of freedom of the system. We choose the grid coordinates for interpolation to be the angular coordinates of the rods $\mathbf{q} = \{\theta_1,\theta_2\}$. We also perform a linear stability analysis across static torque configurations and find a square region $\theta_1,\theta_2 \in [-\frac{\pi}{9},\frac{\pi}{9}]$ that corresponds to stable static states. We also observe that the assumption in eq.(\ref{eq:lambda}) is satisfied in this region. 

We generate two, $100$-second long decaying training trajectories for $N_u=36$ static torque configurations using random initial conditions. The static states live on a square grid $\mathbf{Q}_s \in [-\frac{\pi}{9},\frac{\pi}{9}]^2.$ We observe that after a suitable truncation of the dataset, a $d=2$ dimensional SSM-reduced model captures the decaying dynamics accurately for all static torques. We identify static states with $\theta_1= -\theta_2$ tend to have more nonlinear SSMs compared to others. Setting $n_w = 3$ and $n_r = 3$ provides us with good static SSM approximations that have average $\mathrm{NMTE} \approx 4\% $ across all test data. Since our interpolation grid is square and regular, we employ a spline-based MATLAB interpolation for the aSSM-reduced model's coefficients. For the control deviation term $\mathbf{u}_d(t)$ in eq.(\ref{eq:perturbed_assm}), we can derive the exact form of the linear control matrix as $\mathbf{B}(\mathbf{q}) = \begin{pmatrix}  \mathbf{0} \\ \mathbf{V}^\mathrm{T}(\mathbf{q})\end{pmatrix} \in \mathbb{R}^{4\times 4}$, because we use the phase space as observable space.

We now compare our learned perturbed $2\mathrm{D}$ aSSM-reduced model with the full model simulation for slow control torques that generate a random slow trajectory in $\mathbf{Q}_s$, combined with a chaotic control deviation. The random slow trajectory $\mathbf{S}(\mathbf{u}^s(\epsilon t))$ is generated by a gradient noise scheme called Perlin noise (see \citet{perlin85}) and is rescaled to fit into the grid window $\mathbf{Q}_s$. The chaotic control deviation is $\mathbf{u}^d(t) = \delta \tilde{x}(t)$, where $\tilde{x}(t)$ is a unit rescaled scalar trajectory of one component from the response of a $3\mathrm{D}$ chaotic Lorenz model (see \citet{lorenz63}) launched from a random initial condition.  

\begin{figure*}
    \begin{centering}
    \includegraphics[width=1\textwidth]{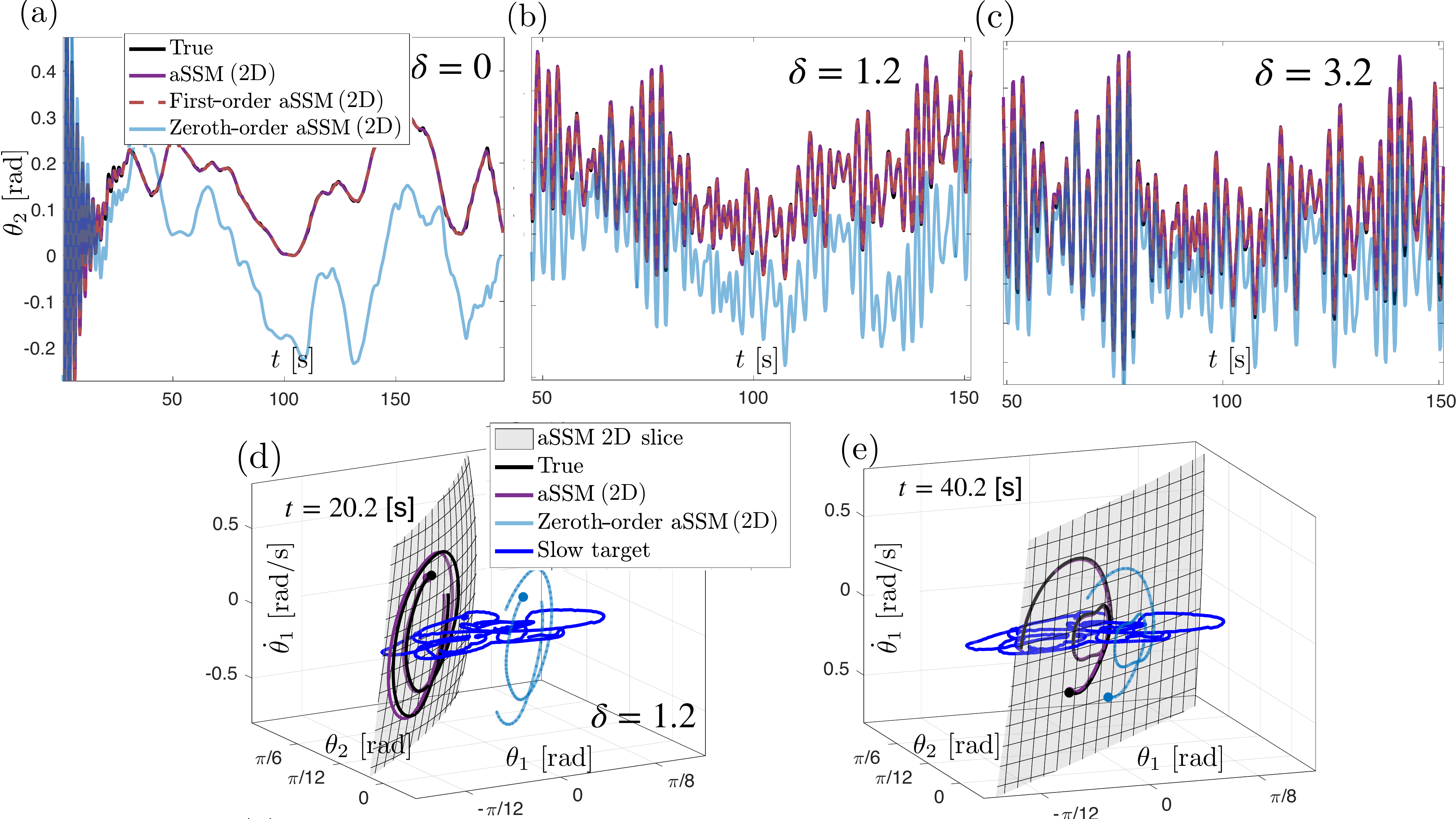}
    \par\end{centering}
    \caption{(a) For $\delta =0$, comparison of different aSSM approximations with the full system response. (b) For $\delta =1.2$, comparison plots in the time window $[50,150]$. (c) The same for $\delta =3.2$. (d) aSSM snapshot in the phase space at $t=20.2 \text{ [s]}$ for $\delta =1.2$. (e) aSSM snapshot at a later time $t=40.2 \text{ [s]}$.}
    \label{fig:double_pendulum}
    \end{figure*}

In Fig. \ref{fig:double_pendulum}a, we compare the different orders of aSSM approximation for an initial condition launched on the aSSM for $\delta =0$ and $\epsilon = 0.005$. The aSSM and its first-order approximation provide high-accuracy predictions. The zeroth-order approximation does well initially but loses accuracy in the long term. Figure \ref{fig:double_pendulum}b depicts the case when $\delta=1.2$, showing that the perturbed aSSM-reduced model and its first-order approximation accurately match the full model. Selecting $\delta=3.2$ yields a similar result, thus proving the robustness of aSSM-reduced models (see Fig. \ref{fig:double_pendulum}c). We also plot in Figs. \ref{fig:double_pendulum}d-e snapshots of the $2\mathrm{D}$ aSSM slices and dynamics in the full phase space. These plots highlight the geometric significance of the aSSM, showing that the true solution lies close to the aSSM slice and is accurately captured by the aSSM-reduced dynamics. The aSSM slice is anchored to the slow target and translates along it changing shape, whereas the prediction of the zeroth-order aSSM approximation remains close to a fixed aSSM slice for all times.

\section{Controlling a soft trunk robot}
\begin{figure*}
    \begin{centering}
    \includegraphics[width=0.8\textwidth]{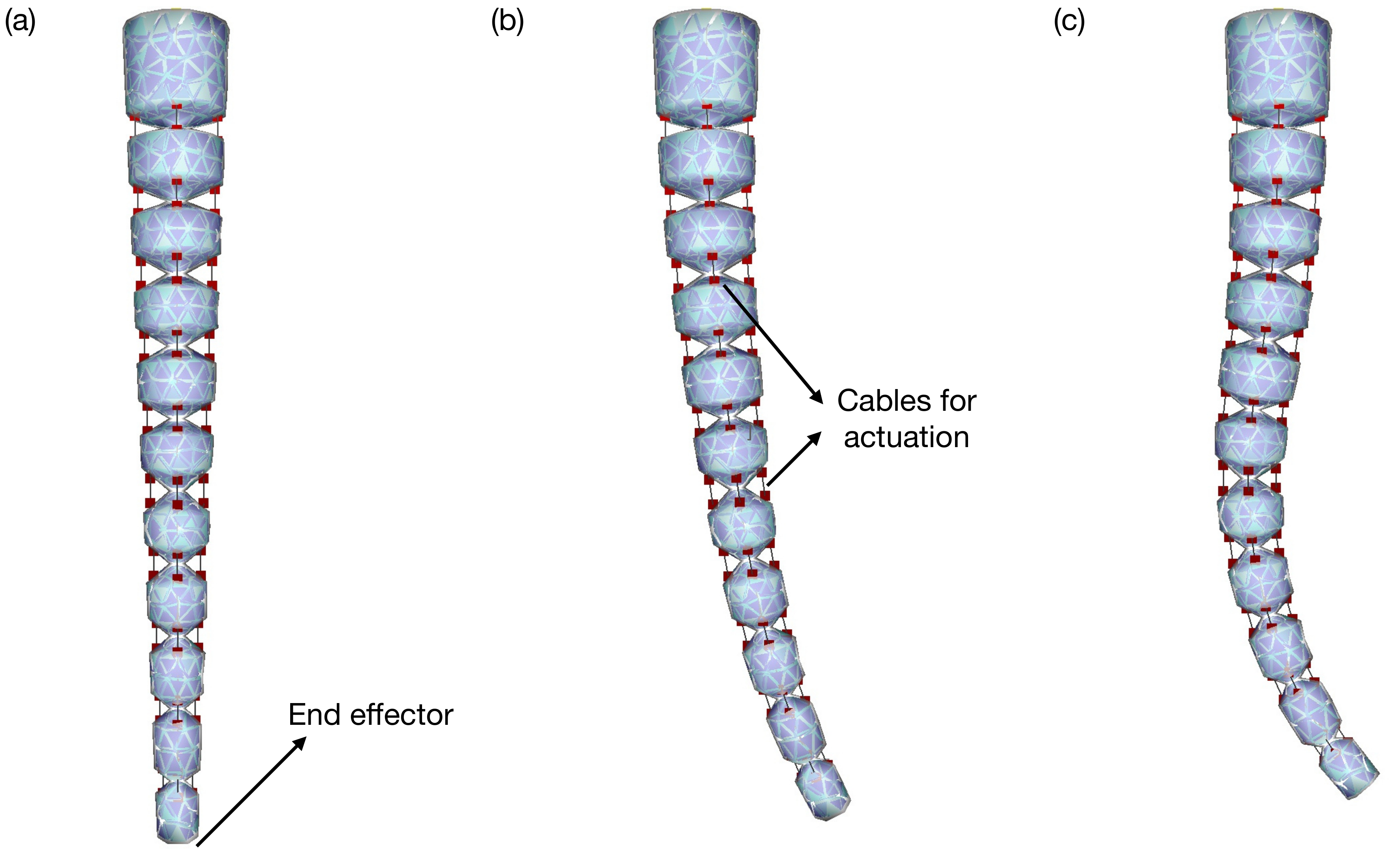}
    \par\end{centering}
    \caption{(a) The trunk at rest. (b) The trunk in a curved static configuration. (c) The trunk in a S-shaped static configuration}
    \label{fig:trunk_snaps}
    \end{figure*}

We now apply our adiabatic MPC control strategy, described in eq.(\ref{eq:aSSM_mpc}), to control a finite-element model of a soft trunk robot. Shown in Fig. \ref{fig:trunk_snaps}, the soft trunk robot weighs  $42 \text{ [g]}$, measures $19.5 \text{ [cm]}$ in height and is made up of soft silicone material with a Poisson's ratio of $0.45$ and a Young's modulus of $0.45 \text{ [MPa]}$. The trunk is actuated using a tensioning mechanism, along its length run four short wires that end at its mid point and four long wires that end at its tip. Hence, the control input space is $n_u = 8$ dimensional with control input $\mathbf{u}\in \mathbb{R}^{8}$. The phase space dimension of the finite-element model of the trunk is $n = 4,254$. 
The open-source package SOFA (see \citet{faure2012sofa}) provides a platform to conduct real-time simulations of this model. 

Our interest is to control the trunk's end effector (lowest point of the trunk) to move along prescribed tracks. Accordingly, we define our workspace to be $\mathbf{z}= \{x_{ee},y_{ee},z_{ee}\} \in \mathbb{R}^3$, where $\mathbf{z}$ represents the position vector of the end effector. We assume to have access only to this positional data for training and testing. SOFA also enforces actuation space constraints given by $\mathcal{U} \in \{|\mathbf{u}_i|\leq 800 \text{ [mN] }, \forall i\leq 8\} $, these constraints ensure the trunk does not break during operation. For any static input in this domain, the trunk robot always admits asymptotically stable configurations, thus satisfying assumption (\ref{eq:critical_manifold}).  

The robot is fixed at its top end and hangs against gravity. 
This geometry, coupled by the physical symmetry of the uncontrolled/unforced trunk in the $x_{ee}-y_{ee}$ directions, implies an approximate $1\colon 1$ resonance between the slowest modes. Depending upon the strength of the damping in the trunk, these slowest decaying modes can be either oscillatory (underdamped) or non-oscillatory (overdamped).

\subsection{Training details of static SSM models}
\begin{figure*}
    \begin{centering}
    \includegraphics[width=1\textwidth]{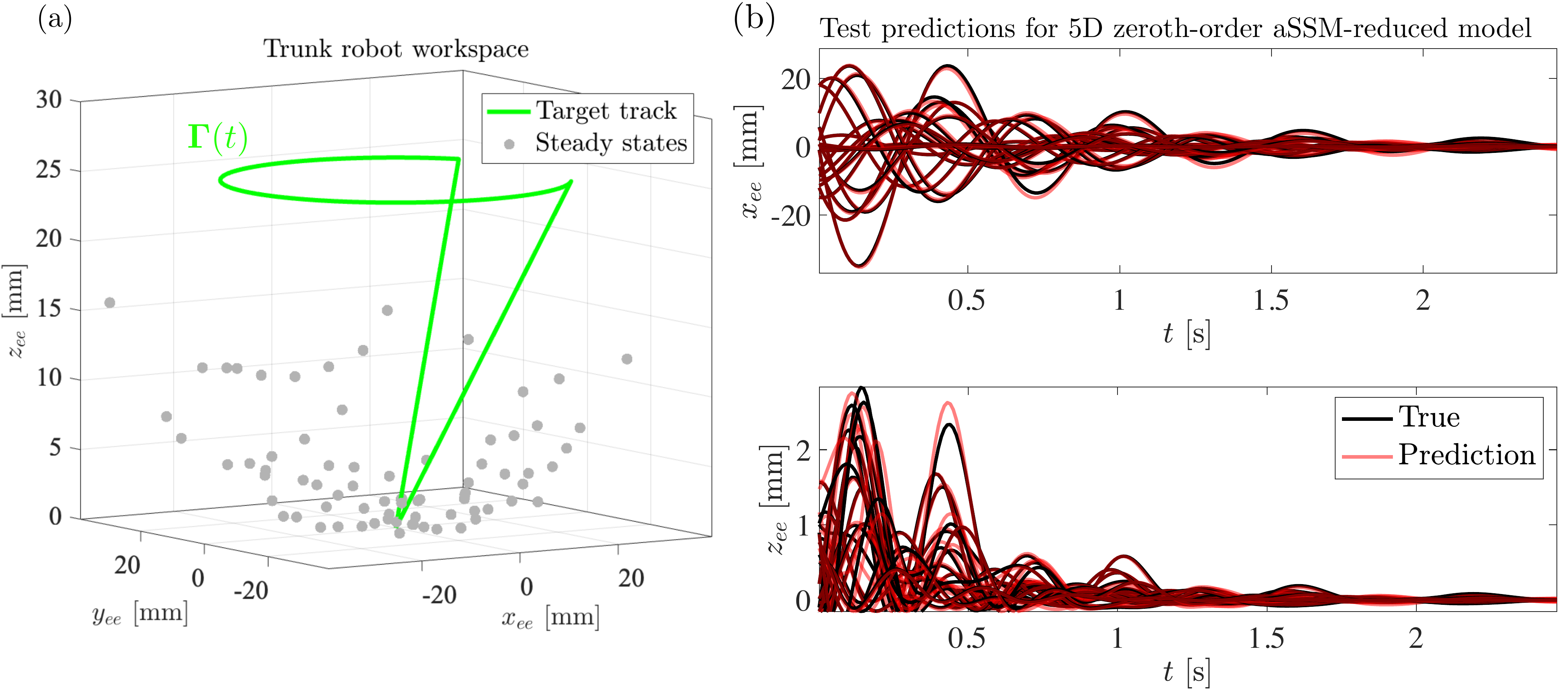}
    \par\end{centering}
    \caption{(a) The workspace of the trunk robot. Static steady state are shown in gray and a target trajectory in green. (b) $x_{ee}$ and $z_{ee}$ test predictions for a $5$D static SSM model anchored at the origin}
    \label{fig:training}
    \end{figure*}

We collect $3$-second long samples of $40$ decaying trajectories for each of $N_u=100$ statically forced equilibrium configurations of the trunk. We split the $40$ decaying trajectories into $15$ for training and the rest for testing. This gives a total of $1500$ training trajectories for the whole workspace of the robot. Each of the $100$ static control input vectors $\mathbf{u}^s$ was generated by random sampling of $100$ distinct combinations of $8$ values from the set $\{0,150,350\}$. $75$ of these static steady states densely cover a spherical cap having radius $20 \text{ [mm]}$ and height $5 \text{ [mm]}$ in the workspace (see Fig. \ref{fig:training}a). We also ensure the combination corresponding to the zero control input vector is part of this collection.

We concentrate on the collected trunk decay data about the unforced configuration. We first split each set of $40$ collected trajectories into $15$ for training and the rest for testing. Next, we truncate the training data to half a second. We follow this up with the construction of a $3p$- dimensional delay embedded observable space $y = (\mathbf{z}(t),\mathbf{z}(t+\tau), \dots ,\mathbf{z}(t+(p-1)\tau))^\top \in \mathbb{R}^{3p}$, where $\tau = 0.01 \text{ [s]}$. 
We observe that for $p=4$, a $d=5$ dimensional SSM-reduced model for $n_w = 4$ and $n_r = 2$ accurately captures the decay in a $12$-dimensional delay embedded observable space, with an average  $\mathrm{NMTE} \approx 3 \%$ on test data (see Fig. \ref{fig:training}b). 

Since the $z_{ee}$ coordinate is directly affected by gravity, the decay data in the coordinate is over-damped. This is reflected by a real eigenvalue of the linear part of our 5D SSM-reduced model. Specifically, the spectrum of the linear part of the $5\mathrm{D}$ zeroth-order aSSM approximation is 
\begin{multline*}
    \lambda_1 = \bar{\lambda}_2 = -1.5799 + i 10.8780,  \\
\lambda_3 = \bar{\lambda}_4 = -1.7705 + i 11.0357, \quad \text{and} \quad
\lambda_5 =  -19.5397. 
\end{multline*}

The eigenvalues $\lambda_1$ and $\lambda_3$ are not in exact resonance, but the gap between them is small compared to $\lambda_5$. This near-resonance arises from the $x_{ee}-y_{ee}$ symmetry of the trunk's equilibrium configuration. The resonance is not exact because the finite-element model of the robot is not c onstrained to respect this symmetry. 

If we truncate the training trajectories at $1$ second, we find that a $4$D zeroth-order aSSM approximation already captures the local dynamics accurately. The spectrum of the reduced dynamics in that case is  
\begin{multline*}
    \lambda_1 = \bar{\lambda}_2 = -1.5337+ i 10.7272,  
 \quad \text{and} \quad\\ 
    \lambda_3 = \bar{\lambda}_4 = -1.5450+ i 10.7968. \quad 
\end{multline*}
The truncation allows us to get a refined spectrum that is near resonant up to the first decimal but comes with the cost of losing the overdamped signatures of the trunk.  

Using the same parameters for the learned 4D and 5D zeroth-order aSSM approximations, we construct the static SSM models for 4D and 5D aSSM-reduced models. We will compare these models in open-loop tests and choose the best performing one for closed-loop tasks.

\subsection{Training details for control calibration sets}

We further collect controlled trajectory data following the procedure outlined in Section \ref{sec:control_calib}. The random control deviation $\mathbf{u}^d(t)$, with a maximum norm of $\delta = 50$, is generated using the Latin hypercube sampling technique (see \citet{latin_hypercube}). We generate one 200-second long random control input deviation. To this we add the random 100 static inputs from earlier and pass the resulting control inputs to the robot to generate controlled responses. We use these controlled responses and their corresponding inputs to learn local linear control matrices $\mathbf{B}_s$ by solving eq.(\ref{eq:control_calib}). We repeat the same procedure but we keep the autonomous reduced dynamics fixed to the zeroth-order aSSM approximation to arrive at $\mathbf{B}_s$ for the first-order aSSM-reduced model. 

For learning $\mathbf{B}_{\mathbf{0}}$ (see eq.(\ref{eq:b0})) for the zeroth order aSSM-reduced approximation, we generate a random control deviation $\mathbf{u}^d(t)$ with $\delta = 350$. We train this model for larger magnitudes of inputs to better anticipate large control deviations that will arise in our closed loop control tasks. 

\subsection{Open-loop testing results}
\label{sec:open_loop}
To test our aSSM-reduced models on unseen general control inputs, we generate new random control input sequences $\mathbf{u}(t)$ with a maximum norm of $\delta = 350$. The trajectory generated by this input traces out a spherical cap in the workspace. We further generate faster versions of the same random input $\mathbf{u}(t)$ that result in faster mean speeds achieved by the trunk robot in the workspace. We run controlled simulations with $\mathbf{u}(t)$ and its faster versions on the trunk robot.

On each controlled response and their corresponding inputs, we perform a detailed validation exercise given by :
\begin{itemize}
    \item Randomly select $50 \%$ points from the control response trajectory.
    \item Use these points as initial conditions to construct an aSSM-reduced model appearing in eq.(\ref{eq:aSSM_mpc}). The construction of the aSSM-reduced model is based on interpolation or regression methods discussed in Section \ref{sec:interp_aSSM}.
    \item Simulate this model for 0.05 seconds using the corresponding general control input values that occur within the time window of the actual response, starting from the same initial condition. 
    \item Calculate the NMTE error metric between the true response of the trunk and the prediction from the aSSM-reduced model.
    \item Plot the initial conditions and shade the plot with a color gradient, where low NMTE values are represented by green and high NMTE values by red. 
\end{itemize}
Further, we perform the above validation on the following aSSM-reduced interpolation models : 
\begin{itemize}
    \item $5$D aSSM-reduced model with modified inverse distance weighting (MIDW) interpolation, with $p=2$ in eq.(\ref{eq:midw}), constructed on a scattered grid in workspace coordinates $\mathbf{q} =\mathbf{z} \in \mathbb{R}^3$. 
    \item $5$D aSSM-reduced model with quadratic polynomial regression (QPR), setting $n_q =2$ in eq.(\ref{eq:poly_reg}), constructed on a scattered grid in the first two reduced coordinates $\mathbf{q} =\{(\mathbf{V}^{\top}_0\mathbf{y})_{1},(\mathbf{V}^{\top}_0\mathbf{y})_{2}\} \in \mathbb{R}^2$.
\end{itemize}
We compare these results with the $4$D aSSM-reduced model counterparts and their zeroth-order approximations as well. 

\begin{figure*}
    \begin{centering}
    \includegraphics[width=1\textwidth]{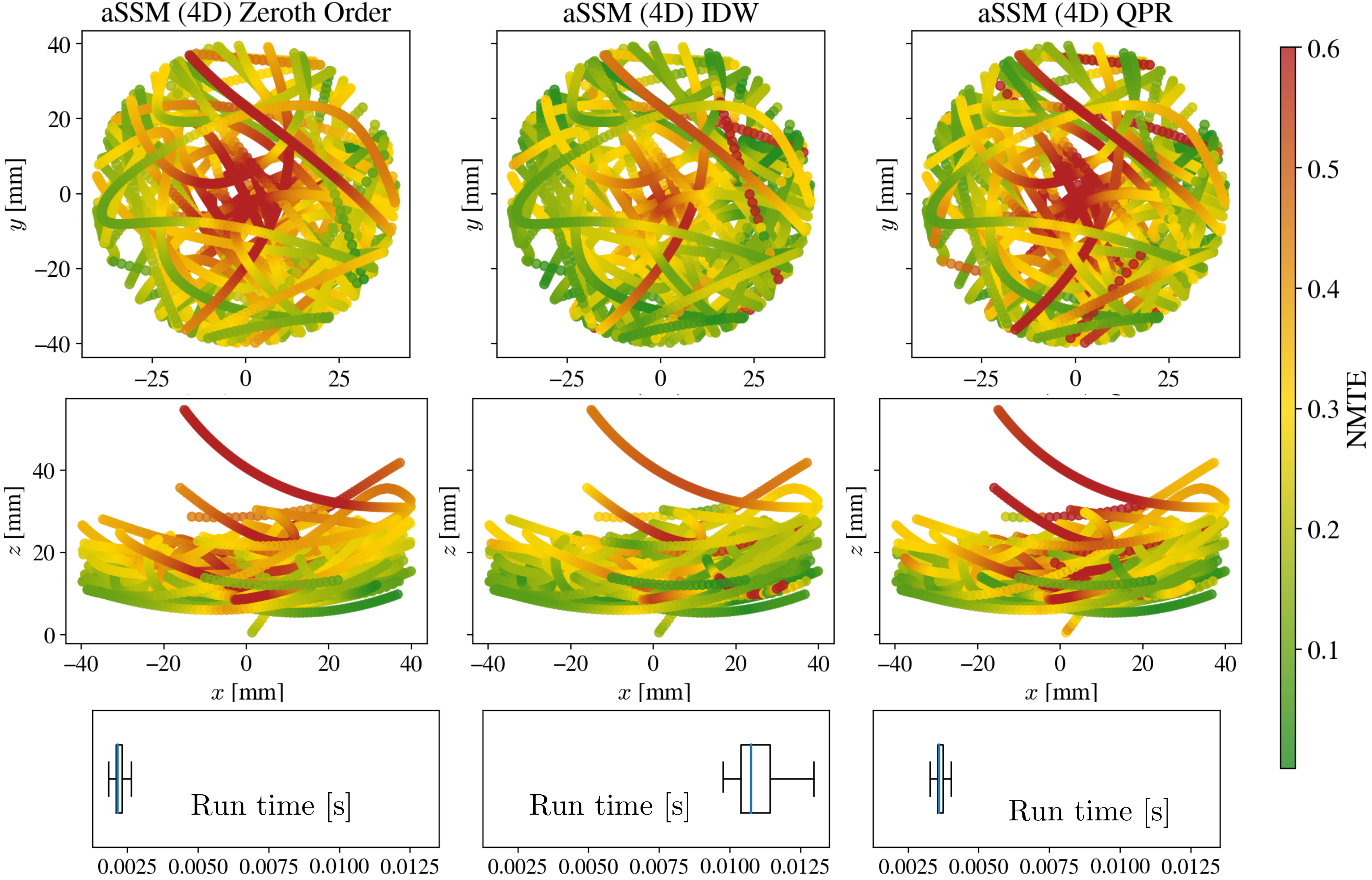}
    \par\end{centering}
    \caption{(Top) $x_{ee}-y_{ee}$ scatter color plots of the random initial conditions for $4$D aSSM-reduced model simulations. Color indicates NMTE values for the open loop prediction. (Middle) $z_{ee}-x_{ee}$ scatter color plots of the random initial conditions. (Bottom) Average run-times for the $4$D aSSM-reduced model simulation.}
    \label{fig:open_loop4D}
    \end{figure*}

\begin{figure*}
    \begin{centering}
    \includegraphics[width=1\textwidth]{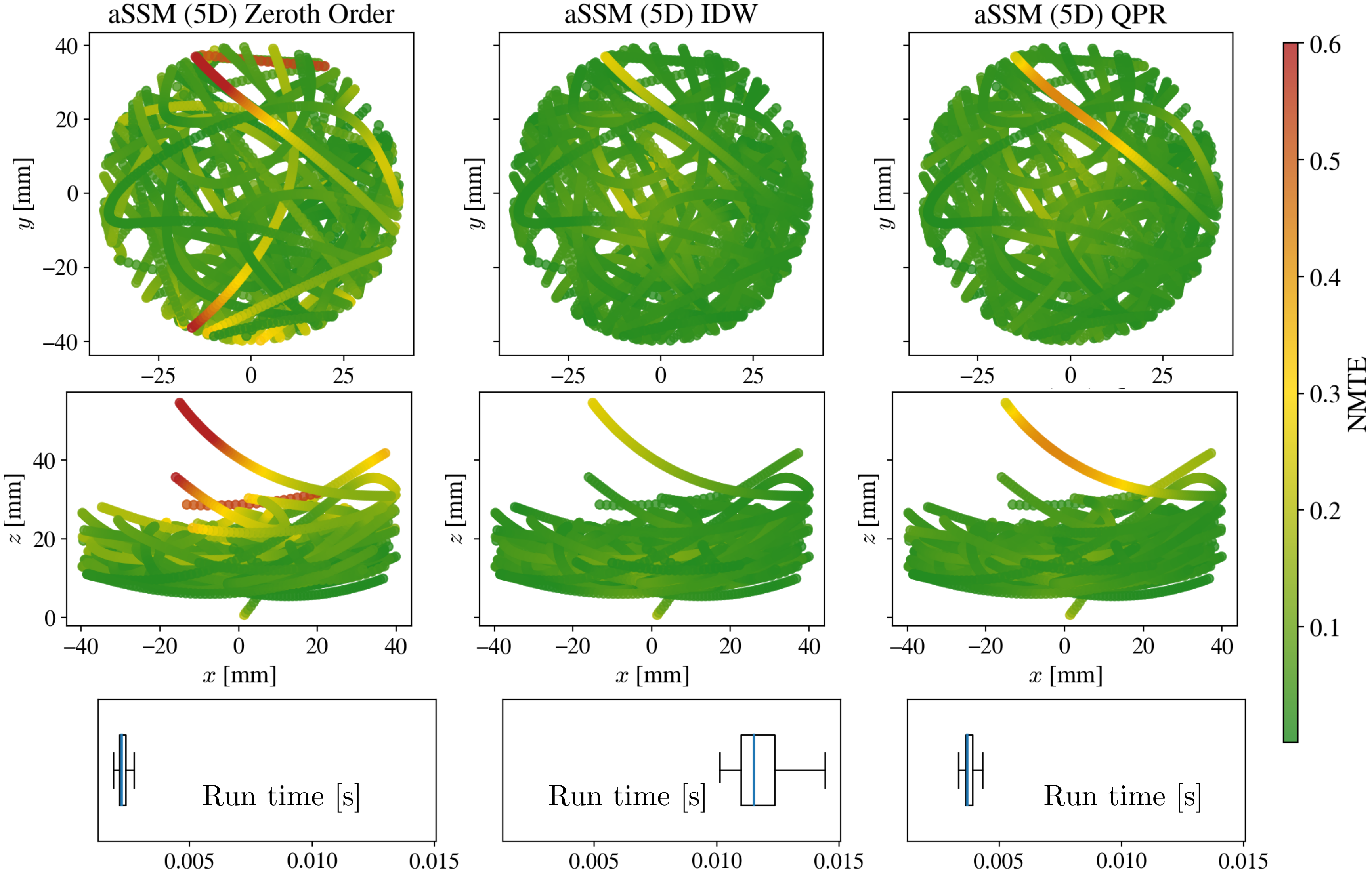}
    \par\end{centering}
    \caption{(Top) $x_{ee}-y_{ee}$ scatter color plots of the random initial conditions for $5$D aSSM-reduced model simulations. Color indicates NMTE values for the open loop prediction. (Middle) $z_{ee}-x_{ee}$ scatter color plots of the random initial conditions. (Bottom) Average run-times for the $5$D aSSM-reduced model simulation. }
    \label{fig:open_loop5D}
    \end{figure*}

Figure \ref{fig:open_loop4D} shows NMTE color plots of the performance of the $4$D models. Notice that these models perform poorly for initial conditions having a non-zero $z_{ee}$ value. This is expected as these models do not have predictive power that models the overdamped behavior of the $z_{ee}$ direction. In Fig. \ref{fig:open_loop5D}, we have the results for the $5$D aSSM-reduced models. In comparison to the $4$D case, we see nearly complete green patches signaling low NMTE values for nearly all the initial conditions. This informs us that the our $5$D aSSM-reduced models have significantly greater generalizability compared to the $4$D aSSM-reduced models. At the same time, the plots signal that $4$D aSSM-reduced model is an optimal choice for closed-loop control tasks restricted to the $x_{ee}-y_{ee}$ plane. 

When comparing the MIDW interpolation and QPR methods, we observe that MIDW offers slightly better performance than QPR for the $5\mathrm{D}$ aSSM-reduced model. However, this comes with longer mean run time for MIDW due to complexity in the interpolation procedure. The QPR method has lower mean run time which aids in achieving closed-loop performance in real-time, and it also comes with good prediction accuracy on unseen inputs. Based on these findings, we choose the $5$D aSSM-reduced model approximated using QPR in our MPC schemes for closed loop control. 

The zeroth-order aSSM-reduced model's performance also meets our expectation. Indeed, a small region close to the origin has green patches in Fig.\ref{fig:open_loop5D} but as we move further away from the origin the errors start to grow. 

For the above results, the average instantaneous speed of the trunk's tip during the full control response is approximately $\approx 125 \text{ [mm/s]}$. This gives the non-dimensional slowness measure $r_s \approx 1.7$. We show similar results when the average instantaneous speed is about $265 \text{ [mm/s]} $ with an $r_s \approx 3.7$ value for the $5$D aSSM-reduced models in Appendix \ref{app:D}. 

\section{Closed-loop control results for soft trunk}
\label{sec:cl_trunk}
To test our aSSM-reduced models in closed-loop control, we use various target tracks $\boldsymbol{\Gamma}(t)$. We set the planning horizon to have a span of $0.02 \text{  [s]}$ when we have no workspace constraints. When we have constraints, the planning horizon is increased to $0.05 \text{  [s]}$. We set the workspace cost matrix $\mathbf{Q}_z = \mathbf{1}_{3 \times 3}$ and the actuation space cost matrix $\mathbf{R}_u = 0.001 \cdot \mathbf{1}_{8 \times 8}$. These choices penalize all inputs and workspace coordinates equally. For $3$D closed-loop control tasks, the $5$D aSSM-reduced model is compared with its zeroth- and first-order approximations. In the case of a target in the $x_{ee}-y_{ee}$ plane, we also add comparisons using the $4$D aSSM-reduced model. We further add comparisons with linear modeling methods: trajectory piecewise linearization (TPWL) by \citet{tonkens21} and Koopman static pregain by \citet{haggerty23}. We discuss implementation details of these linear methods in the Appendix \ref{app:A.TPWL}-\ref{app:A.Koop}, and data collection and training time comparisons with aSSM-reduction in Appendix \ref{app:A.training_time}.

We evaluate closed-loop performance by calculating the integrated square error (ISE) across the target horizon $\mathcal{T} = [t_i,t_f]$, defined as 
\begin{equation}
    \text{ISE} = \int_{t_i}^{t_f} \|\mathbf{z}_{pred}(s) - \boldsymbol{\Gamma}(s) \|^2 ds.
\end{equation}
All our closed loop control tasks were run on an AMD Ryzen 7 1800X eight-core processor @ 3.6 GHz with OS Ubuntu 22.04.5 LTS.

\subsection{Figure-8 track}

\begin{figure*}
    \begin{centering}
    \includegraphics[width=1\textwidth]{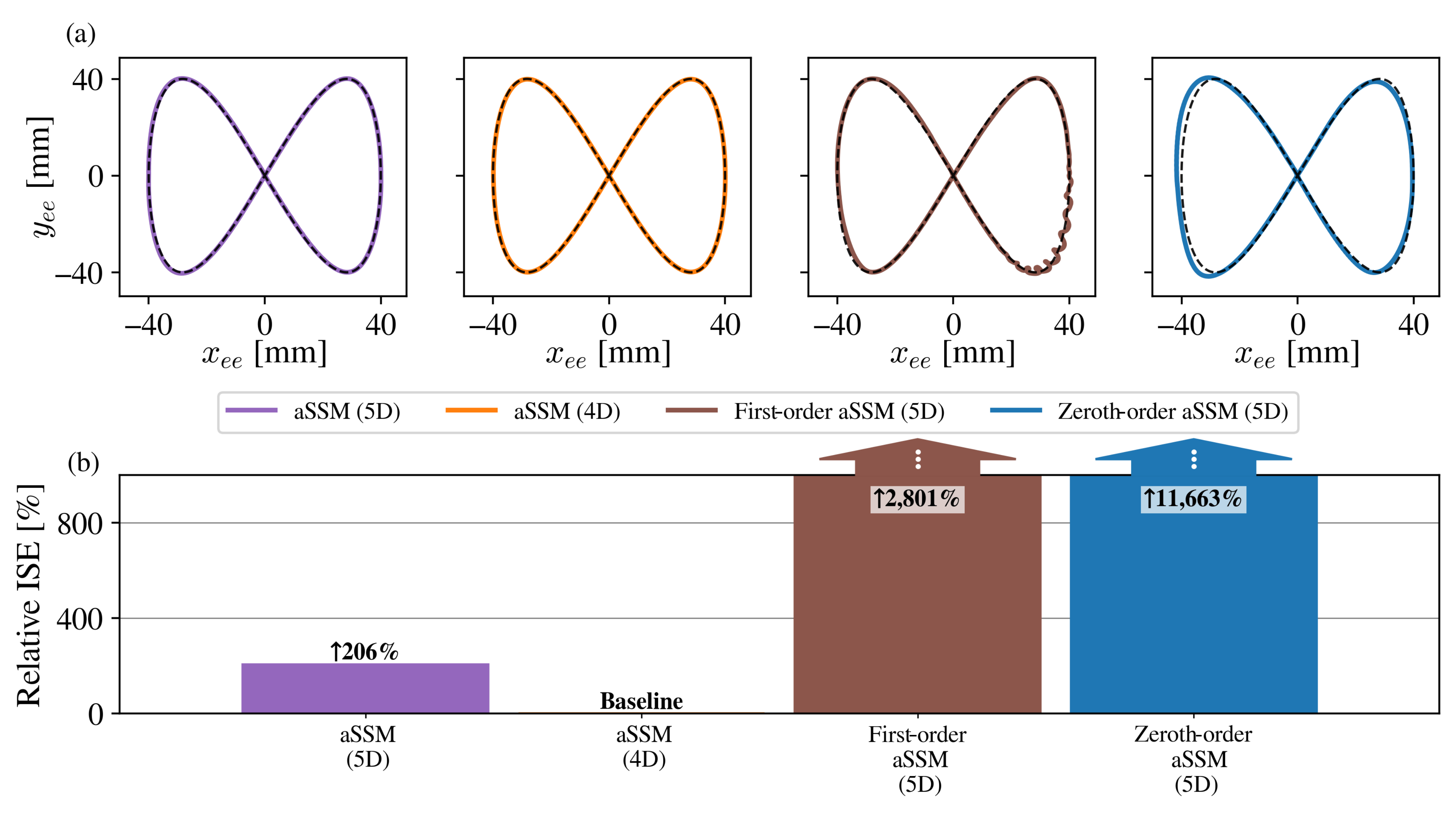}
    \par\end{centering}
    \caption{ (a) Closed-loop results for the figure-$8$ track plotted in the $x_{ee}-y_{ee}$ plane. (b) Bar plots of relative ISE, with the 4D aSSM-reduced model as the baseline.}
    \label{fig:figure8}
    \end{figure*}

Our first target $\boldsymbol{\Gamma}(t)$ is a figure-8 trajectory in the $x_{ee}-y_{ee}$ plane. This figure-$8$ target has a 10-second period and a maximum amplitude of $40 \text{ [mm]}$. The average instantaneous speed achieved along the target is $37.74 \text{ [mm/s$^2$]}$ with slowness measure $r_s \approx 0.5$. The target trajectory starts at the origin and traverses the figure-$8$ in the clockwise direction. Figure \ref{fig:figure8} shows results from four different models used in a MPC control scheme: $5$D aSSM-reduced model, $4$D aSSM-reduced model,  $5$D zeroth-order aSSM-reduced model and $5$D first-order aSSM-reduced model. We find that the $4$D aSSM-reduced model has the lowest ISE. Using this as a baseline, we evaluate relative ISE for the rest of the models (see Fig. \ref{fig:figure8}b). These results are consistent with the open-loop analysis presented in Section \ref{sec:open_loop} in terms of the performance of the various models.

Despite the relative ISE of $206 \%$, the $5$D aSSM-reduced model also achieves similar control performance along the target as the $4$D case. We also observe that the $5$D first-order aSSM-reduced model does better than the $5$D zeroth-order aSSM-reduced model. The $5$D zeroth-order aSSM-reduced model traces the target near the origin with good accuracy, but as it moves radially outward, it starts to overestimate the target. 

    On the other hand, the first-order aSSM-reduced model tracks the majority of the figure-8 precisely but begins to jitter on the bottom curve on the right lobe causing the relative ISE to shoot up to $2,801 \%$. We attribute this phenomenon to model error arising from the control calibration performed in a region in the workspace where the first-order aSSM approximation is no longer valid. 

    We conclude aSSM-reduced models are generalizable, as none of these models were trained on inputs that resembled the figure-$8$. We also find that our aSSM-reduced models to extrapolate well in regions of the $x_{ee}-y_{ee}$ plane, $\sqrt{x_{ee}^2 +y_{ee}^2}>20 \text{ [mm]}$, where they have no prior information of the static steady states and SSM computations. In Appendix \ref{app:C}, we present comparisons of the aSSM-reduced models with linear methods. The results from  Appendix \ref{app:C} indicate that the trunk is only mildly nonlinear in $x_{ee}-y_{ee}$. Indeed, both linear methods are able to trace out the figure-$8$ but lack precision compared to the aSSM-reduced models. 

\begin{figure*}
    \begin{centering}
    \includegraphics[width=0.6\textwidth]{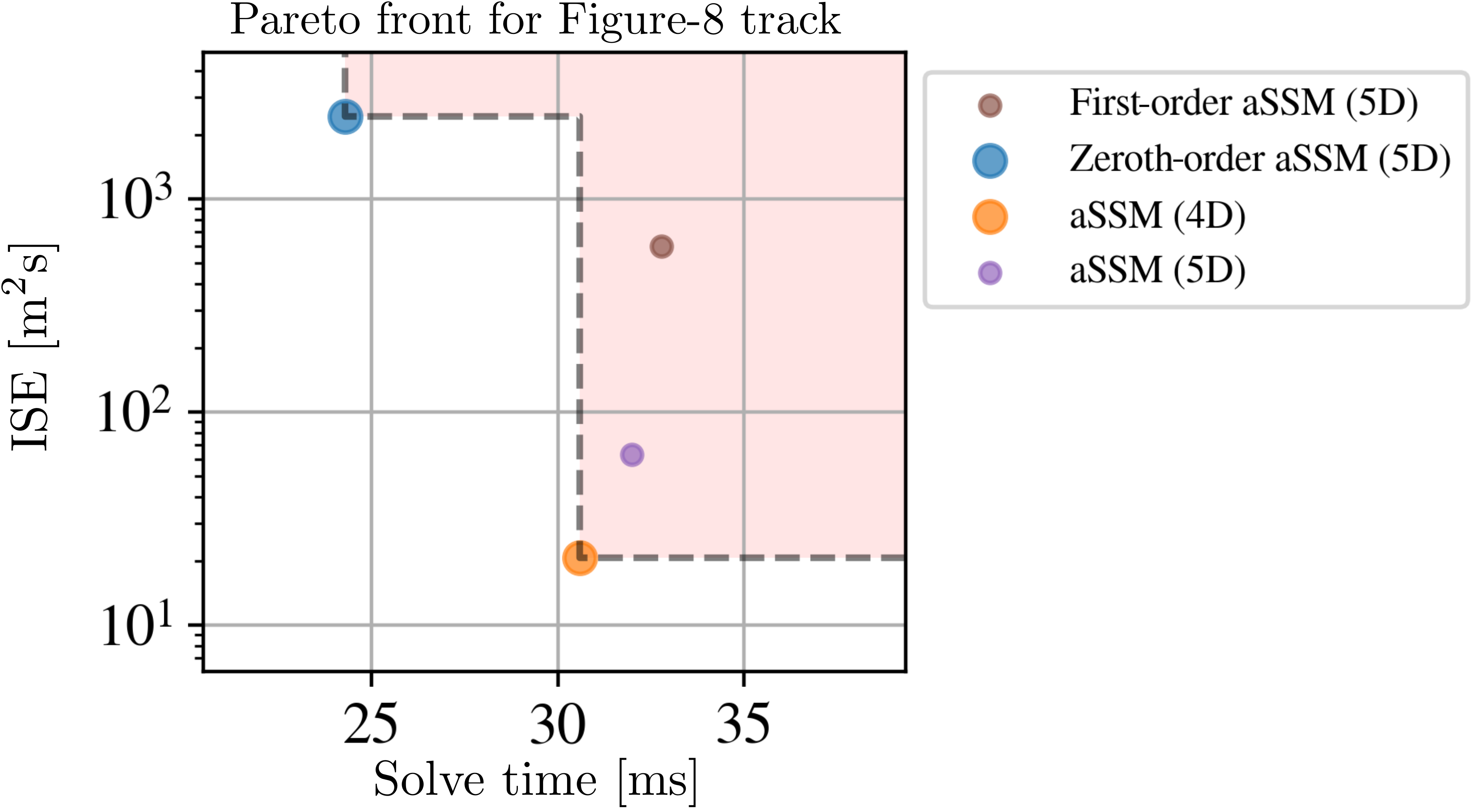}
    \par\end{centering}
    \caption{Pareto front for the closed loop performance on the  figure $8$. The horizontal axis is the average solve times for the MPC and the vertical axis represents the ISE.}
    \label{fig:figure8_pareto}
    \end{figure*}
 
We also investigate the trade-off between model accuracy and finite-horizon optimization solve times. Figure \ref{fig:figure8_pareto} plots the Pareto front for the figure-$8$ trajectory. The Pareto front here, qualitatively depicts the trade-off between control accuracy and control solve times. The models appearing on the Pareto front are said to be Pareto optimal, as their performance in terms of accuracy and solve time is not strictly dominated by any other model. We find the $4$D aSSM-reduced model and the $5$D zeroth-order aSSM-reduced model to lie on the Pareto front. The $4$D aSSM-reduced model provides us precision control with average solve time of about $31 \text{ [ms]}$ whereas the $5$D zeroth-order aSSM-reduced model in comparison has a $10 \text{ [ms]}$ lower solve time but comes at a loss of control accuracy as it has the largest ISE. Depending upon the use case for the robot, the practitioner can make a choice between the two Pareto efficient models for planar trajectory tracking. 

\subsection{Spherical random track}

We now consider a $3$D target track in the workspace, randomly generated using a Perlin noise scheme (see \citet{perlin85}). The track starts at the origin and varies smoothly in all workspace directions. The target $\boldsymbol{\Gamma} = \{\alpha(t),\beta(t), \sqrt{R^2 - s^2(\alpha(t)^2 - \beta(t)^2)}  \}$, where $\alpha(t)$ and $\beta(t)$ are unit normalized scalar signals, generated by such a scheme. We set the maximal radial span of the target at $s=40 \text{ [mm]}$ and the maximal height at $h = 20 \text{ [mm]} $, which gives $R=\frac{h^2+s^2}{2h}$.  

\begin{figure*}
    \begin{centering}
    \includegraphics[width=1\textwidth]{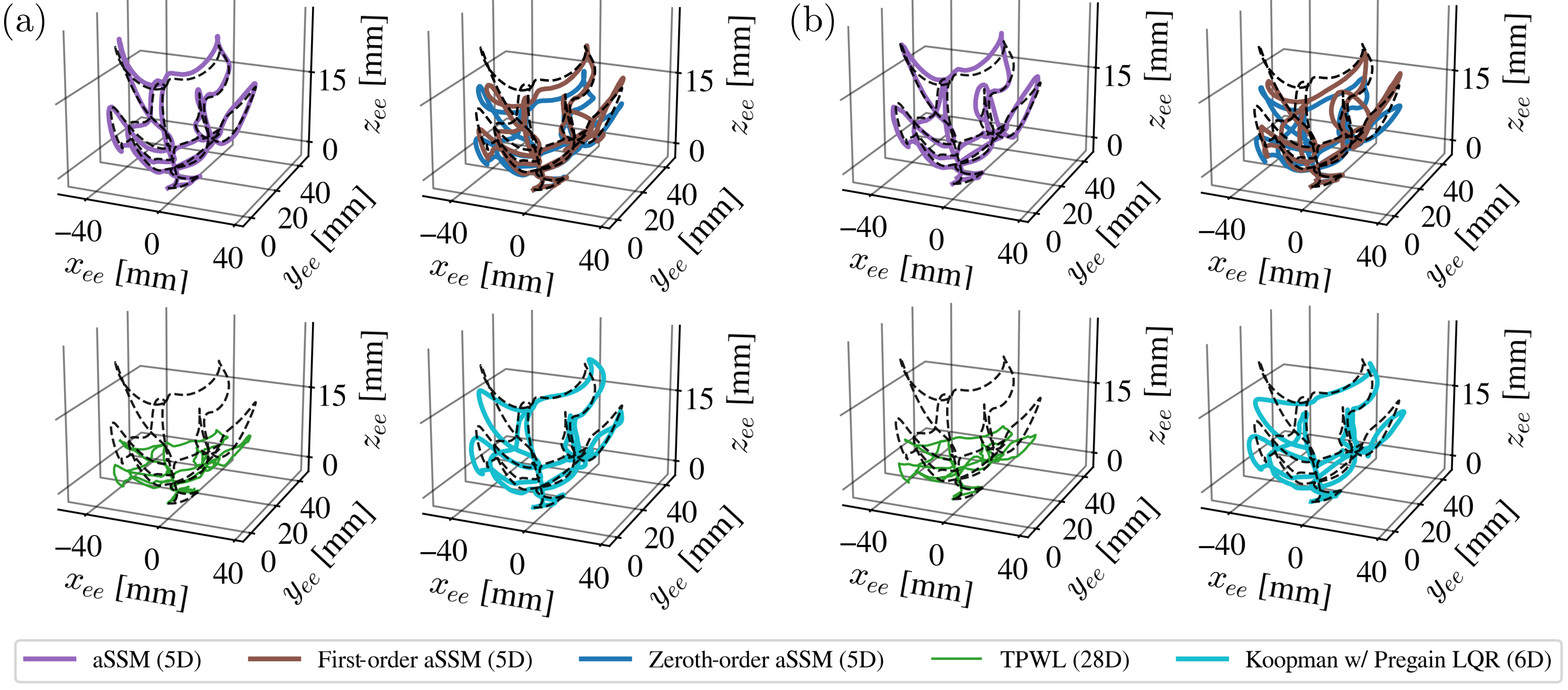}
    \par\end{centering}
    \caption{(a) Closed-loop results for the slow random spherical target plotted in the trunk robot's workspace. (b) Closed-loop results for the fast random spherical target plotted in the trunk robot's workspace.}
    \label{fig:random}
    \end{figure*}

To test our aSSM-reduced MPC schemes for speed variations along the given track, we consider a mean instantaneous speed of $50.16 \text{ [mm/s$^2$]}$ with slowness measure $r_s \approx 0.7$ and faster variation at a mean instantaneous speed of $100.32 \text{ [mm/s$^2$]}$ with $r_s \approx 1.4$. In Fig. \ref{fig:random}a, we show closed-loop performance for the slower varying case using the aSSM-reduced models as well as linear models. We find the aSSM-reduced model and its approximations to perform significantly better than the linear methods. TPWL method performs visibly poorly and is unable to reach the maximal height. The Koopman static pregain method is able to roughly trace out the random target's behavior but fails to follow it accurately. 

\begin{figure*}
        \begin{centering}
        \includegraphics[scale=0.25]{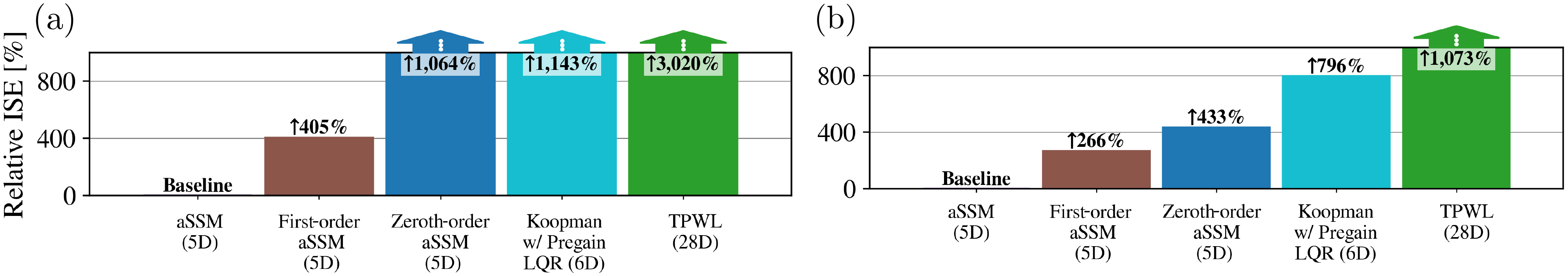}
        \par\end{centering}
        \caption{(a) Bar plots of relative ISE for slower track, with the 5D aSSM-reduced model as the baseline. (b) Bar plots of relative ISE for fast case, with the 5D aSSM-reduced model as the baseline.}
        \label{fig:random_ISE}
        \end{figure*}

Figure \ref{fig:random_ISE}a confirms our visual assessment quantitatively with the relative ISE bar plots indicating the same trend. Results for the faster version of the same track are presented in Fig. \ref{fig:random}b. Again, a visual assessment suggests that the aSSM-reduced model has the best performance, confirming the robustness of our aSSM-reduced model. We again verify this assessment quantitatively from the ISE bar plots for the various models in Fig. \ref{fig:random_ISE}b. As seen from the plot, Koopman static pregain does significantly worse here than in the slow case. This is because the method assumes the inputs to be quasi-statically varying and the mapping between the static inputs and steady states to be linear, neither of which holds for the trunk. Generally, the key reason for the failure of the linear methods is their non-robustness to unseen control inputs. 

\begin{figure*}
    \begin{centering}
    \includegraphics[width=0.8\textwidth]{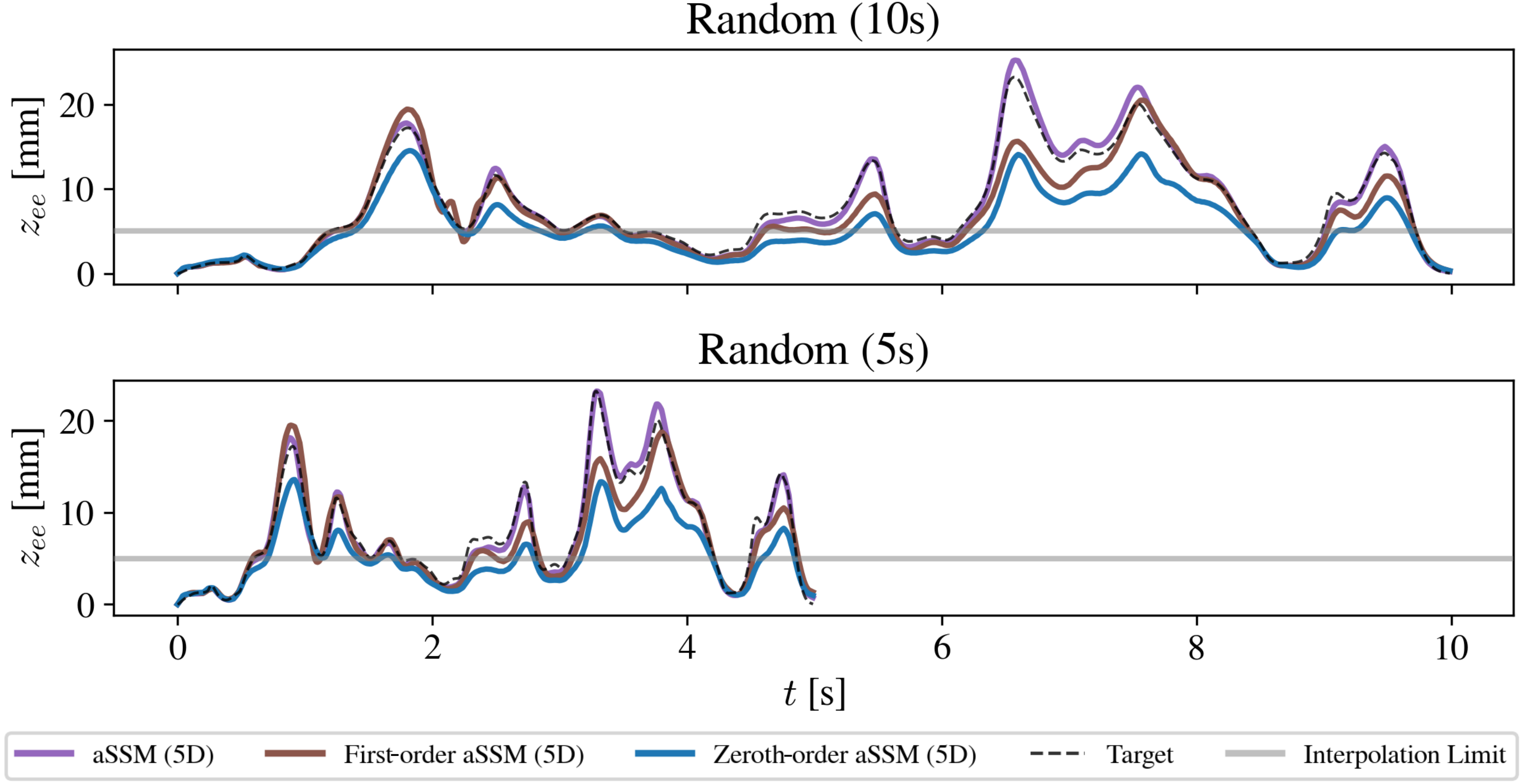}
    \par\end{centering}
    \caption{(Top) $z_{ee}$ closed loop tracking for the slow  random spherical target. (Bottom) $z_{ee}$ closed loop tracking for the fast random spherical target. The gray line represents the interpolation limit, beyond which the aSSM-reduced models start extrapolating.}
    \label{fig:random_z}
    \end{figure*}

We compare more closely the aSSM-reduced models and their approximations by plotting the performance in the $z_{ee}$ direction in Fig. \ref{fig:random_z}. The results confirm the local domain of validity of the $5$D zeroth-order aSSM-reduced model showing degraded performance along the height of the workspace for both speeds of the target track. The first-order aSSM-reduced model shows a slight improvement in performance but fails to reach the heights of the target, even though it tracks the variations smoothly. We again note that $z_{ee}$ values explored by the target are much further away from the training regime of the static SSM models.

\begin{figure*}
    \begin{centering}
    \includegraphics[width=0.6\textwidth]{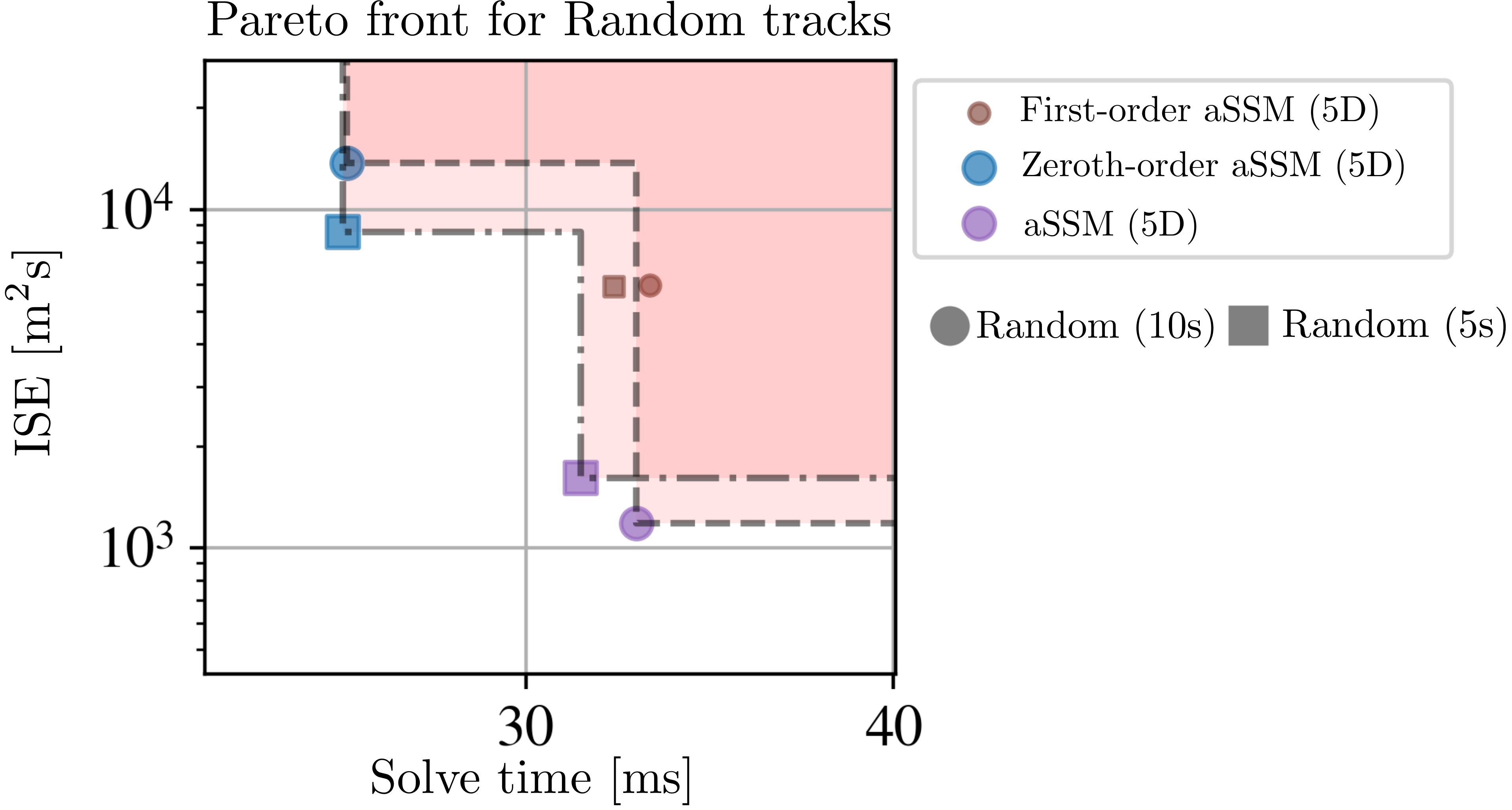}
    \par\end{centering}
    \caption{Pareto fronts for the closed loop performance on the  spherical random target for two different speeds. The horizontal axis is the average solve time for the MPC and the vertical axis represents the ISE.}
    \label{fig:random__pareto}
    \end{figure*}

In Fig. \ref{fig:random__pareto}, we plot the Pareto fronts for the slow and fast random tracks. Similarly to the figure-8 results, we find the zeroth-order aSSM-reduced and the aSSM-reduced model to be Pareto efficient. We also notice the average solve times of the aSSM-reduced model decreases with increasing target speed but the ISE increases. Note that the faster target generates a lower ISE under the zeroth-order aSSM approximation. This is because this approximation is robust to uniformly bounded variations in the control inputs. The first-order aSSM approximation is suboptimal because for a slight improvement in ISE, the average solve time is still larger compared to the aSSM-reduced model.  

\subsection{$3$D Pacman track}

In the next example, we increase the difficulty of the control task with a $3$D Pacman track that is mostly outside the training regime of all the models and also has non-smooth variations. The radius of the Pacman track is $30 \text{ [mm]}$ and its height is $25 \text{ [mm]}$. The mean instantaneous speed along the track is $22.9 \text{ [mm/s]}$, which gives the slowness measure $r_s \approx 0.3$.  

\begin{figure*}
    \begin{centering}
    \includegraphics[width=1\textwidth]{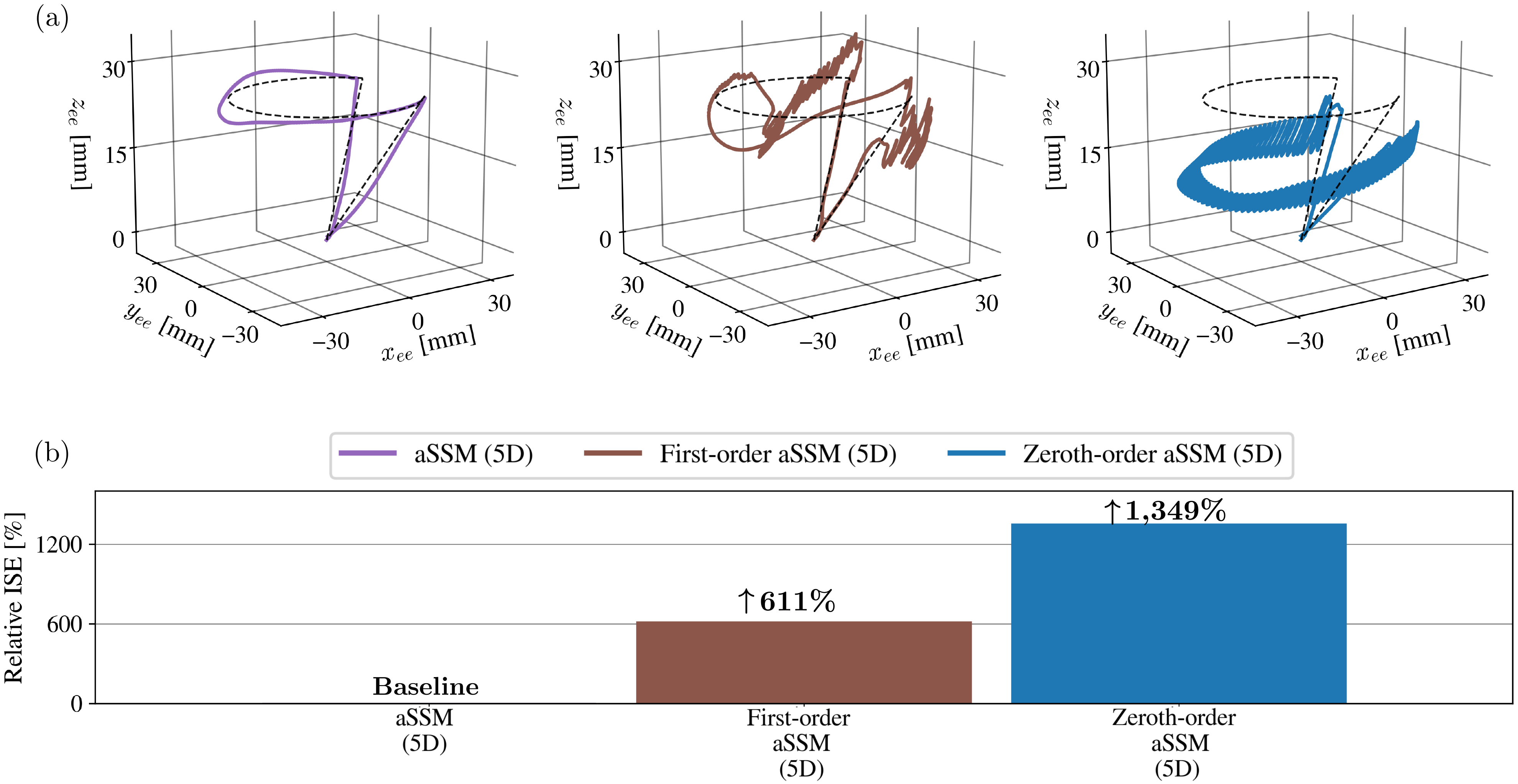}
    \par\end{centering}
    \caption{(a) Closed-loop results for the $3$D Pac-man plotted in the trunk robot's workspace. (b) Bar plots of relative ISE, with the 5D aSSM-reduced model as the baseline.}
    \label{fig:pacman_section}
    \end{figure*}

Figure \ref{fig:pacman_section}a shows the closed-loop performance of the aSSM-reduced models. A visual assessment shows that the aSSM-reduced model performs the best, as expected. The other two approximations struggle to reach the maximum height, although the first-order approximation performs better than the zeroth-order. The first-order aSSM-reduced model is at least able to track the steep slope, but ultimately fails to stay on the circular track. The first-order aSSM approximation results illustrate that even if one knows the existence and approximate location of an aSSM, one generally cannot approximate it as a fiber bundle of constant linear fibers on larger domains because aSSMs tend to be fiber bundles of non-constant nonlinear fibers.
 
\begin{figure*}
        \begin{centering}
        \includegraphics[width=0.8\textwidth]{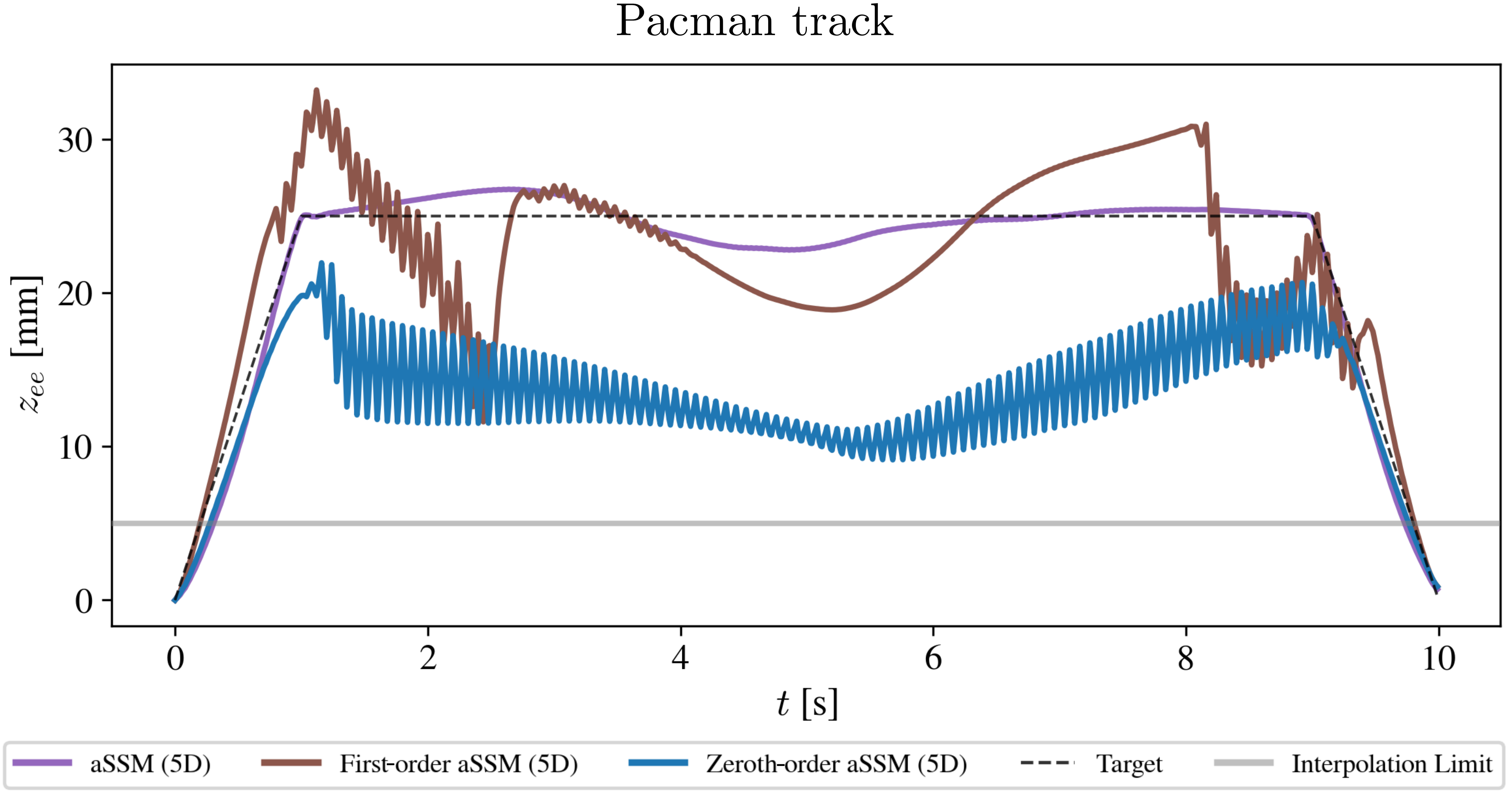}
        \par\end{centering}
        \caption{$z_{ee}$ closed loop tracking for the $3$D Pacman. The gray line represents the interpolation limit, beyond which the aSSM-reduced models start extrapolating.}
        \label{fig:pacman_z}
        \end{figure*}

\begin{figure}
            \begin{centering}
            \includegraphics[width=0.5\textwidth]{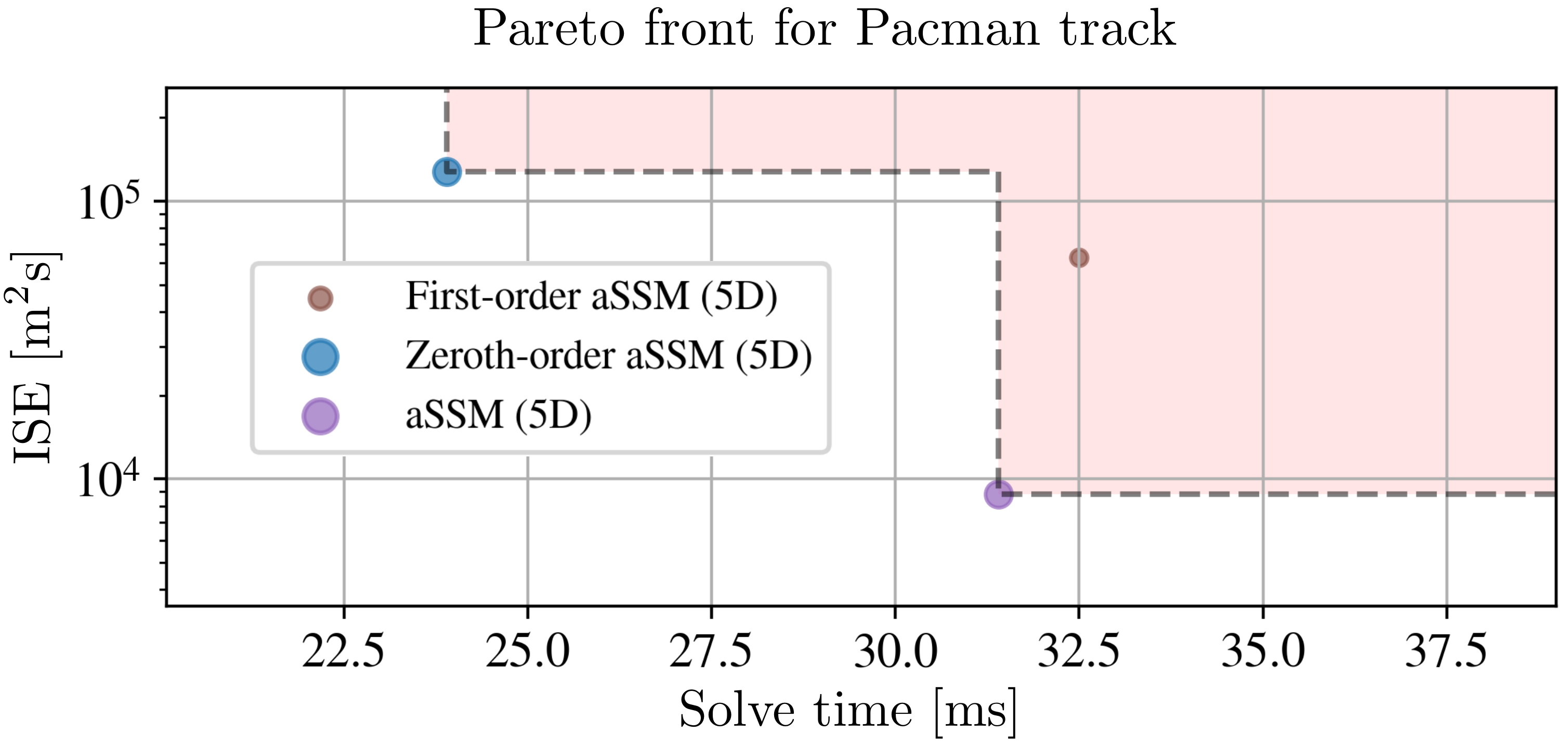}
            \par\end{centering}
            \caption{Pareto front for the closed loop performance on the $3$D Pac-man. The horizontal axis is the average solve time for the MPC and the vertical axis represents the ISE.}
            \label{fig:pacman_z_pareto}
            \end{figure}

 In Fig. \ref{fig:pacman_section}b, the relative ISE plots confirm our visual assessment. For completeness, in Fig. \ref{fig:pacman_z}, we also plot the $z_{ee}$ tracking direction, which yields similar conclusions. The Pareto front is again similar as in the previous examples (see Fig. \ref{fig:pacman_z_pareto}). The aSSM-reduced model has the lowest ISE but has a slightly increased mean solve time of $0.32 \text{ [s]}$ compared to the zeroth-order aSSM at $0.24 \text{ [s]}$. 

\subsection{$3$D Pacman track with spherical constraints}
As a final example, we consider a similarly shaped Pacman track with the same radius but at a height of $15 \text{ [mm]}$ and with five randomly sized and placed spherical constraints in the workspace. The average speed clocked along this track is $21.8 \text{ [mm/s$^2$]}$ with slowness measure $r_s \approx 0.3$. The goal is to control the trunk as closely as possible along the track but at the same time avoid all the constraints. 

\begin{figure*}
    \begin{centering}
    \includegraphics[width=1\textwidth]{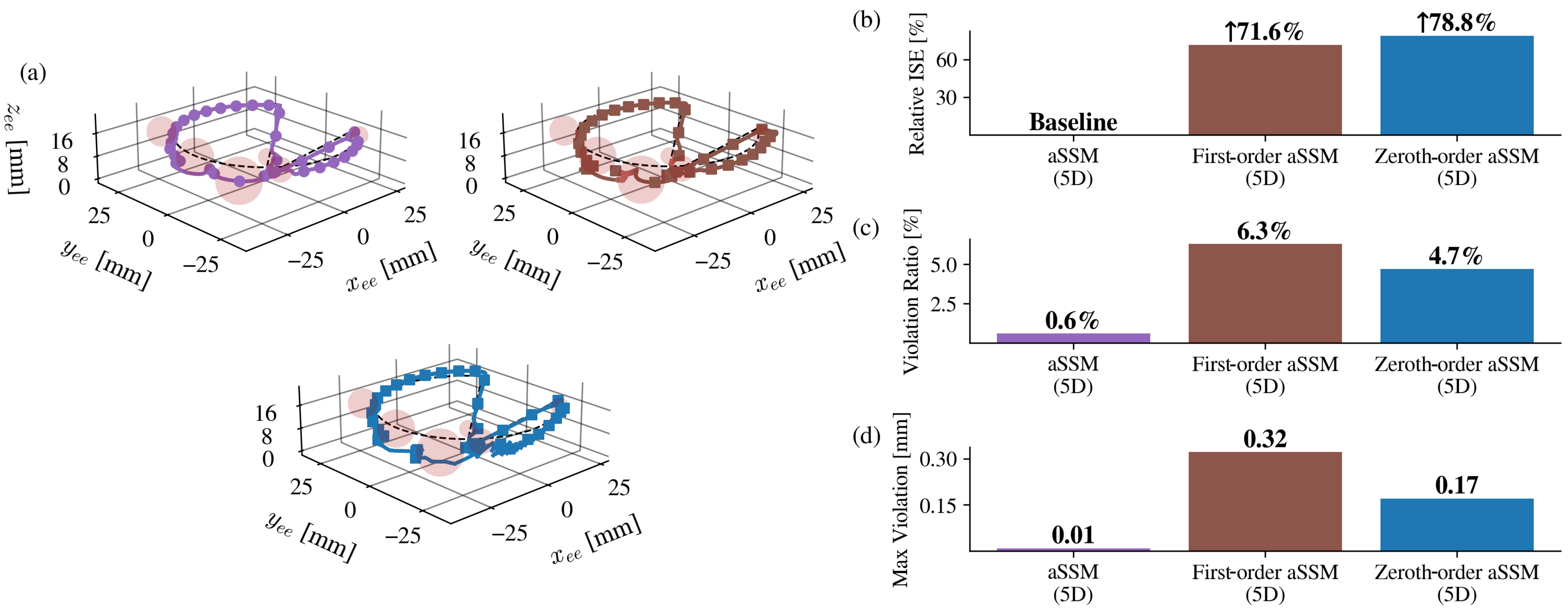}
    \par\end{centering}
    \caption{(a) Closed-loop results for a $3$D pac-man with $3$D spherical constraints in the workspace. (b) Bar plots of relative ISE, with the 5D aSSM-reduced model as the baseline. (c) Bar plots of violation ratio. (d) Bar plots of the maximum violation achieved along the target. }
    \label{fig:pacman_constraints}
    \end{figure*}

In Fig. \ref{fig:pacman_constraints}a, we plot the closed loop predictions for the aSSM-reduced model and its approximations. We also plot the spherical constraints as red transparent spheres in the workspace.  We introduce two error metrics that will aid in our quantification. The first is \textit{violation ratio}, which is the number of times the robot entered the constraint divided by the length of the robot's trajectory. The second is \textit{maximum violation}, which is the maximum distance by which the robot violates the constraints. In  Figs. \ref{fig:pacman_constraints}c-d, we plot these metrics for the aSSM-reduced model, the first-order aSSM-reduced model and the zeroth-order aSSM-reduced model. The radii of the spherical constraints range from $1 \text{ [mm]}$ to $5 \text{ [mm]}$. By maximal violation measure, all the aSSM-reduced models perform consistently; as the maximum violation distance is $0.32 \text{ [mm]}$ across models. We remark that these are indeed hard constraints, so in principle, when they are violated, the program algorithm should exit. However, to test the models across the target horizon, we re-initialize the failed models at an initial condition just outside the constraint, closest to the point of failure.

Overall, the aSSM-reduced model has the smallest violation ratio, the smallest maximum violation distance of $0.01 \text{ [mm]}$ and also the lowest ISE. This effectiveness of the aSSM-reduced model for constrainted workspace configurations illustrates that the mathematical assumptions going into the theory of aSSM-reduction are realistic, at the very least for soft trunk robots.

\section{Controlling soft elastic arms}
We focus here on using our methodology to identify a data-driven aSSM-reduced model for a soft elastic arm modeled as a slender rod obeying the Cosserat rod theory (see \citet{cosserat1909theorie, gazzola18}). The arm's dynamics is time-integrated using the numerical procedure outlined in \citet{gazzola18}. Specifically, we aim to find an accurate aSSM-reduced model for the elastic arm geometry appearing in \citet{naughton21} for two cases: a short and long elastic arm. The short and long arms' undeformed lengths are $1 \text{ [m]}$ and $3 \text{ [m]}$, and both have a radius of $5 \text{ [cm]}$. The short arm has a Young's modulus $10 \text{ [MPa]}$ and material dissipation coefficient per unit length $7 \text{ [kg/(ms)]}$. The long arm has a Young's modulus $20 \text{ [MPa]}$ and a material dissipation coefficient per unit length $5 \text{ [kg/(ms)]}$. Both arms have a Poisson's ratio $0.5$ and material density $1000 \text{ [kg/m$^3$]}$.

On one end, the elastic arm is fixed, while the other end is free to move. The arm is actuated by supplying continuous torques along the arm's length that act normally and binormally to its centerline. The torques act at three equidistant points along the short arm's length and along four equidistant points along the long arm's length, leading to a $n_u = 6$ dimensional control input space $\mathbf{u} \in \mathbb{R}^6$ for the short arm and $n_u = 8$ dimensional control input space $\mathbf{u} \in \mathbb{R}^8$ for the long arm. For our simulations, we set $15 \text{ [Nm]}$ as the maximum torque magnitude attainable for the short arm and $3 \text{ [Nm]}$ for the long arm, ensuring that assumption (\ref{eq:critical_manifold}) is satisfied. See Appendix \ref{app:physical_models} for details on the governing equations and the actuating mechanism of the soft elastic arm geometry. 

\subsection{Training details of aSSM-reduced model}
Our focus is to control the arm's free end (end effector), hence our workspace is defined by the end effector coordinates $\mathbf{z} = \{x_{ee}, y_{ee}, z_{ee}\} \in \mathbb{R}^3$. We collect $10$-second long samples of 4 decaying trajectories for each $N_u = 1000$ randomly sampled static arm configurations. We ensure practical limitations on observability by only assuming access to a $3p$-dimensional observable space $y = (\mathbf{z}(t),\mathbf{z}(t+\tau), \dots ,\mathbf{z}(t+(p-1)\tau))^\top \in \mathbb{R}^{3p}$, where $\tau = 0.0168 \text{ [s]}$. 

The undeformed arm is symmetric along the $x_{ee}-y_{ee}$ and also freely stretches along the $z_{ee}$ direction, due to these reasons we expect three dominant slow modes in the decay profiles of the arm. By Takens' delay embedding theorem, we require $3p \geq 15$ to guarantee capturing decay along these three dominant slow modes. Indeed for $p=5$, we observe $d=6$ dimensional static SSM-reduced models with $n_r = 2$ and $n_w = 2$ to accurately capture the decay with an average $\mathrm{NMTE} = 5 \%$ on test data for the short arm and an average $\mathrm{NMTE} = 4.5 \%$ on test data for the long arm . 

We further collect controlled data by adding a 10-second long randomly generated control deviation input, with a maximum torque magnitude of 1.5, to a step input sequence formed by 100 static inputs, each held for 10 seconds (see Section \ref{sec:control_calib} for details). We use this to learn the effective control calibration set. Specifically, we find a $6$D aSSM-reduced model obtained via quadratic polynomial regression (QPR) on the static SSM coefficient set to provide consistent performance on unseen test data. For a 30-second long random input with slowness measure $r_s = 0.75$, our $6$D aSSM-reduced model's prediction has a $9 \%$ NMTE error for the short arm and a $11 \%$ NMTE error for the long arm. See Appendix \ref{app:training_elastic_arm} for prediction results in the robot's workspace. 

\subsection{Short elastic arm: Closed-loop results for a $3$D Trifolium track}

We design a spherical target track $\boldsymbol{\Gamma} = \{r(t)\cos(2\pi t/T),r(t)\sin(2\pi t/T), \sqrt{1 - r^2(t)}  \}$, where $r(t) = r_0 (1+\cos(6\pi t/T + \pi))$. This track resembles a three-lobed curve, known as the Trifolium curve. We set the Trifolium curve's parameters $r_0=0.45 \text{ [m]}$ and $T=10 \text{ [s]}$.  The mean instantaneous speed along the track is $56.8 \text{ [cm/s]}$ and the slowness measure is $r_s =0.5$. 

We perform closed-loop control using the reduced MPC scheme eq.(\ref{eq:aSSM_mpc}) for the 6D aSSM-reduced model and its first-order approximation. We also add a comparison with the Koopman static pregain method, following similar guidelines in Appendix \ref{app:A.Koop} using identical training data collected for the static SSM models. The planning horizon for the MPC scheme is set to $0.1 \text{ [s]}$ and the workspace cost matrix $\mathbf{Q}_z = 10 \cdot\mathbf{1}_{3 \times 3}$ for all models. We set $\mathbf{R}_u = 0.8\cdot\mathbf{1}_{6 \times 6} $ for the 6D aSSM-reduced MPC scheme and $\mathbf{R}_u = 1.5\cdot\mathbf{1}_{6 \times 6} $ for its first-order approximation and the Koopman static pregain method. For this example, we observed that different $\mathbf{R}_u$ values led to improved performance for the first-order aSSM-reduced approximation and the Koopman static pregain method; hence, the $\mathbf{R}_u$ settings were adjusted to ensure a fair comparison.

\begin{figure*}
    \centering
    \subfloat[]{\includegraphics[width=0.4\textwidth]{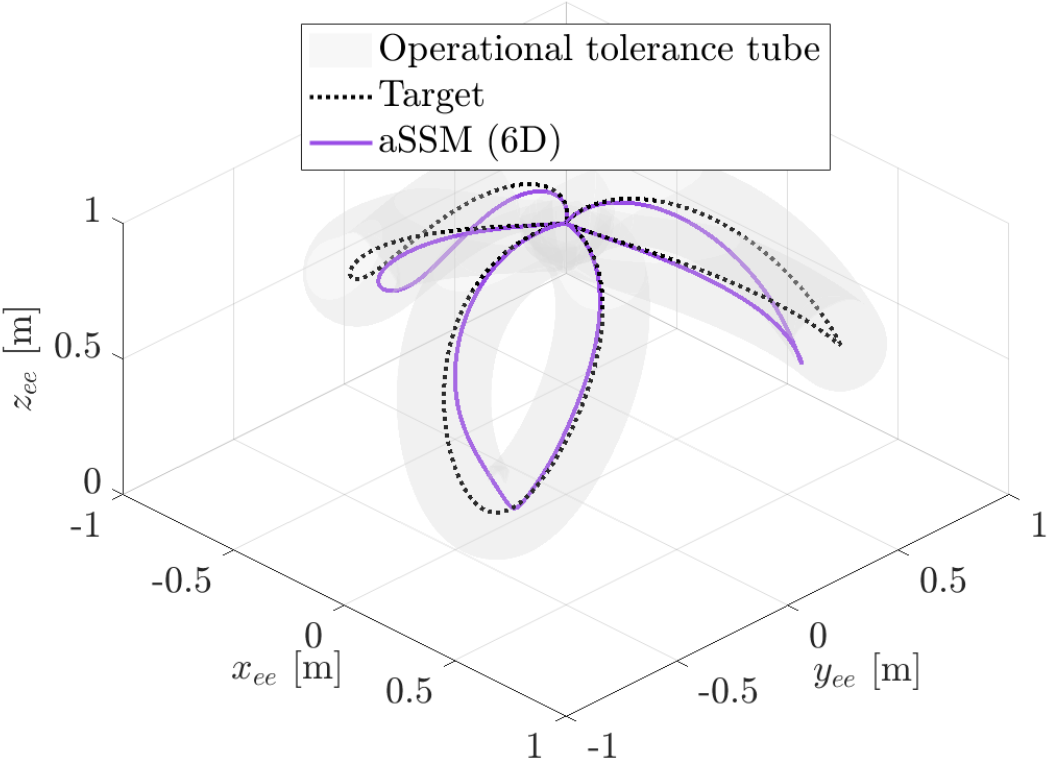}}
    \hspace{0.05\textwidth}
    \subfloat[]{\includegraphics[width=0.4\textwidth]{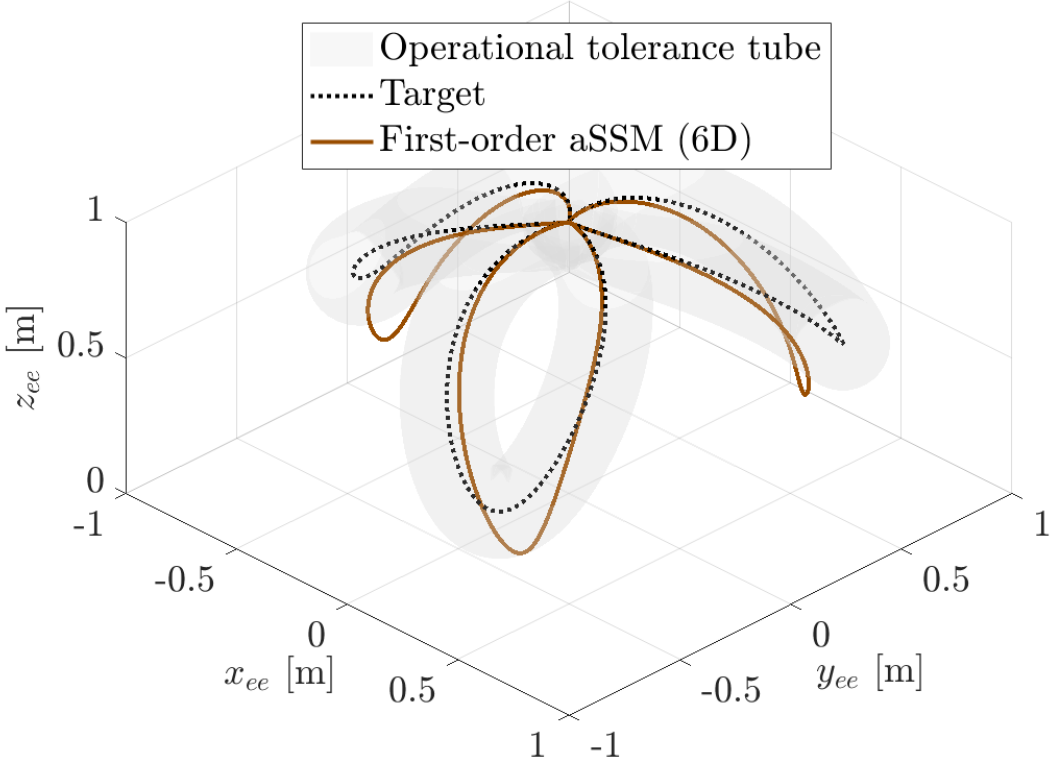}}
    \hspace{0.05\textwidth}
    \subfloat[]{\includegraphics[width=0.4\textwidth]{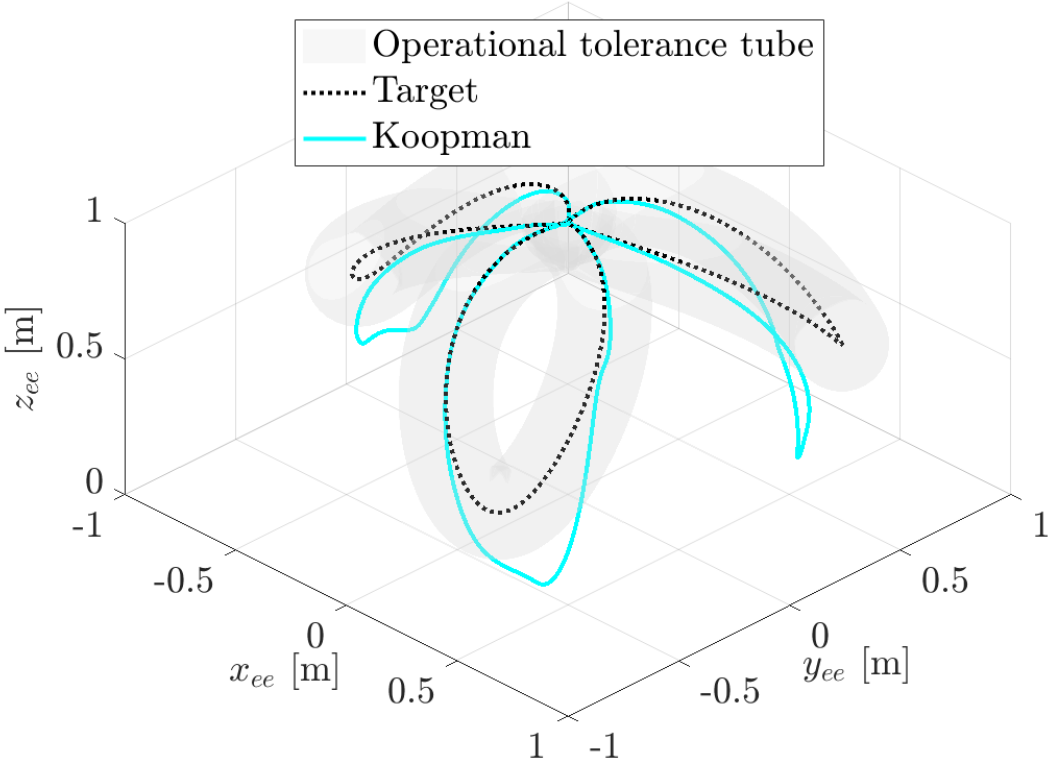}}
    \caption{Closed-loop prediction plots in the short elastic arm's workspace for (a) 6D aSSM-reduced model, (b) 6D first-order aSSM-reduced model and (c) 6D Koopman static pregain method. Target track is plotted in a black dotted line and the gray shaded tube represents the allowed operational tolerance of the soft arm's end effector.}
    \label{fig:cl_arm}
\end{figure*}

In Fig.\ref{fig:cl_arm}a-c, we plot the closed loop predictions in the robot's workspace for all the model-based MPC schemes. It is impossible for the soft arm to exactly track the spherical Trifolium curve as the soft arm is underactuated (see Appendix \ref{app:training_elastic_arm}). Hence, all plots show a gray-shaded cylindrical tube around the target, indicating the allowed operational tolerance of the soft arm's end effector during closed-loop tracking. We set the radius of the operational tolerance tube to be $2\%$ of the target's length, which amounts to a tube of radius $15 \text{ [cm]}$. We clearly observe that only the 6D aSSM-reduced model falls within the operational tolerance tube. Both the 6D first-order aSSM-reduced model and the Koopman static pregain method produce closed-loop inputs that cause the arm to deviate from the target trajectory as it approaches the maximal points.

\begin{figure*}
    \centering
    \subfloat[]{\includegraphics[width=0.7\textwidth]{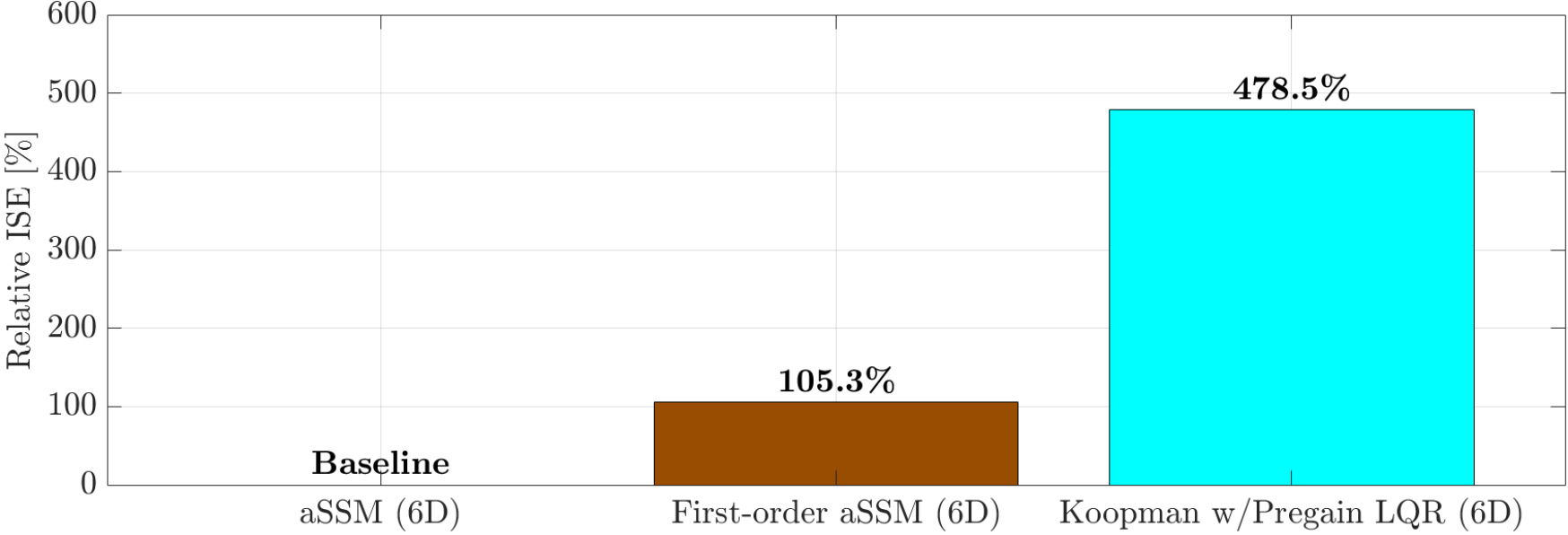}} 
    \subfloat[]{\includegraphics[width=0.3\textwidth]{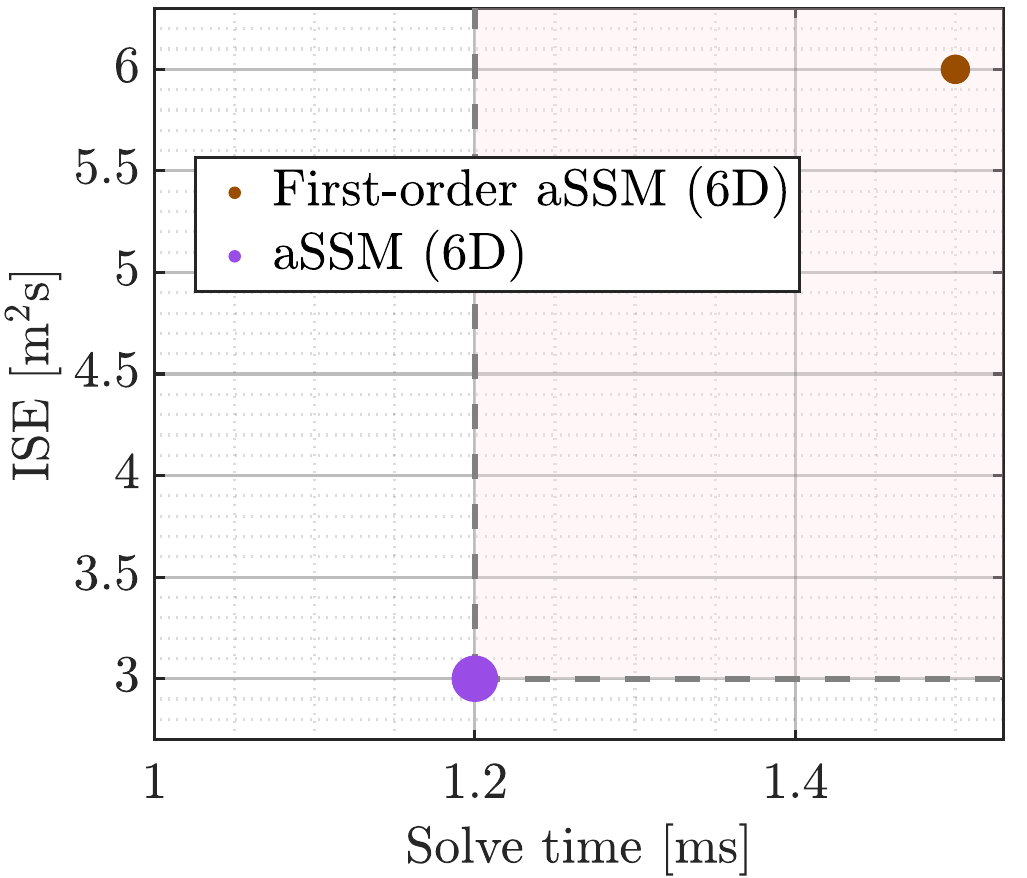}}
    \caption{(a) Relative ISE bar plots with the 6D aSSM-reduced model as the baseline. (b) Pareto plot for closed loop performance on the 3D Trifolium track for the 6D aSSM-reduced model and its first-order approximation.}
    \label{fig:ise_pareto_arm}
\end{figure*}

\begin{figure*}
    \centering
    \subfloat[aSSM (6D)]{\includegraphics[width=0.4\textwidth]{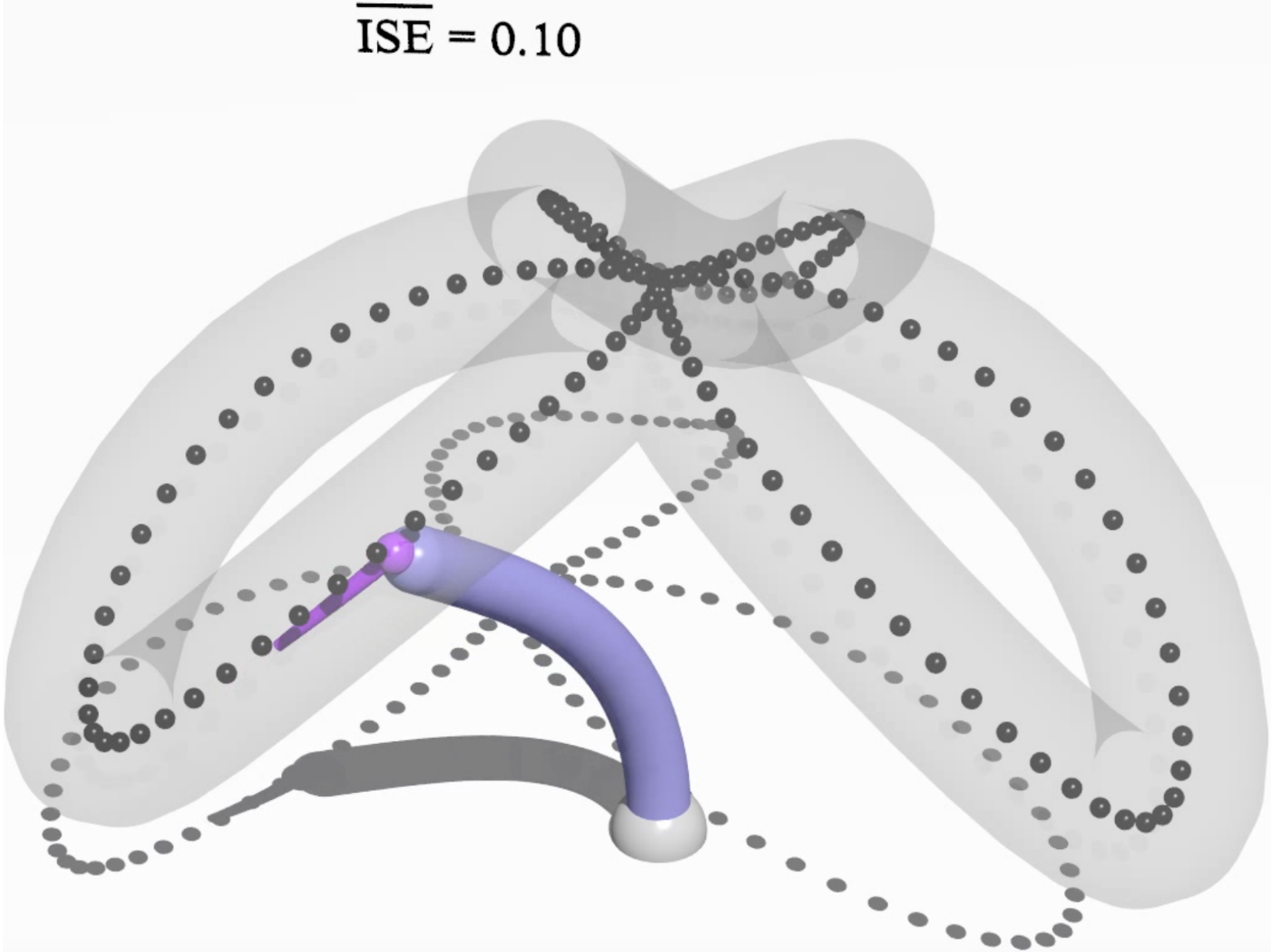}}
    \hspace{0.05\textwidth}
    \subfloat[First-order aSSM (6D)]{\includegraphics[width=0.4\textwidth]{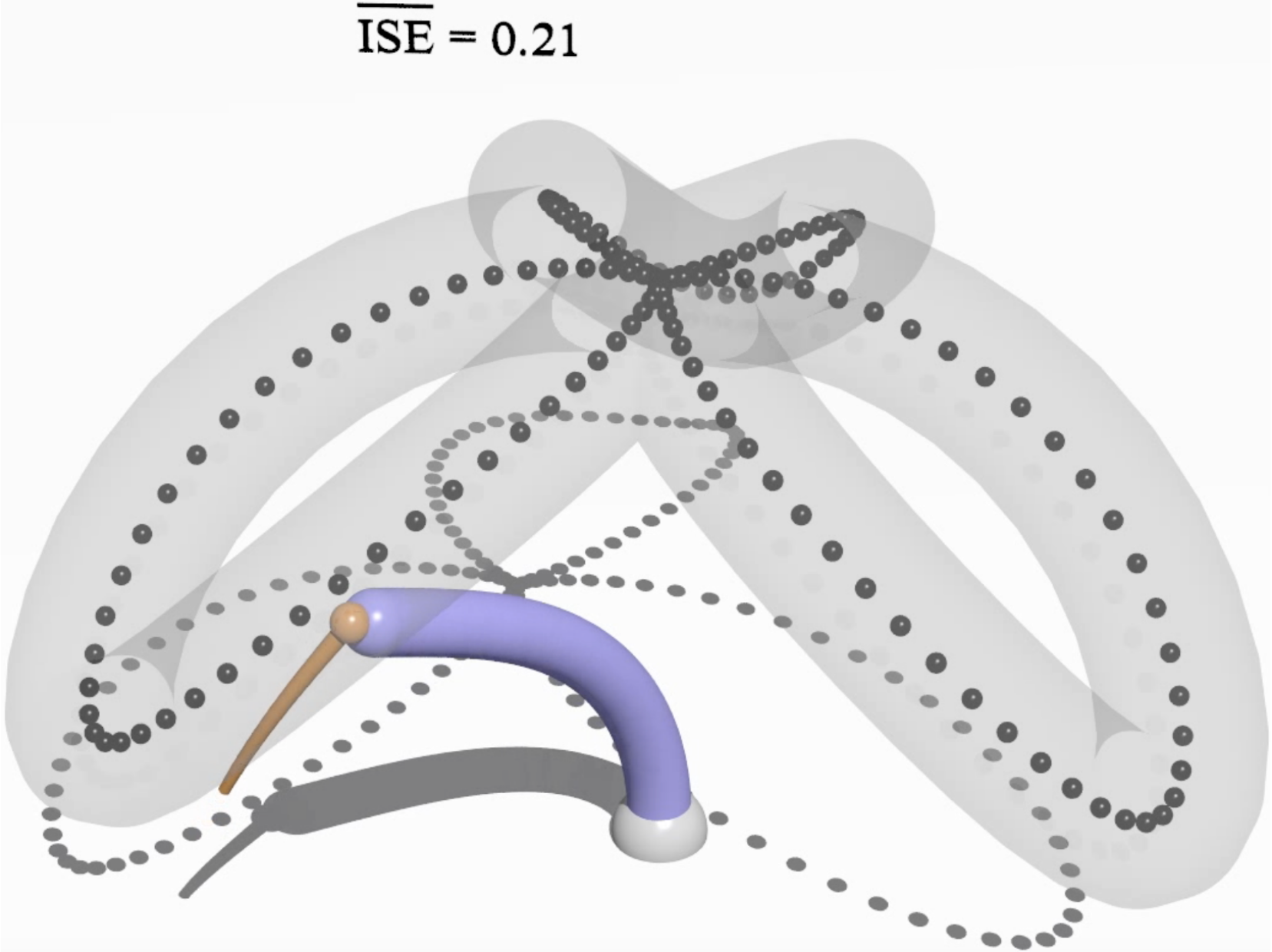}}
    \hspace{0.05\textwidth}
    \subfloat[Koopman w/Pregain LQR (6D)]{\includegraphics[width=0.4\textwidth]{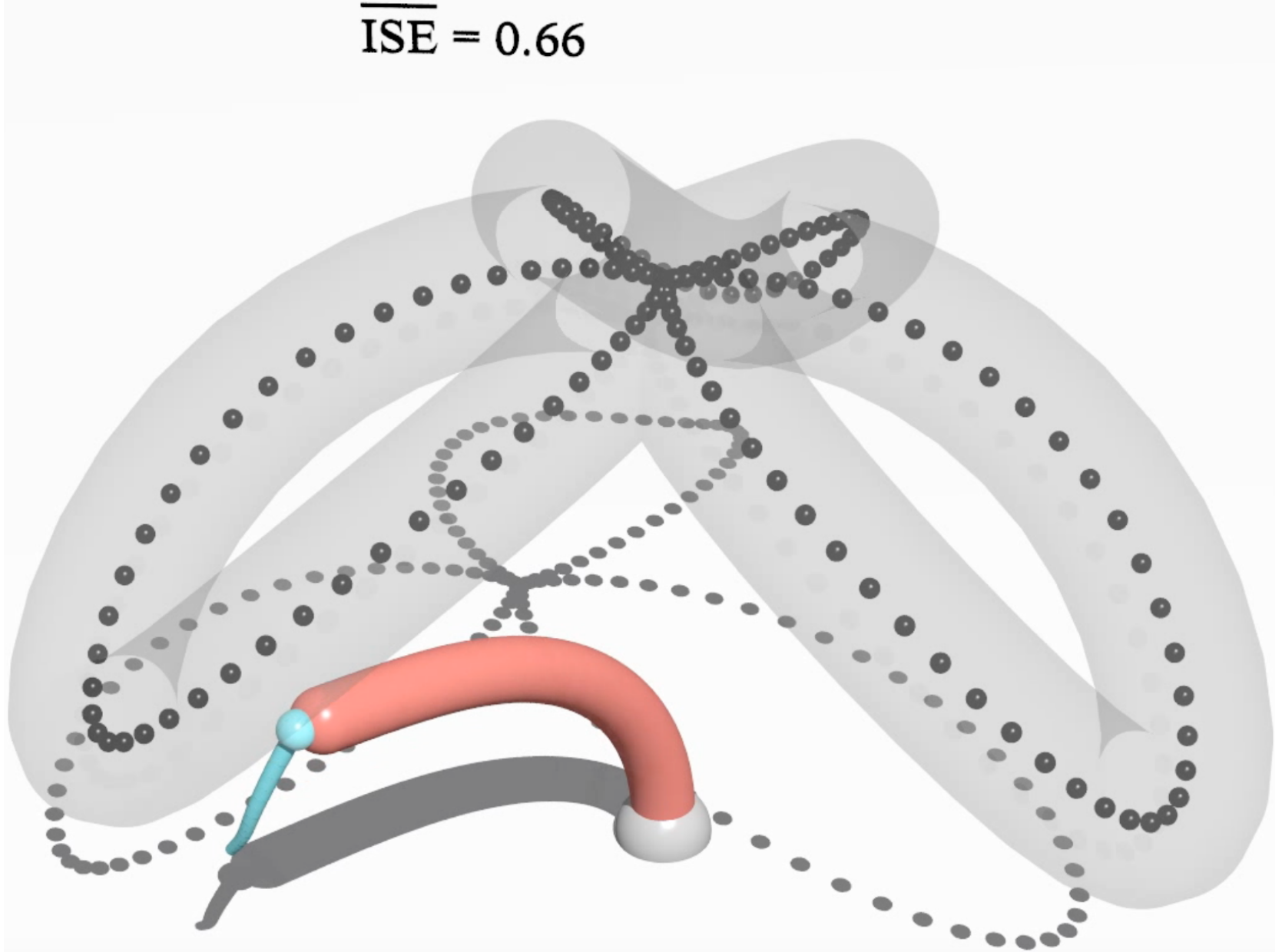}}
    \caption{Short elastic arm snapshots at $t=3 \text{ [s]}$. The target track is depicted with black dots, the operational tolerance tube (OTT) with radius $15 \text{ [cm]}$ in gray and the normalized ISE displayed on top. Model predictions steering away from the OTT result in the soft arm flashing in red. See Multimedia Extension 1 in the Supplementary Material for the full short elastic arm's evolution.}
    \label{fig:arm_full}
\end{figure*}

We support these observations in Fig.\ref{fig:ise_pareto_arm}a-b, by plotting bar plots for the relative ISE and the Pareto plots comparing ISE with solve times. Across all metrics, the 6D aSSM-reduced model offers the lowest ISE $\approx 3 \text{ [m$^2$s]}$  and the lowest average solve time $\approx 1 \text{ [ms]}$. The first-order aSSM-reduced approximation has comparable solve times but overall accumulates errors, leading to larger ISE due to the model's inaccuracy in the extended target regions. The linear Koopman static pregain tracks worse in the extended target regions because the linear approximation is invalid there. Specifically, the linear Koopman model's limitations become evident when it extrapolates to large control inputs, leading to increasing instability in the soft arm's response. The controller counteracts this instability by clipping the predicted control inputs (see Fig.\ref{fig:cl_arm}c). In contrast, our results show that nonlinear aSSM-reduced models achieve robust extrapolation accuracy in these regions (see Fig.\ref{fig:cl_arm}a-b), relying solely on their inherent nonlinearity to predict optimal control inputs without requiring corrective control measures. In Fig.\ref{fig:arm_full}a-c, we also present snapshots of the elastic arm tracking for each model's prediction and display the accumulated normalized ISE\footnote{The ISE is normalized by dividing it by the target's total time and the square of the undeformed arm's length and is denoted as $\overline{\mathrm{ISE}}$.}.

We have not included comparisons with the zeroth-order aSSM approximation, as we know from theory that the method is inapplicable to track large target tracks, which we have systematically demonstrated for the soft trunk robot example in Section \ref{sec:cl_trunk}. We also note that the TPWL method of \citet{tonkens21} cannot be implemented in the elastic arm setup, as it relies on a finite-element model of the arm, which is not readily available.

Our closed-loop MPC scheme for the elastic arm uses optimized evaluation techniques for the aSSM-reduced model, yielding a 6D aSSM-reduced model with significantly lower average solve times of 1 [ms] for a 0.06 [s] MPC horizon. These solve times are 10 times lower than those reported in \citet{alora24} for hardware MPC experiments with simpler 6D SSM-reduced models. With these upgrades, higher-dimensional aSSM-reduced models offer better accuracy with faster solve times. We note that shorter solve times alone do not necessarily translate into improved performance in high-speed tasks. Indeed, the robot’s sampling time is typically hardware-constrained irrespective of the specific task. Therefore, what ultimately matters in a hardware implementation is the reduced model’s accuracy in predicting the dynamics.

\subsection{Short elastic arm: Closed-loop results in the presence of experimental noise}
\label{subsec:noise_short_arm}

In a hardware setting, the elastic arm's end effector is tracked by a motion capture system, which approximates its position as a rigid body using markers embedded on the elastic arm. OptiTrack systems are commonly used in academic robotic hardware systems to record observations. They offer millimeter-accuracy when there are enough cameras. Their placement ensures successful calibration without occlusions and there are no external light sources that introduce noise. However, in most practical hardware applications, it is difficult to meet these requirements.

From the OptiTrack documentation \citet{optitrack_docs}, a poorly calibrated setup results in an error range of approximately 0.5-1 [cm]. To mimic this practical limitation in our simulations, we add noise sampled from a bounded Gaussian distribution with an upper bound of 1 [cm] to the feedback and use it as the initial condition for the MPC. To account for this additional operational uncertainty, we increase the operational tolerance radius by 1 centimeter.

\begin{figure*}
    \centering
    \subfloat[]{\includegraphics[width=0.4\textwidth]{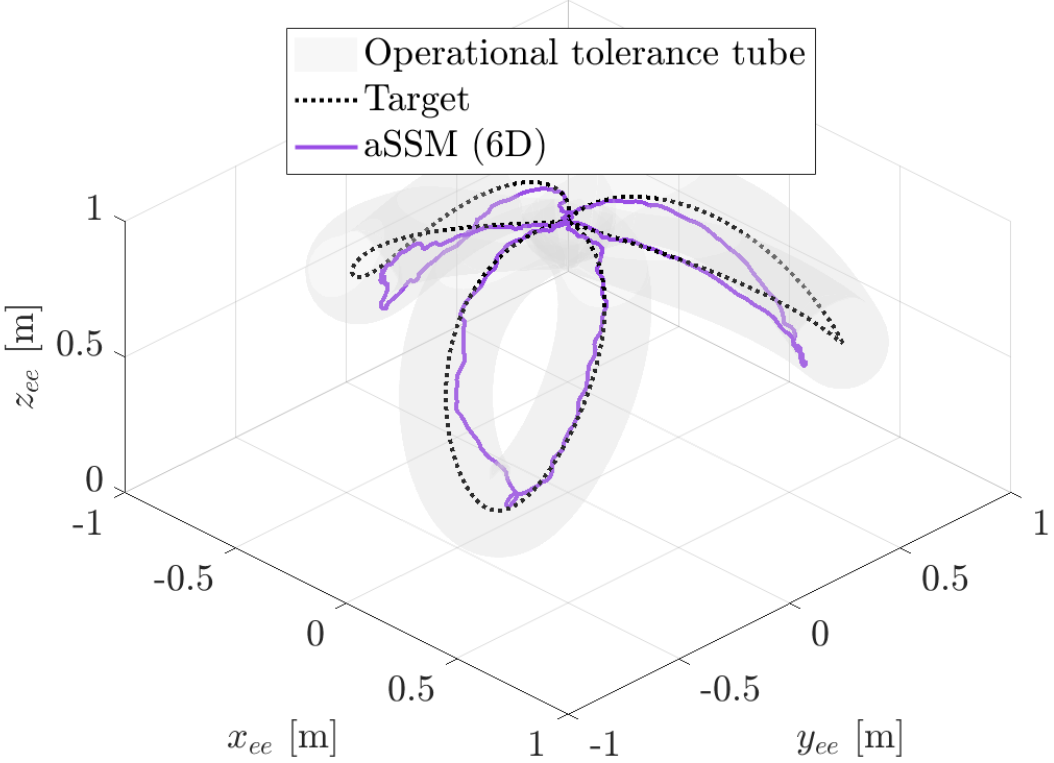}}
    \hspace{0.05\textwidth}
    \subfloat[]{\includegraphics[width=0.4\textwidth]{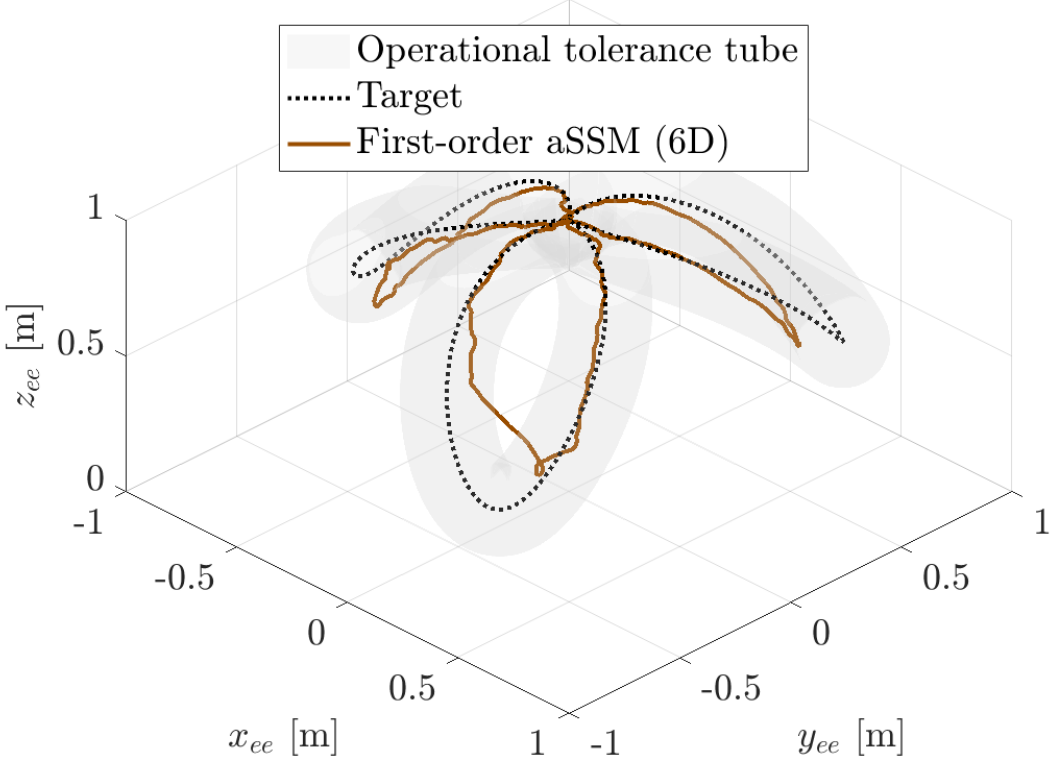}}
    \hspace{0.05\textwidth}
    \subfloat[]{\includegraphics[width=0.4\textwidth]{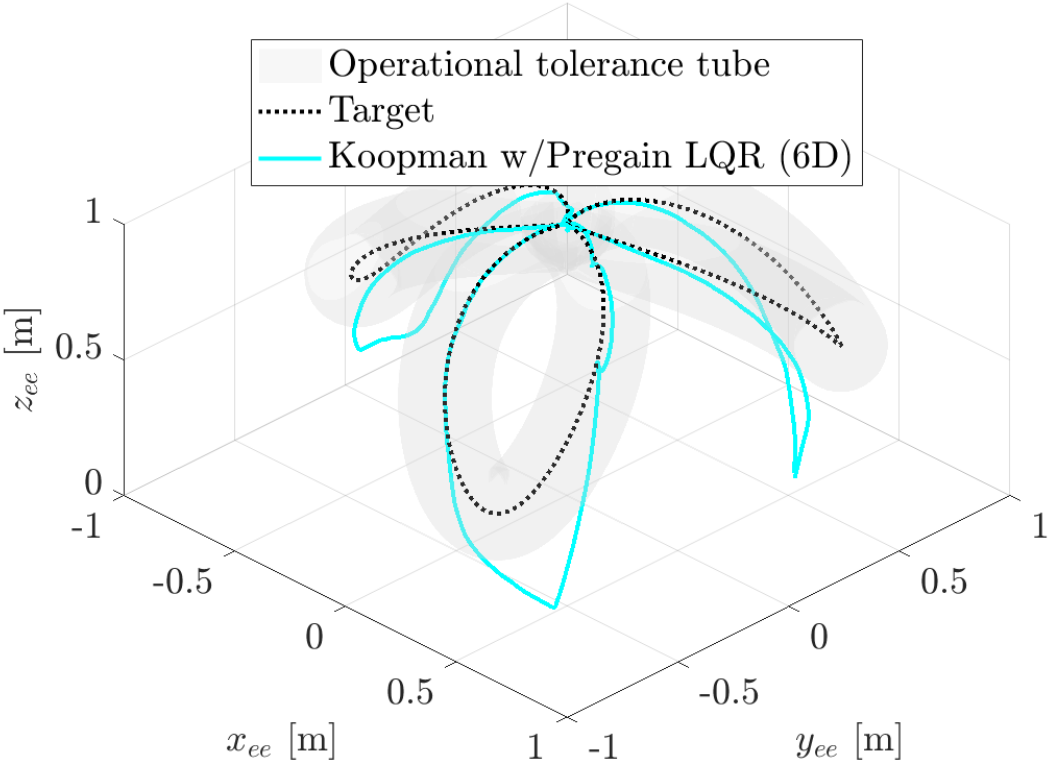}}
    \hspace{0.02\textwidth}
    \subfloat[]{\includegraphics[width=0.57\textwidth]{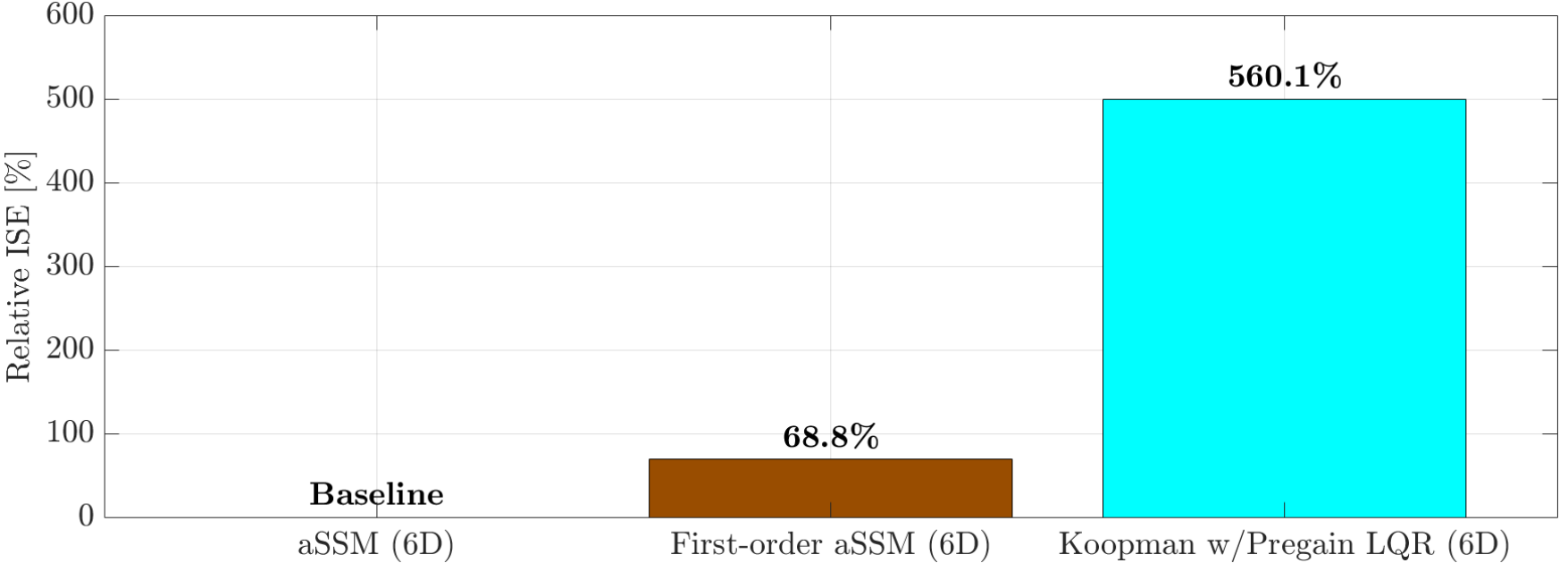}}
    \caption{Closed-loop prediction plots in the short elastic arm's workspace in the presence of noise for (a) 6D aSSM-reduced model, (b) 6D first-order aSSM-reduced model, and (c) 6D Koopman static pregain method. The target track is plotted in a black dotted line, and the gray shaded tube represents the allowed operational tolerance of the soft arm's end effector. (d) Relative ISE bar plots with the 6D aSSM-reduced model as the baseline.
}
    \label{fig:cl_arm_noise}
\end{figure*}

We test the closed-loop performance of all three models for the Trifolium target track discussed in the previous section in the presence of feedback noise discussed above. In Fig. \ref{fig:cl_arm_noise}a-c, we plot the closed-loop predictions in the robot's workspace. We observe that camera feedback noise causes the short elastic arm to jitter in the 6D aSSM-reduced model and in its first-order approximation. The jitter is not visible in the Koopman static pregain model, as the method is dominated by the linear mapping that acts directly on the target track, and the dynamic feedback contribution is minimal and does not significantly affect it. Overall, the 6D aSSM-reduced model remains within the operational tolerance tube when the other methods tend to exit it and develop larger ISE errors (see Fig. \ref{fig:cl_arm_noise}d).

\subsection{Long elastic arm: Closed-loop results for a $3$D Spiral track}

\begin{figure*}
    \centering
    \subfloat[]{\includegraphics[width=0.4\textwidth]{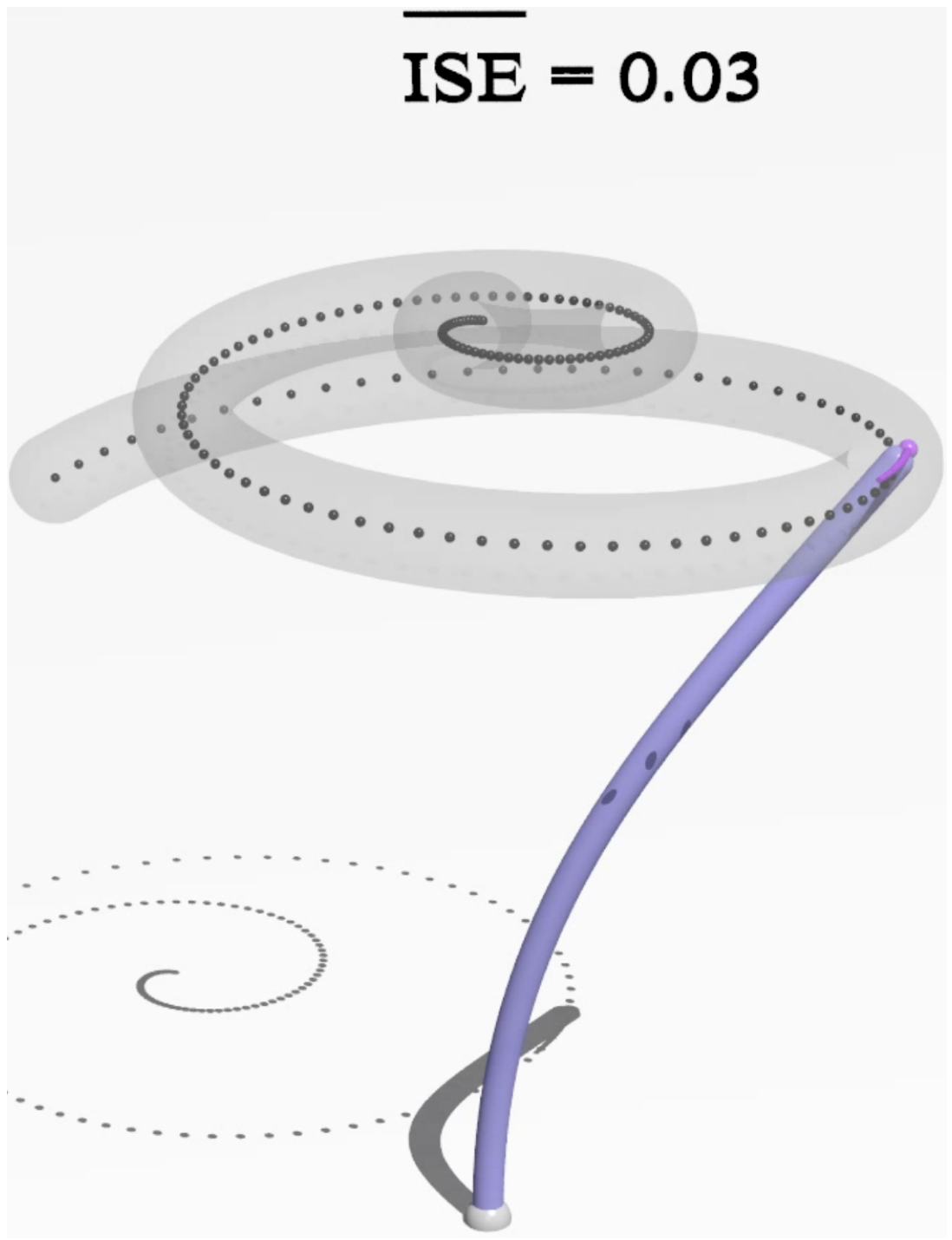}}
    \hspace{0.05\textwidth}
    \subfloat[]{\includegraphics[width=0.4\textwidth]{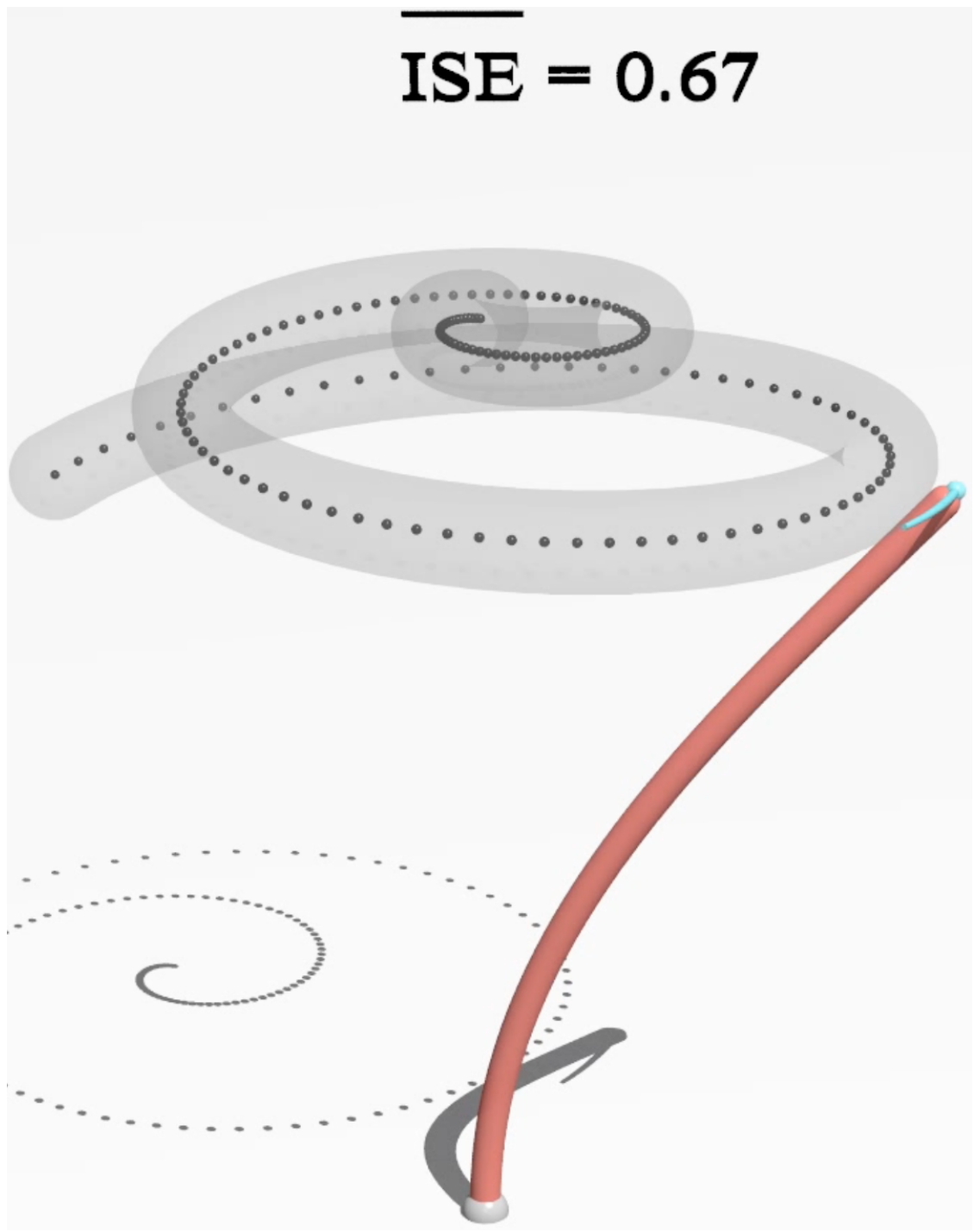}}
    \hspace{0.05\textwidth}
    \subfloat[]{\includegraphics[width=0.45\textwidth]{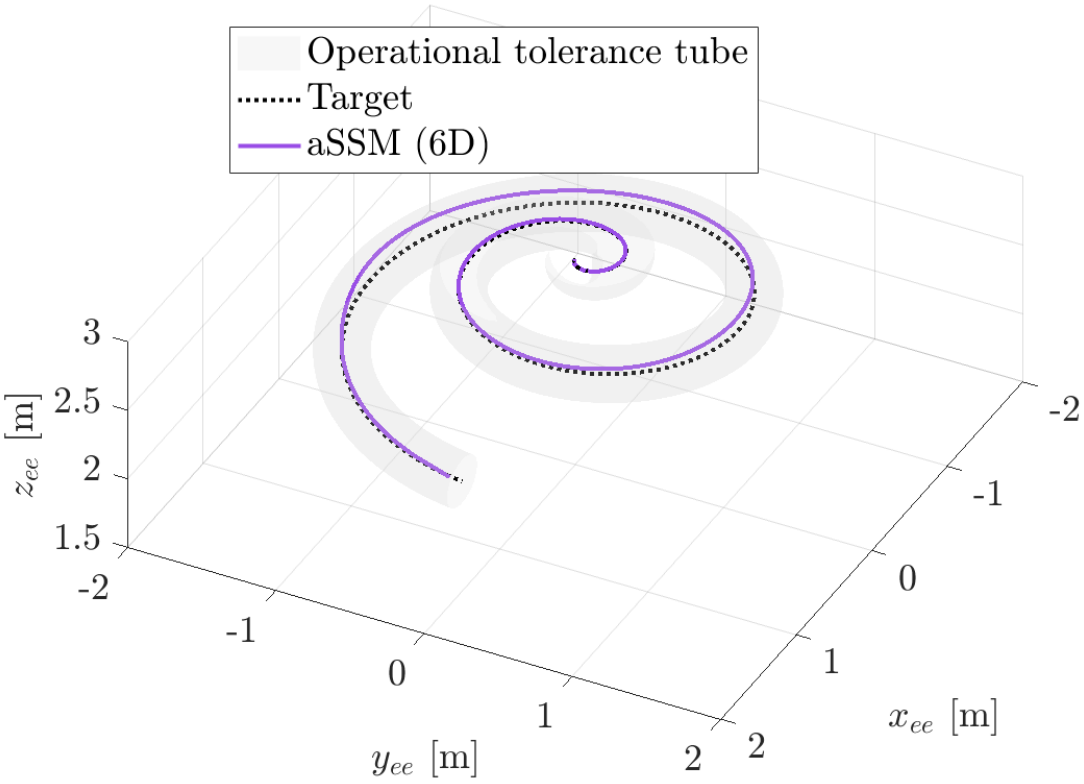}}
    \hspace{0.05\textwidth}
    \subfloat[]{\includegraphics[width=0.45\textwidth]{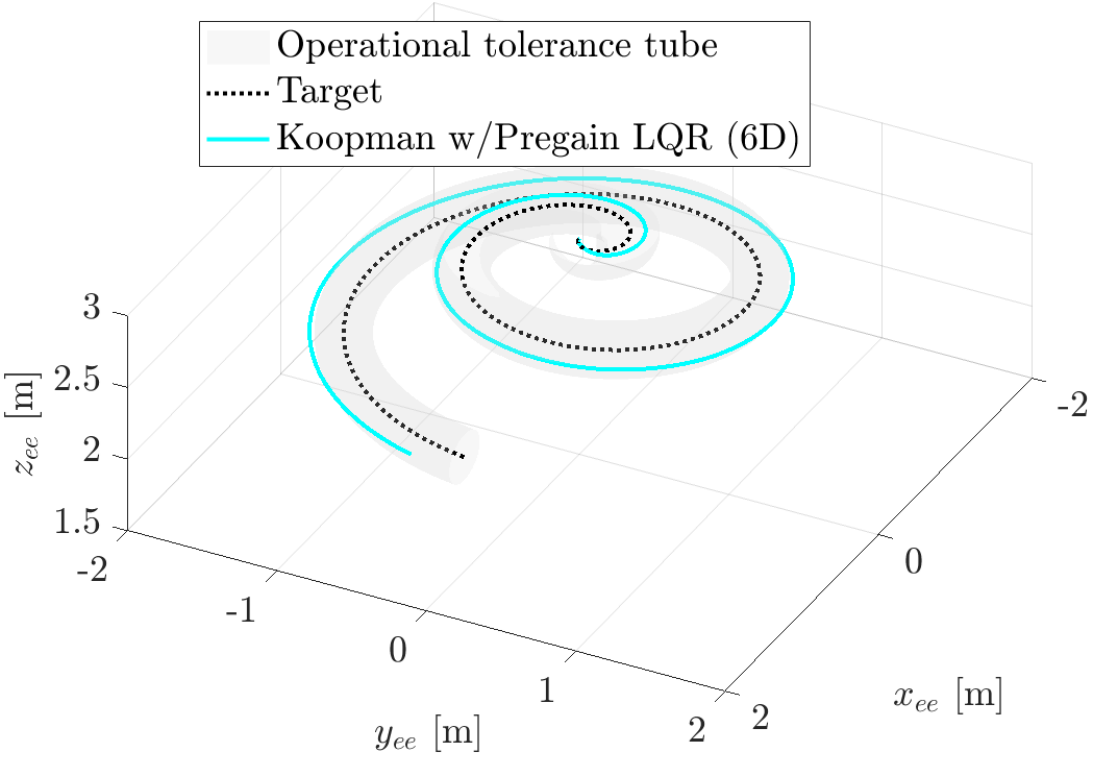}}
     \hspace{0.05\textwidth}
    \subfloat[]{\includegraphics[width=0.45\textwidth]{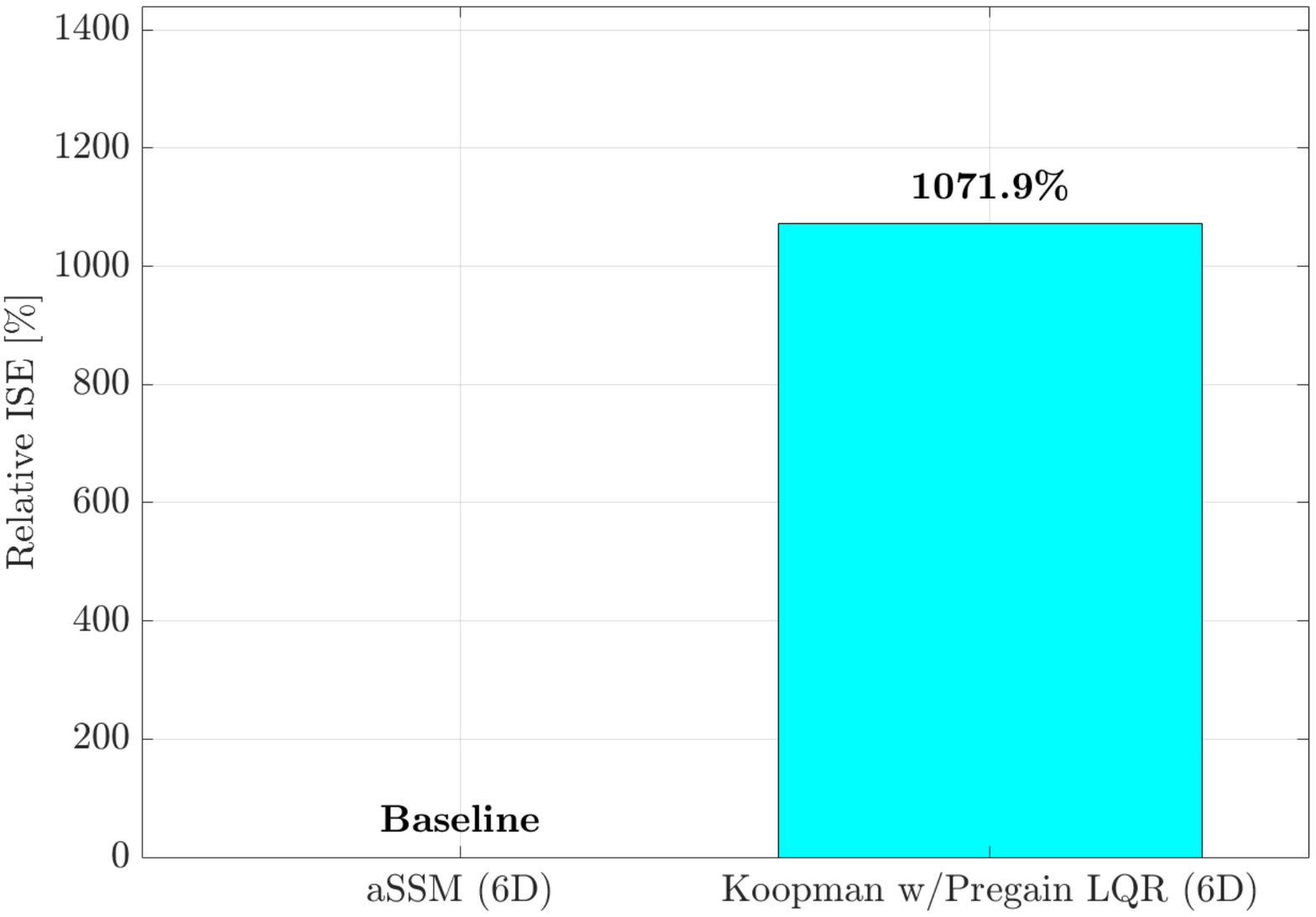}}
    \caption{(a)-(b) Long elastic arm snapshots at $t=7 \text{ [s]}$. The target track is depicted with black dots, the operational tolerance tube (OTT) with a radius of $17.5 \text{ [cm]}$ is shown in gray, and the normalized ISE is displayed on top. Model predictions steering away from the OTT cause the soft arm to flash red. See Multimedia Extension 1 in the Supplementary Material for the full evolution of the long elastic arm. Closed-loop prediction plots in the long elastic arm's workspace for (c) 6D aSSM-reduced model and (d) 6D Koopman static pregain method. (e) Relative ISE bar plots with the 6D aSSM-reduced model as the baseline.}
    \label{fig:long_arm_results}
\end{figure*}

We conclude this section with a closed-loop demonstration for the long elastic arm on a spiral-shaped target track with a mean instantaneous speed of $103.3 \text{ [cm/s]}$ and slowness measure $r_s = 1.3$. The track starts from the undeformed arm's configuration and spirals downward to a height of $2.3 \text{ [m]}$. The spiral track is depicted as a dotted curve in Fig. \ref{fig:long_arm_results}a. For this case, we perform closed-loop comparisons for the 6D aSSM-reduced model and the Koopman pregain static method. We set the planning horizon for the MPC scheme as $0.1 \text{ [s]}$ and the workspace cost matrix $\mathbf{Q}_z = 100 \cdot\mathbf{1}_{3 \times 3}$ for both models. We identified that $\mathbf{R}_u = 0.5\cdot\mathbf{1}_{8 \times 8} $ for the 6D aSSM-reduced MPC scheme and $\mathbf{R}_u = 0.01\cdot\mathbf{1}_{8 \times 8} $ for the Koopman static pregain method produced reasonable optimal control inputs that led to the lowest errors during closed-loop. We omit the first-order approximation of the aSSM-reduced model here because the results showed performance gains similar to those already demonstrated for the short elastic arm.

Figure \ref{fig:long_arm_results}a-b shows snapshots of the closed-loop results of the long elastic arm for the 6D aSSM-reduced model and the Koopman static pregain method. In both plots, we denote the operational tolerance tube (OTT) in gray, with a radius of 17.5 [cm] amounting to $2\%$ of the spiral target's length. We observe that the aSSM-reduced model stays within the OTT and provides the lowest ISE (see Fig. \ref{fig:long_arm_results}c-e) compared to the Koopman static pregain method. The example yet again demonstrates the limitations of linear methods for accurately tracking the target in extended regions, where the trunk assumes nonlinear shape profiles to course-correct and achieve lower errors, which our nonlinear aSSM-reduced model can do. In Appendix \ref{app:noise_long_arm}, we also briefly present the long elastic arm's closed-loop results in the presence of experimental noise, as in the short elastic arm's case, our aSSM-reduced model robustly tracks the target track.

\section{Conclusions}
We have developed a data-driven methodology for designing low-dimensional predictive models to control soft robots. Our approach is built on the recent theory of adiabatic spectral submanifolds (aSSMs), which are very low-dimensional attracting invariant manifolds tangent to dominant eigenspaces of dynamical systems at their fixed points. We construct aSSMs from uncontrolled trajectory data and learn the impact of control on the aSSM dynamics from randomly generated control inputs. The resulting model is then universally applicable to arbitrary tracks in the workspace of a soft robot.

Control design from aSSM reduction is applicable under a relative time scale separation assumption that holds for
typical soft robotic applications: the desired trajectory must be slow relative to the rate at which the internal oscillations of the robot decay. In contrast, linear methods, such as the TPWL and Koopman operator methods, assume that the soft robot behaves linearly. This is not expected to hold for a geometrically and materially nonlinear soft robotic structure undergoing large deformations, as we have indeed found in our examples. Specifically, our systematic comparisons in Appendix \ref{app:C} and \ref{app:D} highlight that these linear methods fail to track even slow or moderately fast targets.

We have shown across different target sizes, speeds, and hard constraints, that aSSM-reduced models provide the best control performance on trunk robots when compared with other available data-driven modeling techniques. The reason for this remarkable performance is rooted in the robustness of the adiabatic SSM theory. When challenged to account for bounded fast feedback control deviations in closed loop control, the structural stability of aSSMs enabled us to extend our approach to more general control input ranges, thus allowing slow targets to have additional fast bounded variation along their slow time horizon. This extension can also be viewed as combining existing theories on temporal SSMs for slow or weak forcing  discussed and proved by \citet{haller24_wa}.  
We validate and illustrate this by adding chaotic torque control deviation to a slowly moving double pendulum. These extensions for the aSSM theory were used to devise an effective finite-time horizon control strategy in Section \ref{sec:fh_horizon}.  

Our aSSM-reduced MPC scheme (see eq.(\ref{eq:aSSM_mpc})) and aSSM learning methodology can also incorporate practical constraints, such as dealing with limited observations, quick model run times, model applicability for a user-defined MPC horizon, and generalizability to all possible tracks.
The aSSM-reduced MPC scheme enhances the earlier SSM-based MPC scheme used for control of soft robots (see \citet{alora23b}, \citet{alora23} and \citet{alora24}). Staying true to these practical constraints, we learn $4$D, $5$D or $6$D aSSM-reduced MPC schemes for a finite-element model describing a soft trunk and a Cosserat rod model of a soft elastic arm from just tip positional data. We further systematically evaluate the performance of these schemes on previously unseen inputs across operational ranges not encountered during training. Our analysis shows that a $4$D aSSM-reduced model can track general planar targets and a $5$D or $6$D aSSM-reduced model generalizes well to any user-defined target in the full workspace. We emphasize the low-dimensionality of our reduced-order models, in contrast to models yielded by linear reduced-order modeling methods. 

Our future objective is to effectively transfer this learning methodology to hardware experiments on soft robots. Our current learning methodology relies on detailed training data collection about each static SSM. We seek to automate this aspect for real-life experiments by using autonomous training data from random step control input responses. Eventually, we plan to perform aSSM identification from that dataset directly. We also expect challenges in the control calibration procedure and expect to learn possibly linear or nonlinear control actions as well. We also anticipate further hardware-related noise and latency effects from camera feedback that will influence the currently reported closed-loop performance. For the former, within our simulation setups, we have demonstrated the robustness of aSSM methods to noise.

While a hardware implementation lies beyond the scope of this paper (which is primarily focused on the theoretical foundations of our methodology), we have strong reasons to expect that the hardware experiments will corroborate the simulation results presented here. Indeed, SSM-based control was already implemented in a hardware setting by \citet{alora23b,alora24}, \citet{yan2024refinedmotioncompensationsoft} and outperformed the TPWL and Koopman approaches, as predicted by prior numerical simulations, despite noise, delay, and unmodeled dynamics.  A reason for this is that aSSMs are known to be normally hyperbolic invariant manifolds and hence are structurally stable (i.e., robust under small perturbations). A hardware implementation of our trunk robot example is, in fact, currently underway at Stanford University and will be the subject of a forthcoming publication.


\begin{acks}
This work has been partially supported by a grant from the Swiss National Science Foundation (SNF).
\end{acks}

\bibliographystyle{SageH}
\bibliography{SSM_bibliography,sample}

\appendix
    
\makeatletter
\let\originalsection\section

\renewcommand{\section}[1]{%
  \refstepcounter{section}
  \addcontentsline{toc}{section}{\thesection\ #1}
  \originalsection*{APPENDIX \thesection: #1}
}
\makeatother

\section{Mechanical system definitions for robotic simulators}
\label{app:physical_models}
\subsection{Soft trunk}
he soft trunk robot geometry is densely meshed using smaller polyhedra or hexahedra 3D objects. We denote the generalized displacement of all nodes in the meshed trunk description as $\mathbf{n} \in \mathbb{R}^{n_f}$. The mesh material needs to be deformable and is governed by a nonlinear elastic continuum law that depends on the material's Young's modulus and Poisson's ratio, determined from the linear elastic regime. 

Along certain mesh regions, nodes make contact with cables actuators, the forces imparted on the mesh due to cable actuation are given by the function $\mathbf{B}(\mathbf{n}) \mathbf{u}$, where $\mathbf{u} \in \mathbb{R}^{n_u}$ denotes the control input vector. Employing the finite element method with these mesh properties and cable actuation constraints, one can approximate the partial differential equations describing the soft trunk into a finite-dimensional second-order ordinary differential equation,
\begin{equation}
\mathbf{M}(\mathbf{n}) \ddot{\mathbf{n}} = \mathbf{F}(\mathbf{n},\dot{\mathbf{n}}) +  \mathbf{B}(\mathbf{n}) \mathbf{u}(t) + \mathbf{F}_{ext}(t).
\end{equation}
Here $\mathbf{M}(\mathbf{n})$ is the mass matrix, $\mathbf{F}(\mathbf{n},\dot{\mathbf{n}})$ is the nonlinear internal forcing, $ \mathbf{B}(\mathbf{n}) \mathbf{u}(t)$ is the forcing due to cable actuation, and $\mathbf{F}_{ext}(t)$ is a generalized external forcing acting on the soft trunk. In our trunk setup, the mesh consists of $n_f = 709$ nodes and $n_u = 8$ cables, with the external force held constant at the Earth's gravitational force. For exact details on the implementation of the nonlinear elastic internal force and the actuation force, see \citet{faure2012sofa}.

\subsection{Soft arm}

The soft arm is modeled as a Cosserat rod with undeformed length $l$, cross-sectional area $A$, mass per unit length $\rho$, second area moment of inertia $\mathbf{I}$, and bending and shearing stiffness matrices $\mathbf{B}$ and $\mathbf{S}$. 

We denote the centerline coordinate of the rod as $\mathbf{x}(s,t) \in \mathbb{R}^3$, the angular momentum of the rod $\boldsymbol{\omega}(s,t) \in \mathbb{R}^3$, and the generalized curvature of the rod $\boldsymbol{\kappa}(s,t) \in \mathbb{R}^3$, where $s \in [0,l]$ is a material coordinate and $t$ is time. We also define the rotation matrix $\mathbf{T} \in \mathbb{R}^{3\times 3}$ that transforms vectors from the lab frame to the rod's local frame. Then, the governing PDEs of the soft arm  are given by
\begin{align}
 \partial_t \mathbf{T} &= \hat{\boldsymbol{\omega}}\mathbf{T}, \nonumber\\
\rho A \partial_t^2 \mathbf{x} &= \mathbf{F}_{SS} \left(\mathbf{T}, \mathbf{S}, \partial_s \mathbf{x}, \mathbf{x}\right) + \|\partial_s \mathbf{x}\| \mathbf{F}_{ext}(t), \\
\rho \mathbf{I} \partial_t \boldsymbol{\omega} &= \mathbf{F}_{B} \left(\boldsymbol{\kappa}, \mathbf{B}, \|\partial_s \mathbf{x}\|\right) 
 + \mathbf{F}_{SC} \left(\mathbf{T}, \mathbf{S}, \partial_s \mathbf{x}\right) \nonumber \\ & \quad \quad + F_{AD} \left(\rho, \mathbf{I},\boldsymbol{\omega},  \|\partial_s \mathbf{x}\|\right) + \|\partial_s \mathbf{x}\|^2 \mathbf{C}_{ext}(t), \nonumber \end{align}
where $\hat{\boldsymbol{\omega}}$ is the skew-symmetric matrix of $\boldsymbol{\omega}$, $\mathbf{F}_{SS}$ represents shearing or stretching forces,  $\mathbf{F}_{B}$ represents bending or twisting forces,  $\mathbf{F}_{SC}$ is an internal coupling force, $\mathbf{F}_{AD}$ represents advection and dilation forces, $F_{ext}$ is the external force and $C_{ext}$ is the external torque. To mimic realistic damping in these systems, a proportional damping term per unit length is introduced after spatial discretization of the soft arm's PDE. The external torques act on user-defined control points along the rod's length, which serve as the control actuation forces for the arm. For exact forms of the internal force terms and the damping force terms, refer to \citet{gazzola18}. For the actuation settings related to our soft arm example, refer to \citet{naughton21}.

\color{black}

\section{Open loop performance for faster varying inputs}
\label{app:D}

\begin{figure*}
    \begin{centering}
    \includegraphics[width=1\textwidth]{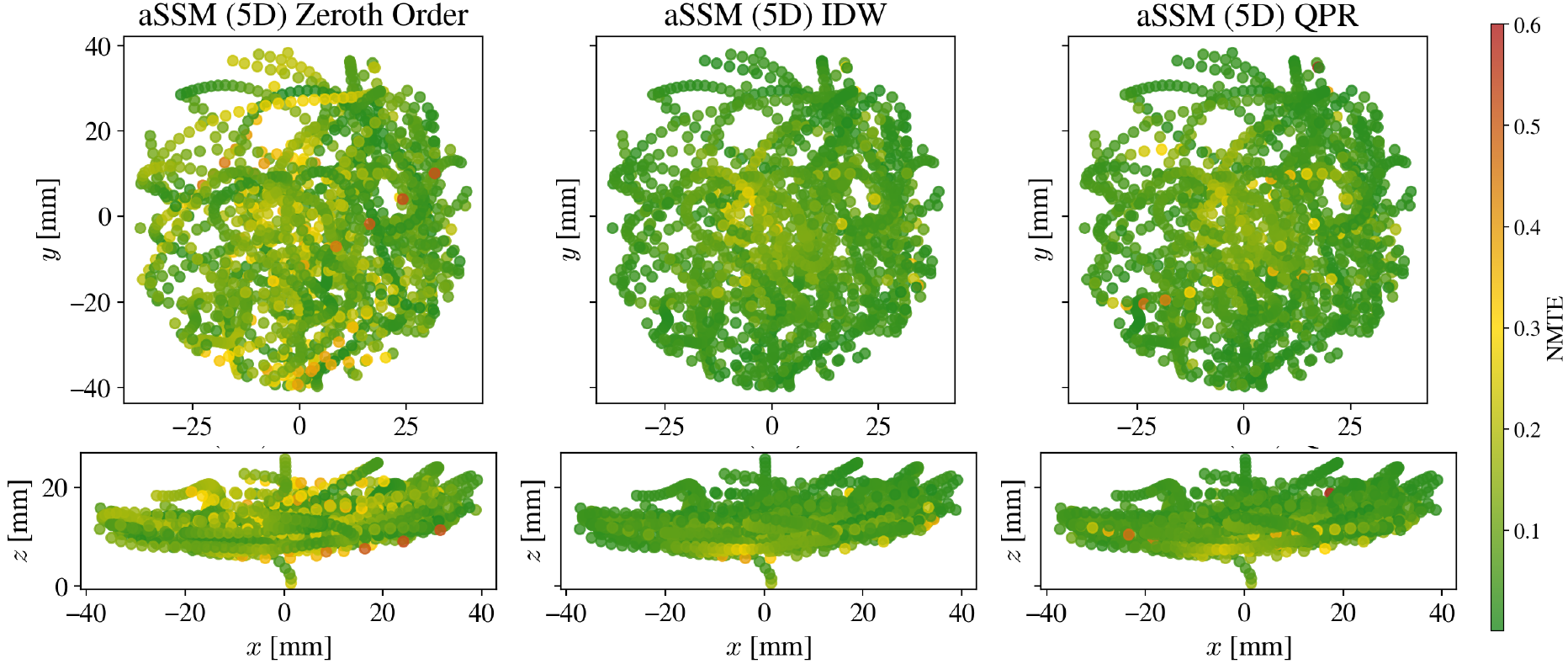}
    \par\end{centering}
    \caption{(Top) $x_{ee}-y_{ee}$ scatter color plots of the random initial conditions for $5$D aSSM-reduced model simulations on a fast varying input. (Bottom) $z_{ee}-x_{ee}$ scatter color plots of the random initial conditions. Color indicates NMTE values for the open loop prediction.}
    \label{fig:open_loop_fast}
    \end{figure*}

We plot open-loop validation results for faster varying control input, with the mean instantaneous speed of $ 265 \text{ [mm/s]}$ for the end effector; the slowness measure evaluates to $r_s \approx 3.7$. The results in Section \ref{sec:open_loop} were for a mean speed of $125 \text{ [mm/s]}$ with $r_s \approx 1.7$.  Figure \ref{fig:open_loop_fast} shows us that even for fast varying inputs, aSSM-reduced models show overall good prediction accuracy. Compared to the slower trajectory, the scatter plots show more yellow-green regions, suggesting a slight dip in accuracy. Still, we expect these errors to be corrected by an optimization algorithm in a closed-loop setting. 

\section{TPWL and Koopman static pregain implementations}
\label{app:A}
\subsection{TPWL implementation for soft trunk robot}
\label{app:A.TPWL}
We closely follow the TPWL method implemented in \citet{tonkens21}. We outline the steps as follows:
\begin{itemize}
    \item Step 1: Collect 5 random control input sequences with a maximum value $\delta = 250$ using a Latin hypercube sampling technique. Subject the trunk robot to these input values.
    \item Step 2: Set up a snapshot matrix for the controlled responses of the trunk. Perform SVD on this dataset and collect singular vectors corresponding to the $28$ largest singular values and form the reduced basis.  
    \item Step 3: Evaluate the Jacobian of the full model at the initial condition of the random control responses. Reduce the Jacobian using the learned basis from Step 2.  
    \item Step 4: Run the SVD-reduced linear model subjected to general control inputs and compare the predictions with the full response. 
    \item Step 5: Stop step 4 when the errors in the predictions exceed a user-defined threshold. Store the SVD-reduced linear model reduced by SVD and the initial condition obtained from step 3. 
    \item Step 6: Repeat steps 3 and 4 at the violation point. Stop after scanning all the random control response trajectories. 
\end{itemize}
We evaluate the error as a pointwise relative error for both velocities and positions and set the threshold to $10 \%$. Here, our method differs from the original TPWL method, which only evaluates the errors in velocities for a threshold value of $200$. Our settings resulted in $52$ piecewise $28$-dimensional linear models. 

\subsection{Koopman static pregain}
\label{app:A.Koop}
We implement the Koopman static pregain method described in \citet{haggerty23} for the bionic trunk robot. We choose the observable space to be a delay embedding of the end effector position $\mathbf{y}=\{\mathbf{z}(t),\mathbf{z}(t+\tau)\} \in \mathbb{R}^{6}$, with $\tau = 0.01$. We generate controlled training data by passing a step input signal that combines all static input probing signals used to train the aSSM-reduced model. We arrange the trajectories and the control input as snapshot matrices $\mathbf{Y} \in \mathbb{R}^{6 \times N_t}$ and $\mathbf{U}\in \mathbb{R}^{n_u \times N_t}$, respectively, where $N_t$ is the total length of both trajectories and $n_u$ is dimension of the robot's actuation space. We fit a linear time invariant system to this dataset:
\begin{equation}
    \left\{\mathbf{A}^*, \mathbf{B}^* \right\} = \operatorname*{arg\,min}_{\mathbf{A},\mathbf{B}} \|\dot{\mathbf{Y}} - \mathbf{A}\mathbf{Y} - \mathbf{B} \mathbf{U} \|^2.
\end{equation}

This is defined as the dynamic Koopman operator which is a linear non-autonomous system in the observable space. A random controlled trajectory will behave linearly and be influenced by a linear control action with probability $0$ in a generic observable space. This simple approximation of the soft robot is expected to fail in predicting the actual nonlinear dynamics of the soft robot. In fact, the authors in \citet{haggerty23} observe this in their results and try to remedy their model by accounting for an additional term.   
 
That additional term is called the Koopman static operator. Using the static inputs computed to learn the static SSM dictionaries, we learn this static operator by fitting a linear mapping $\mathbf{G}$ between steady states in workspace $\mathbf{z}_{s}$ and static inputs $\bar{\mathbf{u}}$, this is defined as:
\begin{equation}
    \bar{\mathbf{u}} = \mathbf{G}\mathbf{z}_s.
\end{equation}

They reformulate the optimization problem eq.(\ref{eq:optim_problem}) as a linear quadratic regulator (LQR) with a linear time-invariant dynamical system constraint. They then treat the finite-horizon problem as an infinite-horizon problem and find exact solutions for the desired control inputs required to track the target $\boldsymbol{\Gamma}(t)$ by solving an algebraic Riccati-difference equation for the pregain matrix $\mathbf{K}$.
\begin{equation}
    \mathbf{u}^{*}(t) = -\mathbf{K} (\mathbf{z}-\boldsymbol{\Gamma}(t)).
\end{equation}
Since this resulted in inaccurate results, they added a static correction to the above as, 
\begin{equation}
    \mathbf{u}^{*}(t) = -\mathbf{K} (\mathbf{z}-\boldsymbol{\Gamma}(t)) + \mathbf{G} \boldsymbol{\Gamma}(t).
\end{equation}
Once this term was added, \citet{haggerty23} reported accurate tracking, and their results also indicated that the dynamic part had negligible contributions. Their method can only be reasoned to work when the target trajectory they wish to control is slowly varying and the robot's dynamics is purely linear. They are invariably using these assumptions to first move the robot along slowly varying circular trajectories and then speed up the robot once they have already latched onto the target during the slow run. These methods will ultimately fail to achieve inertial dynamics if from the very start one challenged the Koopman pregain LQR controller to track a fastly accelerated target. 

\citet{haggerty23} also employ significant simplifications to the optimal control problem. Due to this, we expect their method to fail in tracking various targets precisely. This is exactly what we observed for the examples considered in this paper. Also, we note that the construct in \citet{haggerty23} does not take into account the state and input constraints.

\subsection{Data collection and training time comparisons for aSSM, Koopman and TPWL}
\label{app:A.training_time}
Generating the common training data for the aSSM and Koopman approaches took about an hour and a half for both of our examples.  This time includes the numerical simulation of the systems involved, which would not be needed in an experimental implementation.  Therefore, an hour and a half is an upper estimate of what the data collection time would be in an experimental setting.  Once the training data were available, the model was trained in about 25 seconds for the aSSM approach and about 1 second for the Koopman-based approach.  These training times will be the same in an experimental implementation.  For TPWL, generating the training data took less than an hour, whereas training the model took about thirty seconds.  None of these times is extensive, but the results from the different methods show large variations.  Specifically, the accuracy of the aSSM-based models is consistently orders of magnitude higher than that of the Koopman and TPWL models, even though the latter two are higher-dimensional.  All this is not surprising, given that aSSM-reduction targets a specific invariant manifold family that is known to exist in these highly nonlinear systems.  In contrast, the Koopman-based and TPWL approaches fit linear models to those systems.

\section{Comparisons between aSSM, TPWL and Koopman static pregain for the figure-8 track}
\label{app:B}
\begin{figure*}
    \begin{centering}
    \includegraphics[width=1\textwidth]{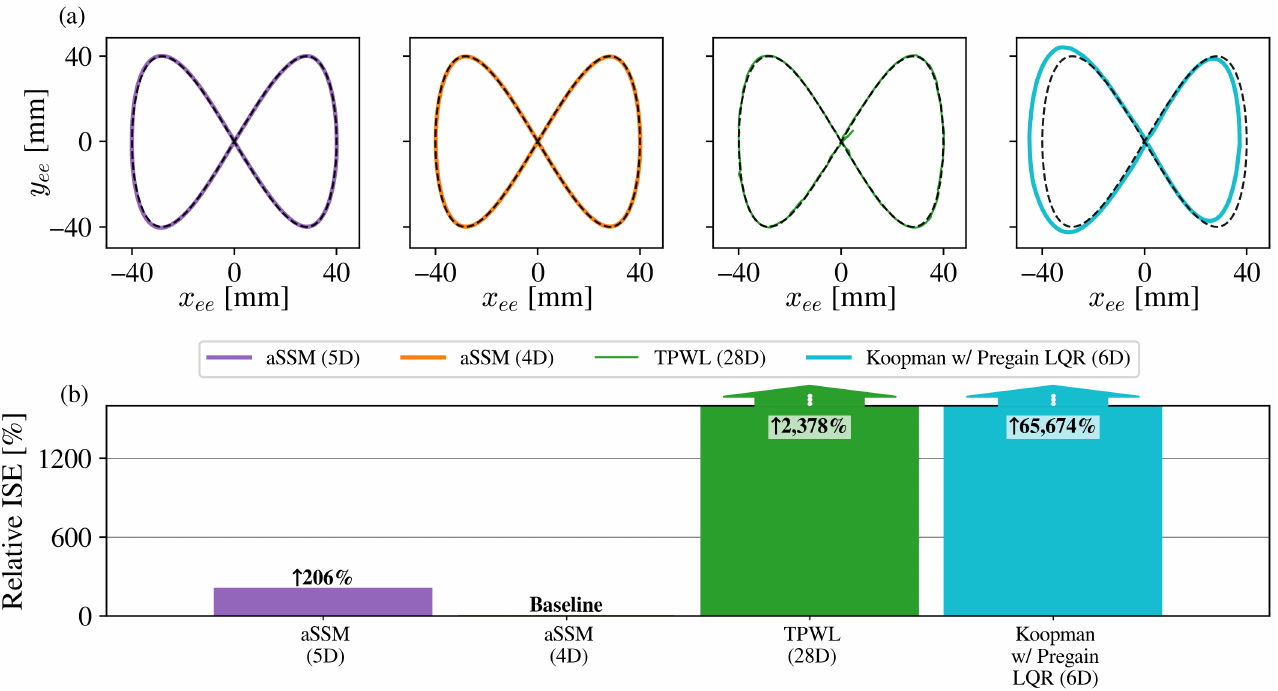}
    \par\end{centering}
    \caption{ (a) Closed-loop results using  linear and aSSM-reduced methods, for the figure-$8$ track plotted in the $x_{ee}-y_{ee}$ plane. (b) Bar plots of relative ISE, with the 5D aSSM-reduced model as the baseline. }
    \label{fig:figure8_appendix}
    \end{figure*}

In Fig. \ref{fig:figure8_appendix}, we plot comparisons of the $5$D aSSM-reduced model with linear methods for closed-loop control of the figure-8 track. 

As expected, we find aSSM models to have the lowest ISE and offer smoother control performance across the full target horizon. TPWL initially falters but works well after latching onto the target. The Koopman pregain overestimates the figure-8 track. 
\section{Comparisons between aSSM, TPWL and Koopman static pregain for the Pacman track}
\label{app:C}
\begin{figure*}
    \begin{centering}
    \includegraphics[width=1\textwidth]{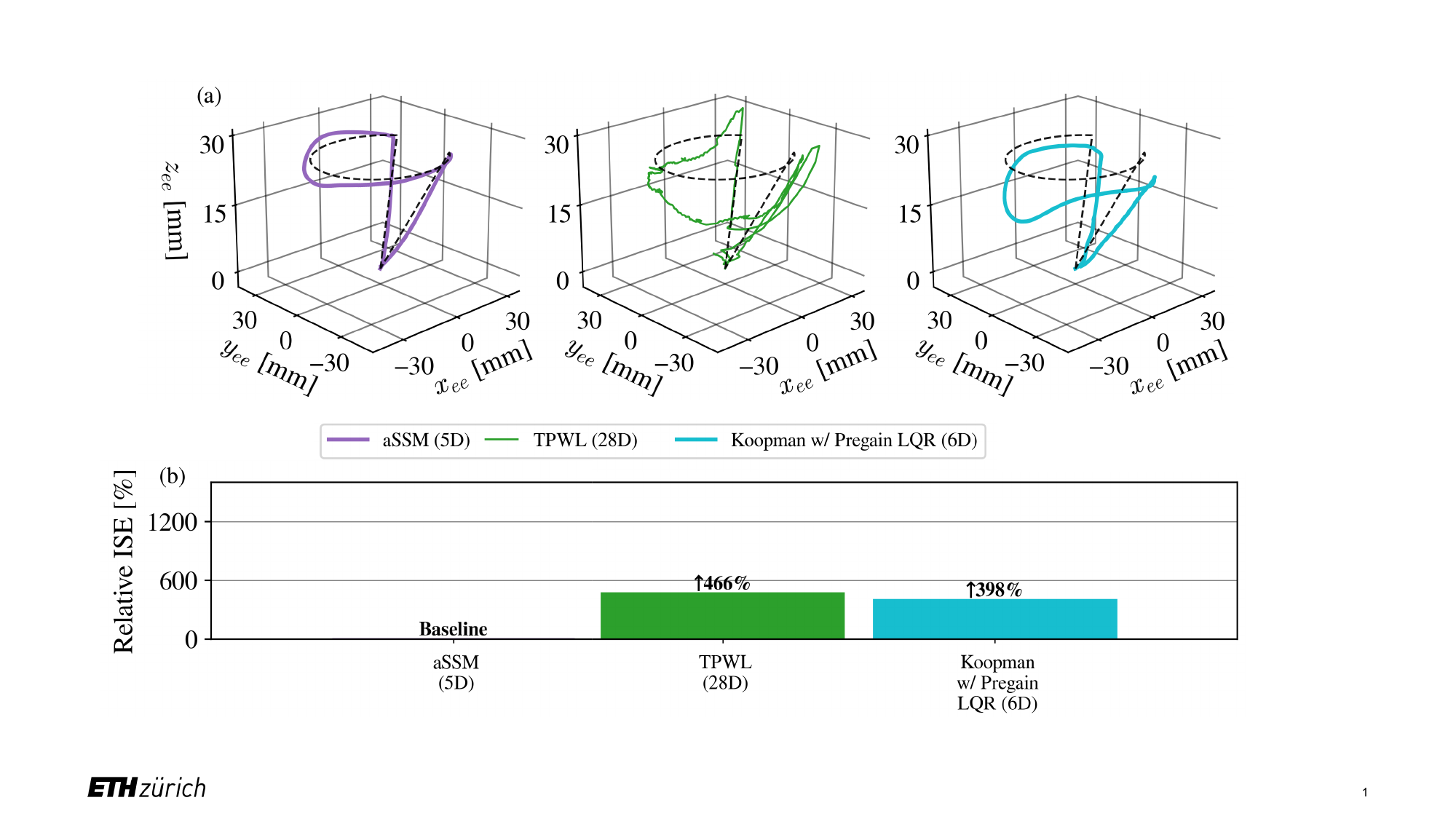}
    \par\end{centering}
    \caption{(a) Closed-loop results using linear and aSSM-reduced methods, for the $3$D Pacman track plotted in the trunk robot's workspace. (b) Bar plots of relative ISE, with the 5D aSSM-reduced model as the baseline. See Multimedia Extension 1 in the Supplementary Material for the full soft trunk robot's evolution.}
    \label{fig:pacman_appendix}
    \end{figure*}

In Fig. \ref{fig:pacman_appendix}, we plot comparisons of the $5$D aSSM-reduced model with linear methods for closed-loop control of the Pacman track. 

We find again that our $5$D aSSM-reduced model has the lowest ISE. This is also the only model capable of lifting the trunk to the right height in the $z_{ee}$ direction. The TPWL method struggles to locate the control inputs that will lift up the trunk. This is because the piecewise linear models have been constructed for inputs that do not correspond to the regions explored by the Pacman. The Koopman pregain approach is again able to trace the target track. However, because of the static assumptions that the model entails, it is unable to make the fast jump from the origin to the start of the Pacman track.

\section{Validation of aSSM-reduced model for short and long soft elastic arms}
\label{app:training_elastic_arm}
\subsection{Steady state distribution}
\begin{figure*}
    \centering
    \subfloat[]{\includegraphics[width=0.45\textwidth]{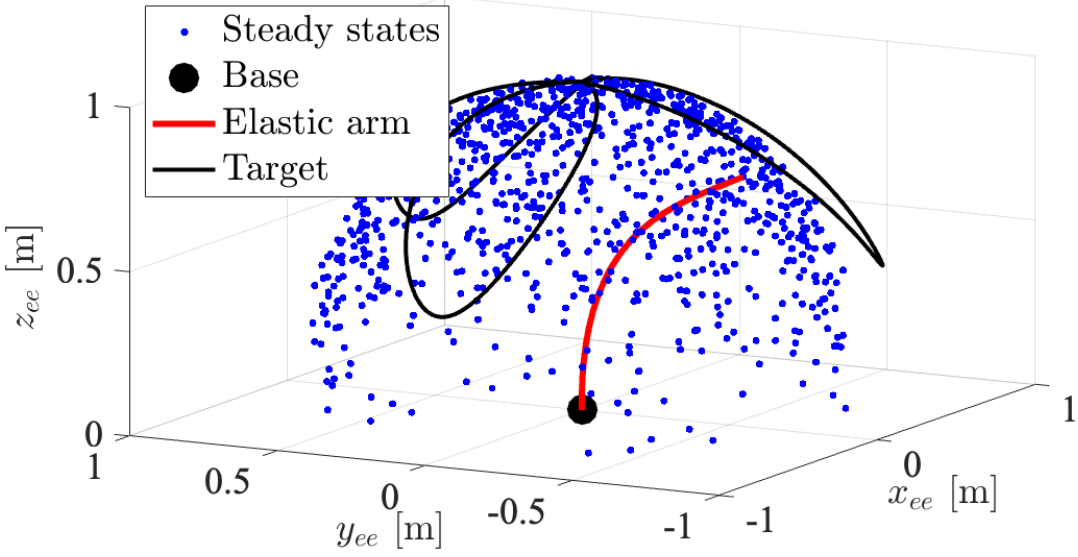}}
    \hspace{0.05\textwidth}
    \subfloat[]{\includegraphics[width=0.45\textwidth]{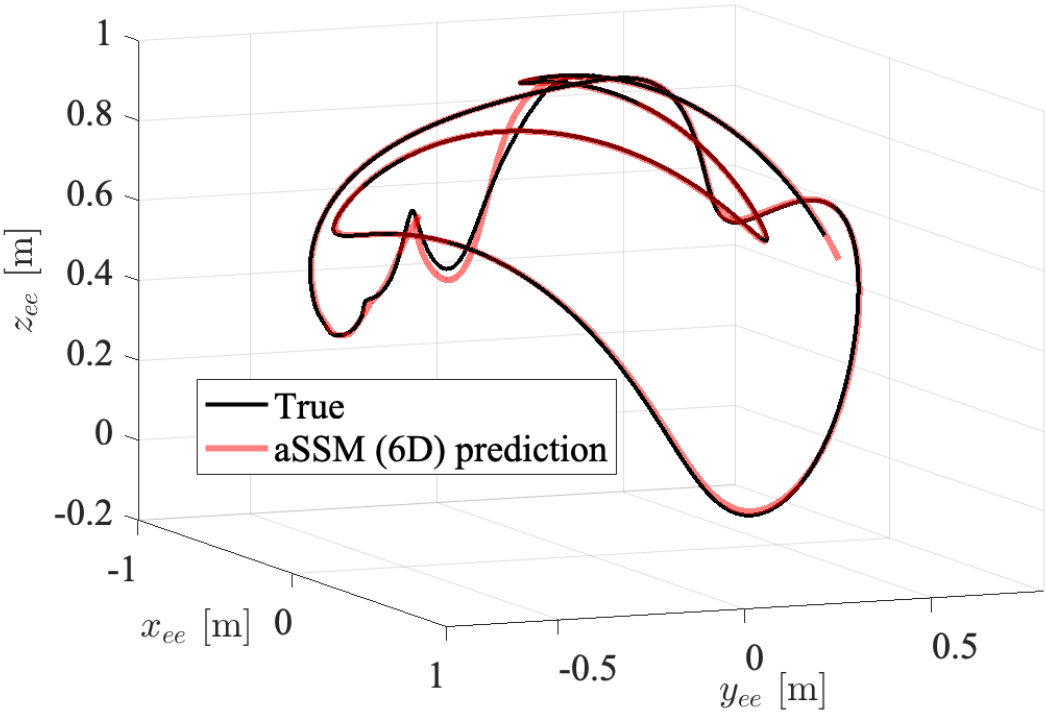}}
    \hspace{0.05\textwidth}
     \subfloat[]{\includegraphics[width=0.45\textwidth]{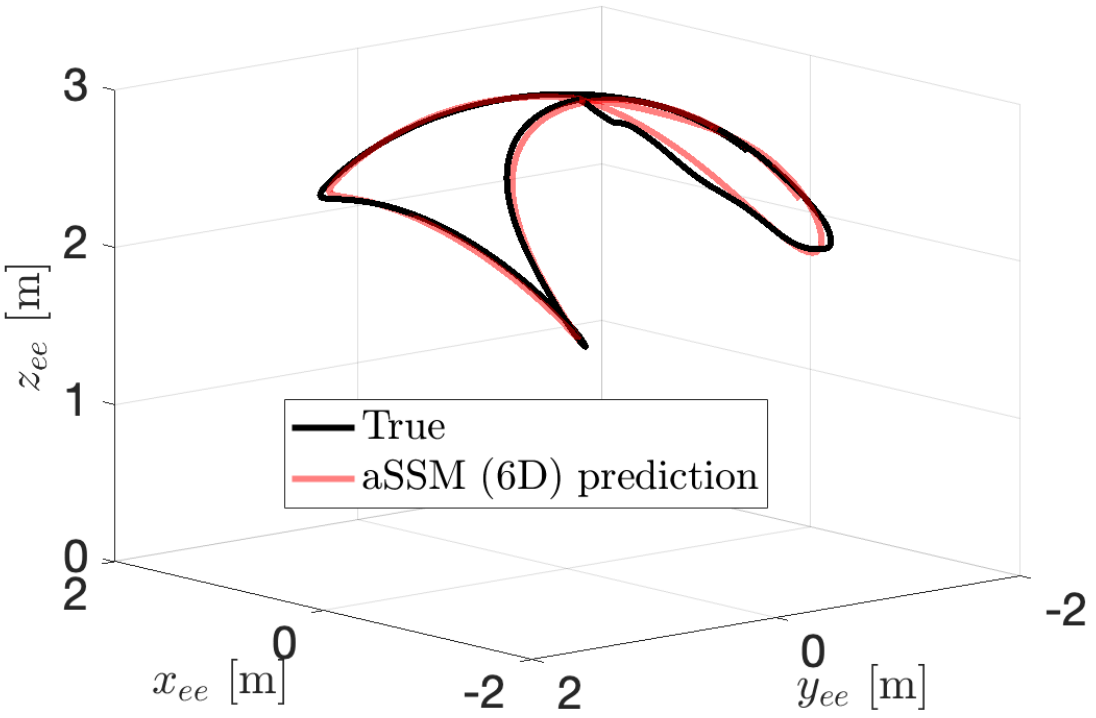}}
     \hspace{0.05\textwidth}
     \subfloat[]{\includegraphics[width=0.45\textwidth]{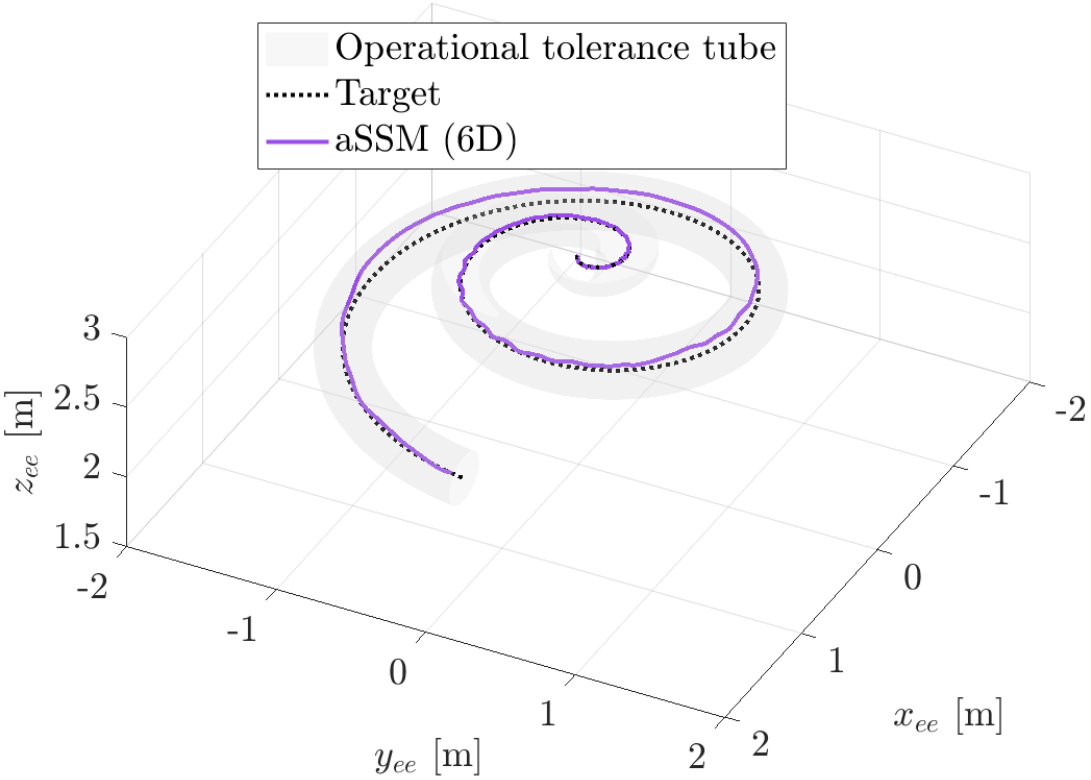}}
    \caption{(a) Short elastic arm steady-state geometry. (b) Test prediction in the short elastic arm's workspace. (c) Test prediction in the long elastic arm's workspace. True solution in black and the aSSM prediction in red. (d) Closed-loop results for the long elastic arm with feedback noise on the spiral track using the aSSM-reduced model.}
    \label{fig:test_aSSM_arm}
\end{figure*}

We isolate $1000$ random static steady states from the collected static input training data. In Fig. \ref{fig:test_aSSM_arm}a, we plot the static steady states in the short arm's workspace. We observe that these states accumulate along a tapered sphere with maximum height $1 \text{ [m]}$ and radial span $0.8 \text{ [m]}$. The tapering is due to under-actuation and limited torque magnitudes. For these reasons, the maximum lobe sections of our planned Trifolium target track cannot be readily reached by the short elastic arm. To assess closed-loop performance, we quantify acceptable error regions by plotting an operational tolerance tube around the target track of radius $15 \text{ [cm]}$. Similar conclusions about the steady-state distributions hold for the long elastic arm, where the tapered sphere has a wider span than in the short arm.  

\subsection{Test result on unseen control input}
We generate responses of the short and long soft arms to a randomly varying 30-second control input, with a slowness measure $r_s = 0.75$, using the undeformed arm configuration as the initial condition. We pass this control input and initial condition to the learned $6$D aSSM-reduced model in the form (\ref{eq:aSSM_reduced}) and output a prediction in the robot's workspace. In Fig. \ref{fig:test_aSSM_arm}b and \ref{fig:test_aSSM_arm}c, we plot the prediction of the aSSM-reduced model and the true response of the short and long soft arms. Our $6$D aSSM-reduced model accurately predicts the transient dynamics from the origin to the slow steady state, as well as the slow steady state itself. The prediction closely matches the ground truth with an NMTE $\approx 9 \%$ for the short soft arm and an NMTE $\approx 11 \%$ for the long soft arm. 

\subsection{Long elastic arm: Closed-loop result in the presence of experimental noise}
\label{app:noise_long_arm}
We add noise sampled from a bounded Gaussian distribution with an upper bound of $10 \text{ [mm]}$ to the feedback term in the aSSM-reduced MPC scheme eq.(\ref{eq:aSSM_mpc}). In Fig. \ref{fig:test_aSSM_arm}d, we plot the closed-loop result for the 6D aSSM-reduced model in the presence of feedback noise. The closed-loop prediction has visible jitter but overall tracks the target and remains within the operational tolerance tube. Here, the operational tolerance tube is enlarged to a radius of 18.5 [cm] to account for the noise uncertainty. The normalized ISE increases by a small factor of 0.01 compared to the case without noise presented in Section \ref{subsec:noise_short_arm}.

\color{black}

\end{document}